%% file: PaperForReview.tex
\documentclass[10pt,twocolumn,letterpaper]{article}

\usepackage[pagenumbers]{wacv} 

\usepackage{graphicx}
\usepackage{amsmath}
\usepackage{amssymb}
\usepackage{booktabs}
\usepackage{multirow}
\usepackage[table,xcdraw]{xcolor}
\usepackage{dsfont}
\newcommand{\cmark}{\ding{51}} %
\newcommand{\xmark}{\ding{55}} %
\usepackage{pifont}
\usepackage{adjustbox}
\usepackage{wrapfig}
\usepackage{stfloats} 
\usepackage{tabularx}

\usepackage[pagebackref,breaklinks,colorlinks]{hyperref}

\usepackage[capitalize]{cleveref}
\crefname{section}{Sec.}{Secs.}
\Crefname{section}{Section}{Sections}
\Crefname{table}{Table}{Tables}
\crefname{table}{Tab.}{Tabs.}

\def\wacvPaperID{788} 
\def\confName{WACV}
\def\confYear{2024}

\begin{document}

\title{Harnessing the Power of Multi-Lingual Datasets for Pre-training: Towards Enhancing Text Spotting Performance}

\newcommand{\superaffil}[2]{\textsuperscript{#1}\,#2}


\author{
\small Alloy Das \superaffil{1}\and
\small Sanket Biswas\superaffil{2}\and
\small Ayan Banerjee\superaffil{2}\and
\small Josep Llad\'{o}s\superaffil{2}\and
\small Umapada Pal\superaffil{1}\and 
\small Saumik Bhattacharya\superaffil{3}\and
\and
\footnotesize{ 
  \textsuperscript{1}CVPR Unit, Indian Statistical Institute, Kolkata
  \quad
  \textsuperscript{2}CVC, Universitat Autònoma de Barcelona
  \quad
  \textsuperscript{3}ECE, Indian Institute of Technology, Kharagpur}
}

\maketitle

\input{tex/abstract}
\input{tex/intro}

\input{tex/related}
\input{tex/method}
\input{tex/sota}
\input{tex/conclusion}
\section*{Acknowledgement}

\noindent
This work acknowledges the Spanish projects GRAIL PID2021-126808OB-I00, DocAI 2021-SGR-01559, the CERCA Program / Generalitat de Catalunya, the FCT-19-15244, and PhD Scholarship from AGAUR 2023 FI-3-00223 and special thanks to Subhankar Ghosh, Prasun Roy and Kunal Biswas for the insightful discussions.

{\small
\bibliographystyle{ieee_fullname}
\bibliography{egbib}
}

\newpage
\input{tex/supply}

\end{document}

%% file: tex/abstract.tex
\begin{figure*}[b]
\centering
\subfloat[TotalText]{\includegraphics[width=0.2\textwidth]{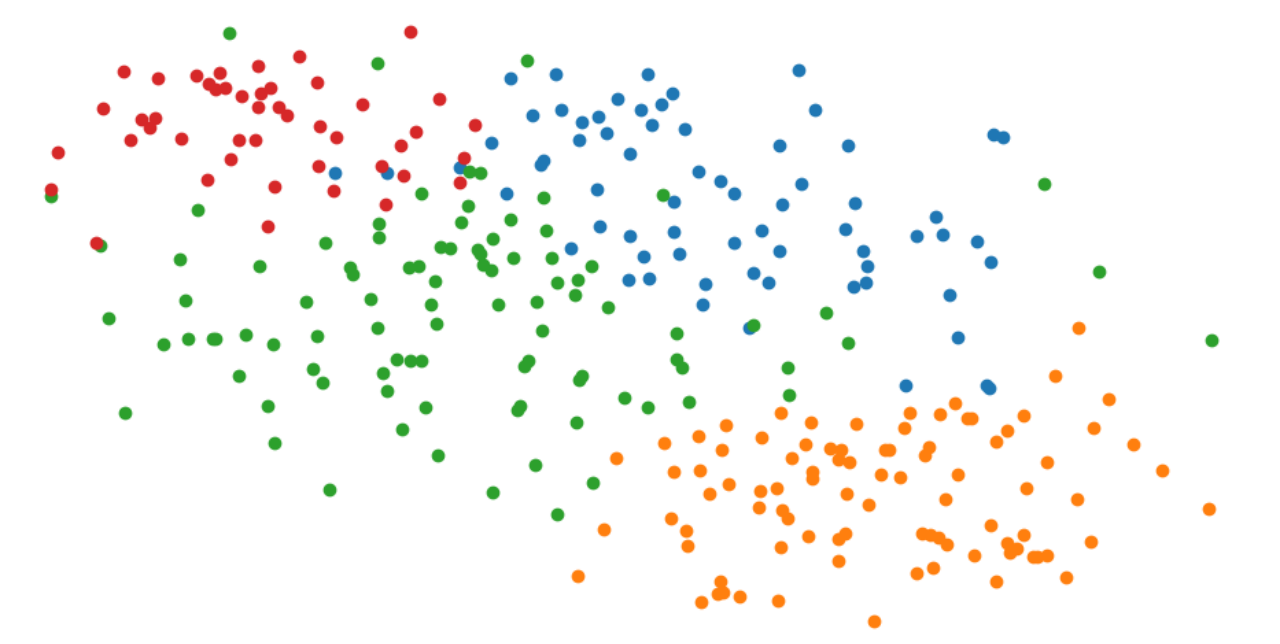}}\hfill
\subfloat[CTW1500]{\includegraphics[width=0.2\textwidth]{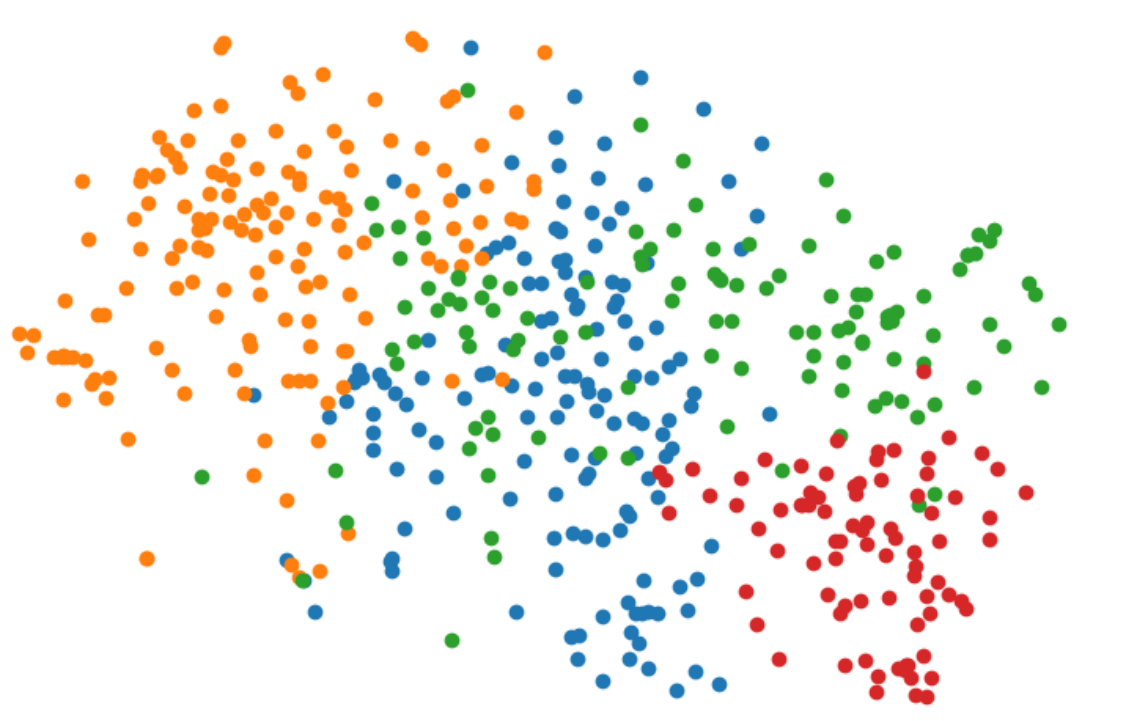}}\hfill
\subfloat[MLT2017]{\includegraphics[width=0.2\textwidth]{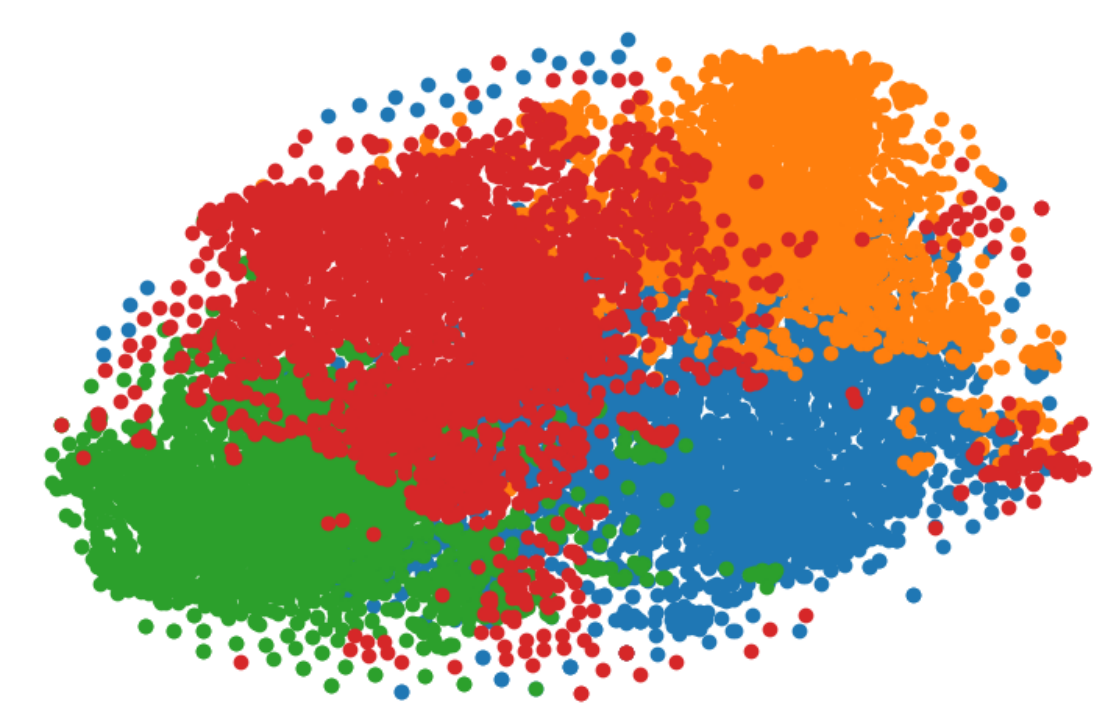}}\hfill
\subfloat[ReCTS]{\includegraphics[width=0.2\textwidth]{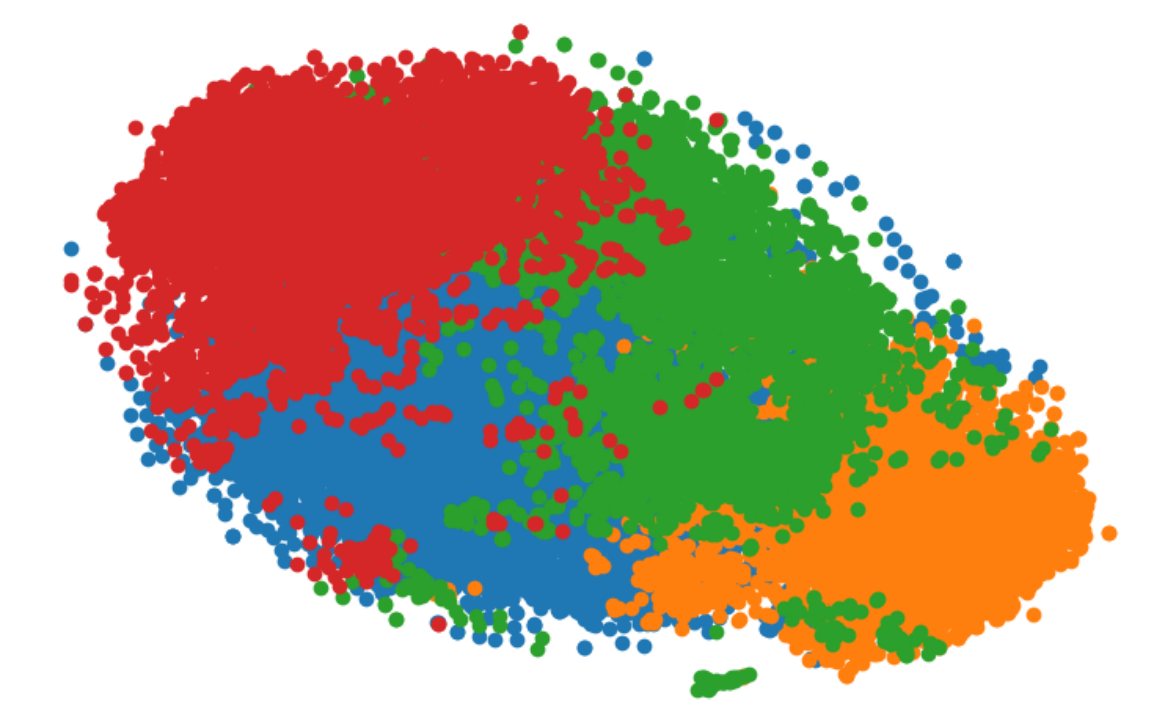}}\hfill
\subfloat[VinText]{\includegraphics[width=0.2\textwidth]{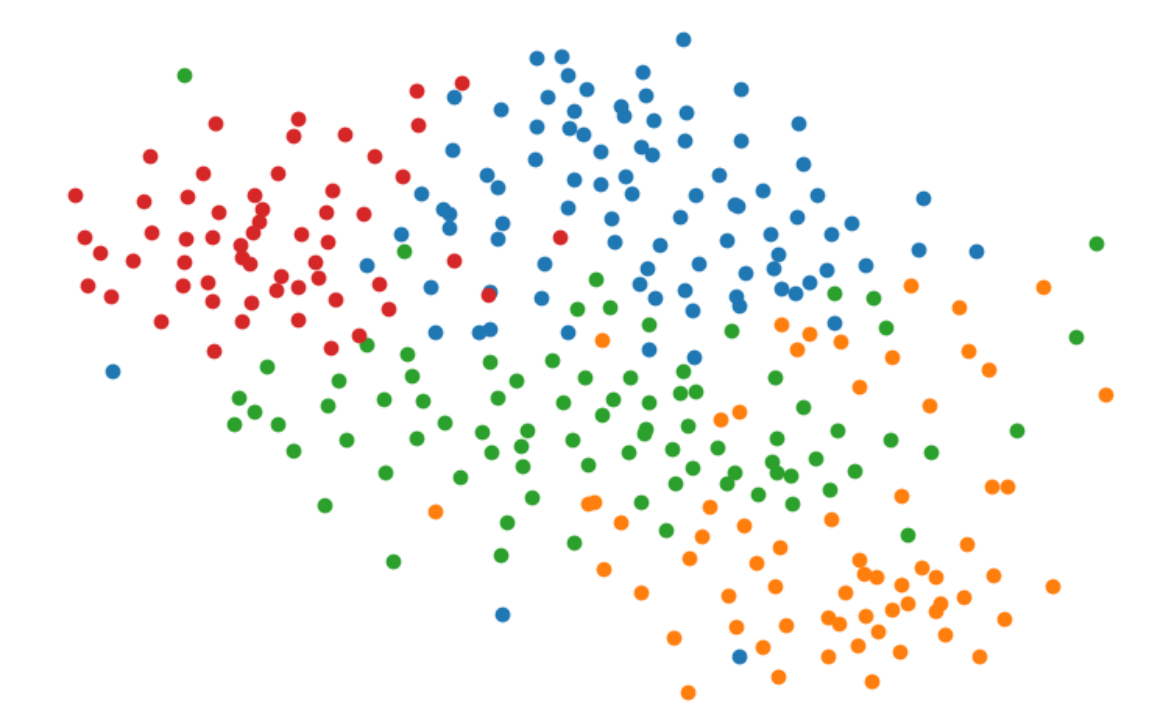}}
\\
\subfloat[TotalText]{\includegraphics[width=0.2\textwidth]{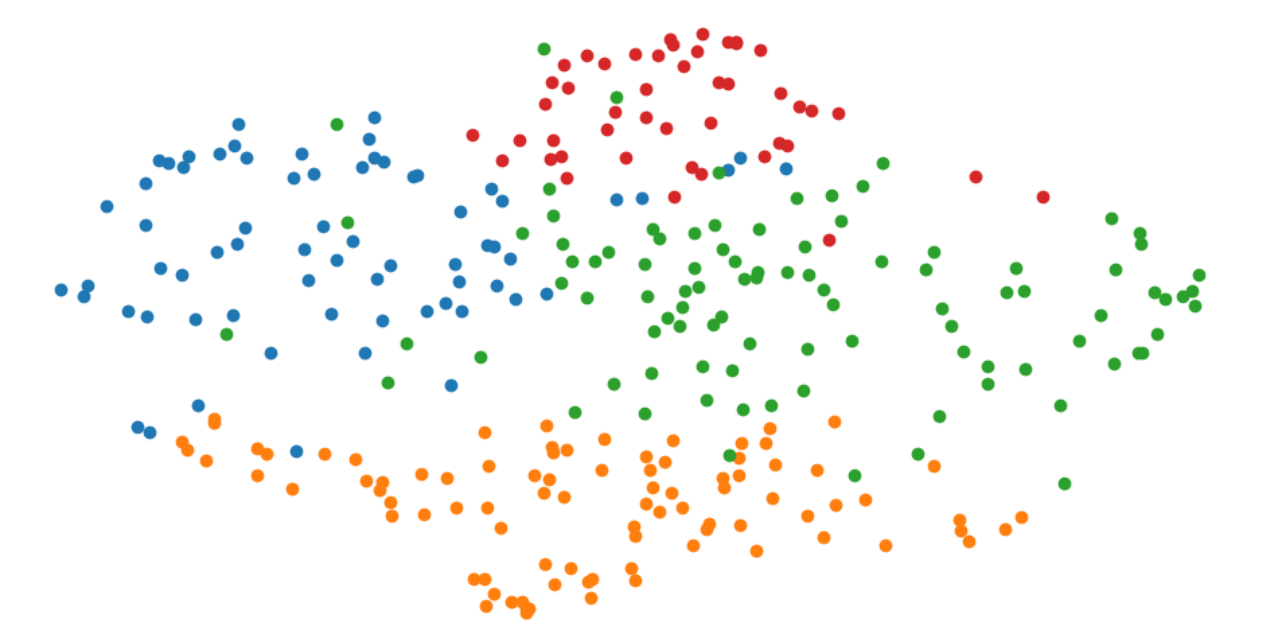}}\hfill
\subfloat[CTW1500]{\includegraphics[width=0.2\textwidth]{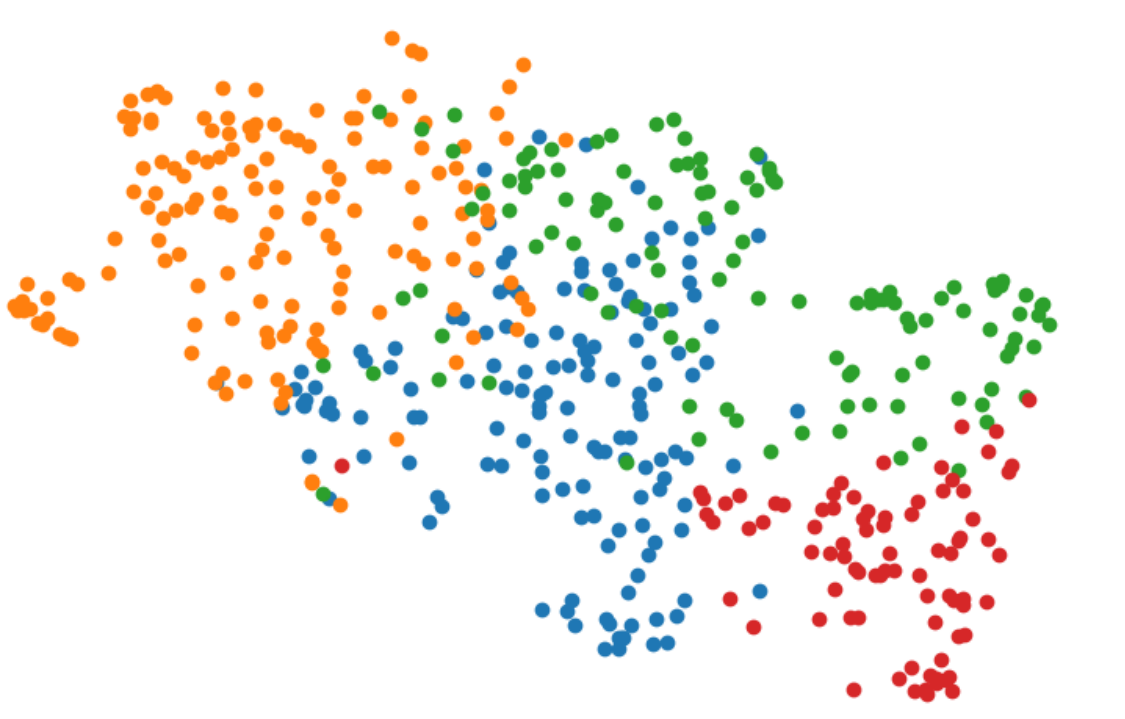}}\hfill
\subfloat[MLT2017]{\includegraphics[width=0.2\textwidth]{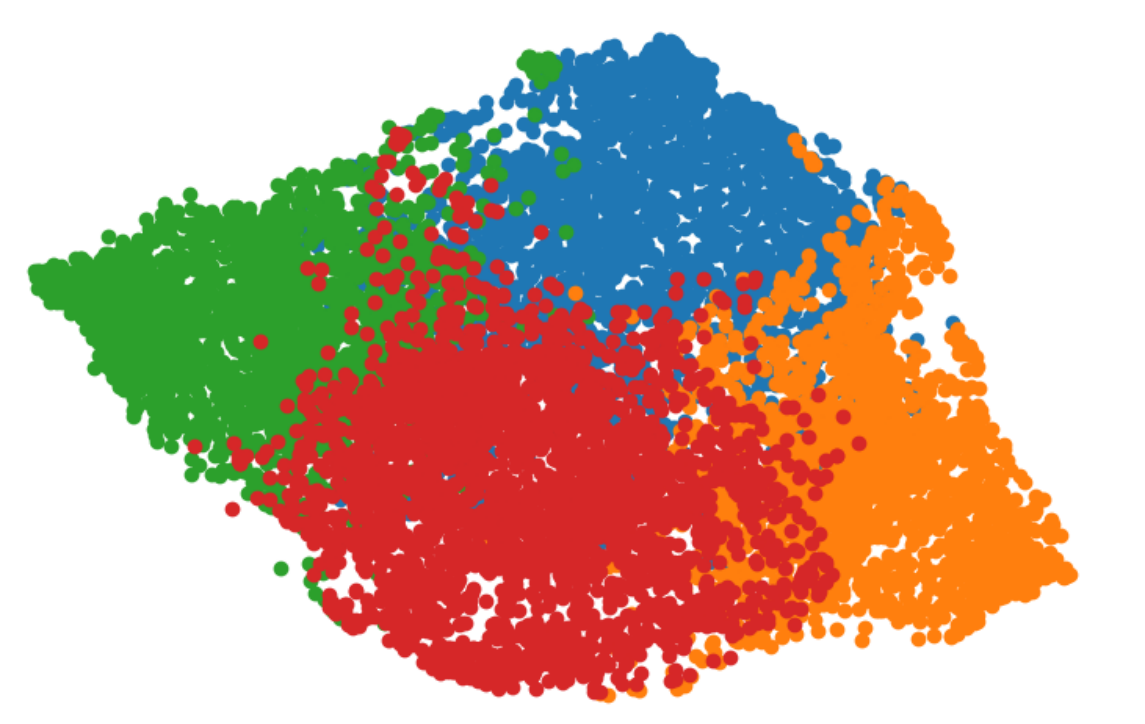}}\hfill
\subfloat[ReCTS]{\includegraphics[width=0.2\textwidth]{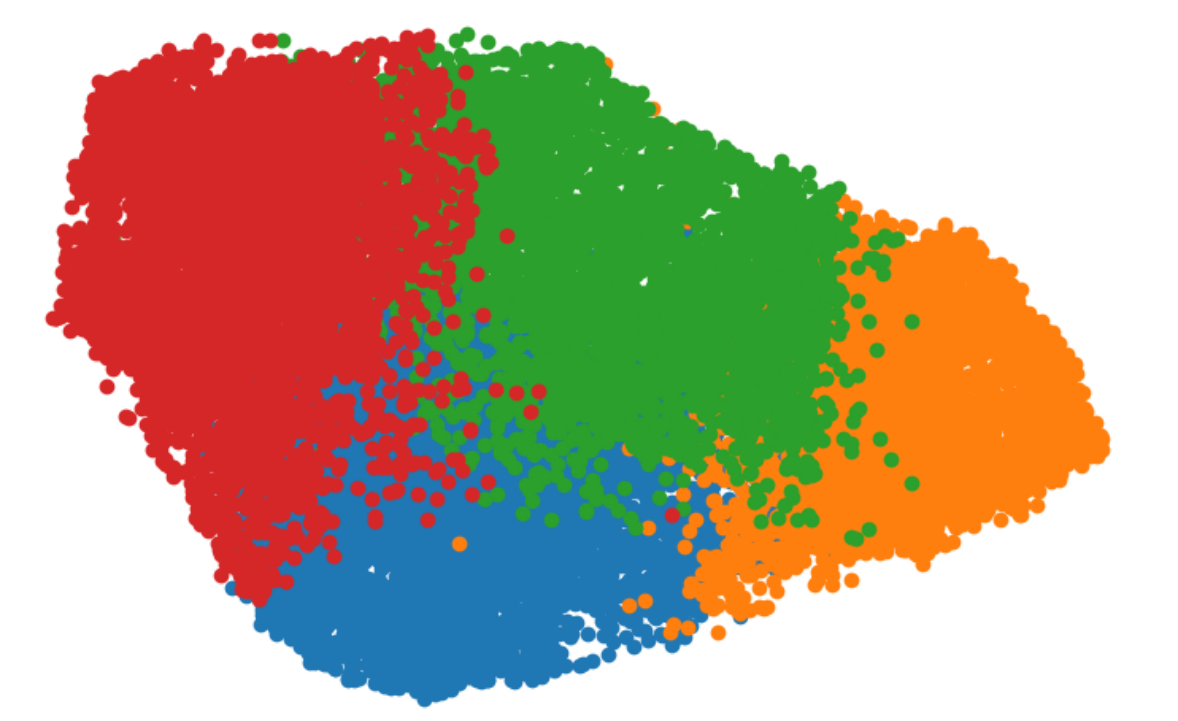}}\hfill
\subfloat[VinText]{\includegraphics[width=0.2\textwidth]{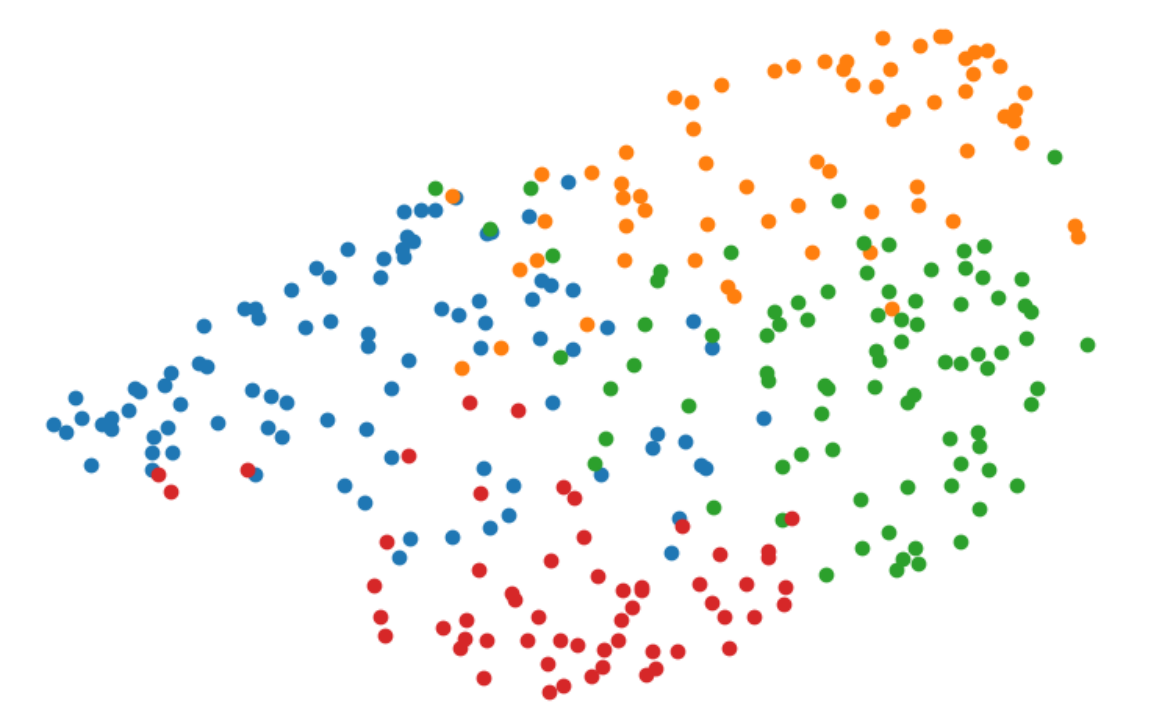}}
\caption{\textbf{Text Spotting Benchmark Diversity.} We employ a t-SNE (top row) and UMAP (bottom row) to visualize the different clusters of similar images corresponding to text orientation characteristics  where \textcolor{blue}{Vertical},  \textcolor{orange}{Circular}, \textcolor{green}{Wavy}, and \textcolor{red}{Horizontal} Curve.}\label{fig:fig1}
\end{figure*}

\begin{abstract}
    The adaptation capability to a wide range of domains is crucial for scene text spotting models when deployed to real-world conditions. However, existing SOTA approaches usually incorporate scene text detection and recognition simply by pretraining on natural scene text datasets, which do not directly exploit the intermediate feature representations between multiple domains. Here, we investigate the problem of domain-adaptive scene text spotting, i.e., training a model on multi-domain source data such that it can directly adapt to target domains rather than being specialized for a specific domain or scenario. Further, we investigate a transformer baseline called \textbf{Swin-TESTR} to focus on solving scene-text spotting for both regular and arbitrary-shaped text along with an exhaustive evaluation. The results demonstrate the potential of intermediate representations to gain significant performance on text spotting benchmarks across multiple domains (e.g. language, synth-to-real, and documents). both in terms of accuracy and efficiency.
\end{abstract}

%% file: tex/intro.tex
\section{Introduction}

End-to-end scene text spotting has been an active research problem in the computer vision community owing to its immense utility in real-world application scenarios like autonomous driving~\cite{garcia2022read,zhang2021character}, intelligent navigation systems~\cite{wang2015bridging}, image retrieval~\cite{jaderberg2016reading} and so on. It can be defined as the joint optimization of scene-text detection and recognition tasks in a unified model pipeline. Research efforts in recent years on text spotting~\cite{fang2022abinet++,huang2022swintextspotter,liu2021abcnet,qiao2021mango,wang2021pan++,zhang2022text} have demonstrated the superiority of deep CNNs and transformer-based approaches on natural scene image benchmarks containing regular text~\cite{karatzas2015icdar}, and arbitrarily-shaped text~\cite{ch2020total,liu2019curved}. \textcolor{black}{However, it is essential for a reading system (OCR) to learn and adapt to new complexities in unseen tasks (domains) without forgetting to read previous tasks (domains)~\cite{xu2020forget}, hence overcoming the phenomenon of \textit{catastrophic forgetting} for neural networks~\cite{goodfellow2013empirical}. }

In recent years, multimodal foundation models pre-trained on multiple source data, have demonstrated impressive performance on several tasks, including text recognition~\cite{yang2021tap,biten2022latr,aberdam2023clipter}, entity extraction~\cite{huang2022layoutlmv3,tang2023unifying}, and document layout analysis~\cite{li2022dit,huang2022layoutlmv3,banerjee2023swindocsegmenter,maity2023selfdocseg}. Given the growing practical significance of pre-training on diverse domains containing both real~\cite{long2022towards} and synthetic data~\cite{liu2020abcnet}, there is a surging interest in analyzing its empirical performance. Recent explorations undertaken in ~\cite{baek2021if} and ~\cite{baek2019character} have delved into different aspects of model design choices and the impact of training and evaluation settings for benchmarking Scene Text Recognition (STR). However, there are open research questions that need to be addressed for a deeper understanding of the performance gain in the SOTA text spotting models. \textbf{Firstly: Given the exuberant amount of data available for pre-training having different sets of text instance properties: which of them are most empirically useful?} This is really difficult to assess given the current training and evaluation scheme followed by every recent SOTA approach. \textbf{Secondly: Why is the multilingual MLT17~\cite{nayef2017icdar2017} most extensively used dataset for pretraining?} As MLT contains samples in multiple languages (6 in total) and scripts (8 in total) in a wide variety of real-world challenging conditions with diverse text instance properties (see Fig.~\ref{fig:fig1}), there are hopes that the intermediate pre-trained representation learned for this task will generalize well to other downstream tasks such as text detection and end-to-end spotting for both arbitrary-shaped text and regular text benchmarks. However, the existing SOTA has not made sufficient attempts towards empirically analyzing this hypothesis. \textbf{Thirdly: Why do we need the domain adaptation settings for scene text spotting?} Since the SOTA text spotting models mostly focus on domain-specific fine-tuning for getting the best results for the target evaluation benchmarks, domain adaptation settings from synthetic (source) to real (target) datasets~\cite{wu2020synthetic, deng2022text} have yet to be investigated more in-depth. The existing domain adaptation approaches utilize adversarial training and discriminative feature alignment to learn some domain-invariant feature representations for text. But these methods are more complex as they are generative and our key objective is to pay attention to the model capacity and computation cost. \textbf{Fourthly: Are the model representations robust enough to be used as an OCR for document layout Analysis (DLA)?} When it comes to measuring model robustness and generalizability, the effectiveness of the text spotting model to detect and read from text regions could be indeed beneficial for DLA tasks without the need for a commercial OCR engine.

\textcolor{black}{Given the bottleneck that annotated real data is expensive and fewer in proportion to the available synthetic data~\cite{gupta2016synthetic}, it further motivates us to study the domain adaptation settings for text spotting task.} Although purely relying on synthetic data doesn't help to get the best model performance, they provide a better initialization point with learnt structural and contextual cues to improve the overall performance with further addition of real data training with far lesser number of iterations. \textcolor{black}{Additionally, as shown in ~\cite{souibgui2022text} a model should inherently learn different degraded scene complexities (like blurs, distortions, occluded objects, ...) to obtain some robustness in it's pre-trained feature representations. In this work, we introduce a simple text spotting baseline called \textit{Swin-TESTR} which performs visual feature extraction using a Swin-Transformer~\cite{liu2021swin} backbone, coupled with a Transformer-based unified single encoder-dual decoder framework as proposed in TESTR~\cite{zhang2022text}. On account of it's higher tranferability~\cite{zhou2021convnets} compared to simple ViTs and CNNs and more interpretable hierarchical feature maps obtained during the feature extraction phase, Swin-TESTR helps to better analyse and establish strong benchmark performances for both domain adaptation and SOTA end-to-end spotting settings.}


\textcolor{black}{The overall contributions of this work can be summarized in four folds:}\textcolor{black}{ 1) To boost the text spotting performance by simply utilizing intermediate representations learned from more diverse and complex datasets (like the multilingual MLT17~\cite{nayef2017icdar2017}, Chinese ReCTS~\cite{zhang2019icdar} or English Total-Text~\cite{ch2020total}), we \textit{analyze the trade-offs for different combinations of tasks (domains) in a domain adaptation setting}. 2) With extensive experimentation, we analyze the \textit{significance of multilingual MLT17 as a strong intermediate representation} after pre-training on synthetic data helps to obtain a fully competitive model for tackling the data distribution shift~\cite{subbaswamy2021evaluating}. 3) A flexible Swin-TESTR baseline has been proposed to show an exhaustive experimental study to address the large domain gap between synthetic and real scene datasets and \textit{explain how diverse intermediate representations could play a role to minimize this gap}. 4) Moreover, we make a \textit{first attempt towards investigating the generalization capability of the Swin-TESTR baseline by evaluating its OCR performance on the DLA task}. Our baseline shows promising results as it outperforms the commercial Microsoft-OCR engine for "text" and "math" regions.} 


%% file: tex/related.tex
\section{Challenges and Limitations in SOTA}

In this section, we examine the different benchmarks that have been adapted for training and evaluation by prior works and address their pitfalls. Through this investigation, we explore the performance inconsistencies across different benchmarks and find shreds of evidence through experimentation and intuitive discussions.  

\subsection{Pitfalls in Pretraining Strategies}
When pre-training text spotting models, the most common practice is to pre-train with the Curved SynthText data containing 150K samples first introduced in~\cite{liu2020abcnet}for detecting arbitrary-shaped scene text. The dataset was curated from the 800K SynText dataset~\cite{gupta2016synthetic} containing quadrilateral bounding boxes. Such synthetic dataset curations have been also used in scene text editing task~\cite{das2023fast,roy2020stefann}. Since then, the most influential SOTA text spotting literature~\cite{liu2020abcnet, liu2021abcnet, fang2022abinet++, qiao2021mango, huang2022swintextspotter, zhang2022text} has persisted with large-scale pretraining with the Curved SynthText. MLT2017~\cite{nayef2017icdar2017} and TotalText~\cite{ch2020total} are the most used real-world datasets that have been used in pre-training the SOTA models. Although the richness and sample diversity of these datasets has been considered to be the key reasons, there have not been enough justifications to prove why they have been chosen. In this work, we try to explore and understand why and how they prove to be beneficial during the model pre-training. Note that the essential pre-training protocols used by the recent methods are mainly of two different kinds. While methods like ABINet++~\cite{fang2022abinet++}, ABCNetv2~\cite{liu2021abcnet} and TESTR~\cite{zhang2022text} mix both synthetic and real data together for pre-training, others like SwinTextSpotter~\cite{huang2022swintextspotter} and MANGO~\cite{qiao2021mango} prefer to generate a synthetic data initialized model checkpoint and then pre-train all the real-world data together on top of it for the rest of the iterations. The problem of how to use real datasets for STR tasks has been previously addressed in~\cite{baek2021if}. Also, the same authors addressed the pitfalls of STR datasets and model evaluation in~\cite{baek2019wrong}. \textcolor{black}{Contrary to the aforementioned works, we delve deep into understanding (see sec.~\ref{domainadaptation} ) the current text spotting practices and finding the balance between the number of datasets used and the model complexities for the overall performance gain.} In this regard, a domain adaptation setting for evaluation has been introduced to actually validate the aforementioned open questions.  

\subsection{Fairness in Evaluating Real-world datasets}
Text spotting has been commonly fragmented into two different sub-tasks in the literature: text detection and recognition. Past approaches~\cite{wang2011end,liao2017textboxes,he2018end,liu2018fots} were mainly evaluated on regular-shaped text spotting benchmarks like ICDAR15~\cite{karatzas2015icdar}. Later,  arbitrary-shaped text benchmarks like Total-Text~\cite{ch2020total} and CTW1500~\cite{liu2019curved} became the standard benchmarks for evaluation protocols.  Recent SOTA text spotting methods ranging from CNN-based~\cite{wang2021pan++,qin2019towards,qiao2020text,liu2020abcnet,liu2021abcnet,qiao2021mango} to recent transformer-based ~\cite{huang2022swintextspotter,fang2022abinet++,zhang2022text,ye2023deepsolo} techniques have mainly done their evaluations with the aforementioned arbitrary-shaped English text benchmarks. Recently, Vintext~\cite{nguyen2021dictionary} and ReCTS~\cite{zhang2019icdar} have emerged as more challenging non-English arbitrary-shaped benchmarks for further tough evaluation. One major issue plaguing all the SOTA methods is they rely heavily upon fine-tuning their models on the corresponding evaluation benchmark to achieve superior performance gain. In this work, we investigate a domain adaptation setting to evaluate the proposed baseline on several combinations of intermediate pretraining and evaluation steps to gain some fairness in the model selection. A few prior works~\cite{wu2020synthetic, das2023diving} have studied such domain distribution mismatches for scene text recognition. Also, we have added model generalization capability when transferred to another task from a different domain as a further evaluation criterion. The closest work related to this task is ~\cite{yu2023turning} and ~\cite{aberdam2023clipter} which harnesses the power of CLIP~\cite{radford2021learning} model for scene text detection and recognition task respectively.

%% file: tex/method.tex
\section{Preliminaries}


The primary goal of our proposed baseline is to learn some domain-agnostic feature representation through an improved intermediate representation  of combining real data with synthetic data in an effective way and address the domain shift problem. Additionally, we also introduce the technical details of our proposed text-spotting architecture with an efficient self-attention unit.

\subsection{Problem Formulation}
In order to formulate the problem statement, let us assume, we have an input space $\chi$ and output space $\xi$. We formulate domain adaptation as a learning problem that is specified by two parameters: a distribution $D$ over $\chi$ (i.e. the domain) and a relation function $R: \chi \rightarrow \xi$ which maps instances to features. Basically, it is a representation that induces the distribution $D$ over $\xi$ to perform the same task in multiple domains. Now what are our source (input space $\chi$) and target domains (output space $\xi$)? We define three pairs of source-target domains towards text spotting. 

\textbf{1. Language to Language:} We take the source domain as the English language and perform training over it and adapt it to the Vietnamese and Chinese text domains.

\textbf{2. Synth to Real:} We initialize training on a synthetic dataset and adapt it to the real text domain. Here we observed that, a potential complex intermediate real dataset (e.g. MLT 2017) boosts up the domain adaptation process.

\textbf{3. Scene Text to Document:} Several previous works \cite{shen2021layoutparser, huang2022layoutlmv3} used OCR tokens in order to boost the performance of Document Layout Analysis. We replace those OCR tokens with the extracted information by the text spotting framework and evaluate it using the same baseline.


\textcolor{black}{In order to formulate the problem of domain adaptation, Let $D_s$ denote the feature distribution of the source domain, $Y_s$ denote the labels of the source domain distribution over the input space $\chi$ and $D_t$ denote the feature distribution of the target domain, $Y_t$ denote the labels of the target domain distribution over $\xi$. The goal of domain adaptation is to learn a labeling rule $f: (D_s, Y_s) \rightarrow (D_t, Y_t)$ that can generalize well on the target domain using labeled data from the source domain and potentially some labeled data from the target domain. We can formulate it as finding a mapping function $f$ that minimizes the distribution mismatch between the source and target domains while ensuring good generalization on the target domain as defined in eq. \ref{eq1}.}

\begin{equation}
\label{eq1}
\min_{f} [\mathcal{L}_s(f) + \lambda \cdot \mathcal{L}_t(f)]
\end{equation}

\textcolor{black}{where, $\mathcal{L}_s(f)$ is the loss on the source domain data, encouraging the model to fit the source domain, $\mathcal{L}_t(f)$  is a distribution discrepancy or adaptation loss between the source and target domains, capturing the differences between the two domains, and $\lambda$ is a balancing parameter that controls the trade-off between fitting the source domain and adapting to the target domain. The goal of solving this eq. \ref{eq1} is to learn a model $f$ that can generalize effectively on the target domain even when the distribution of the source and target domains is different. The focus is on mitigating the negative effects of domain shift and improving the transferability of the learned knowledge from the source to the target domain.}
\begin{figure}[t]
\centering
\includegraphics[width=.48\textwidth]{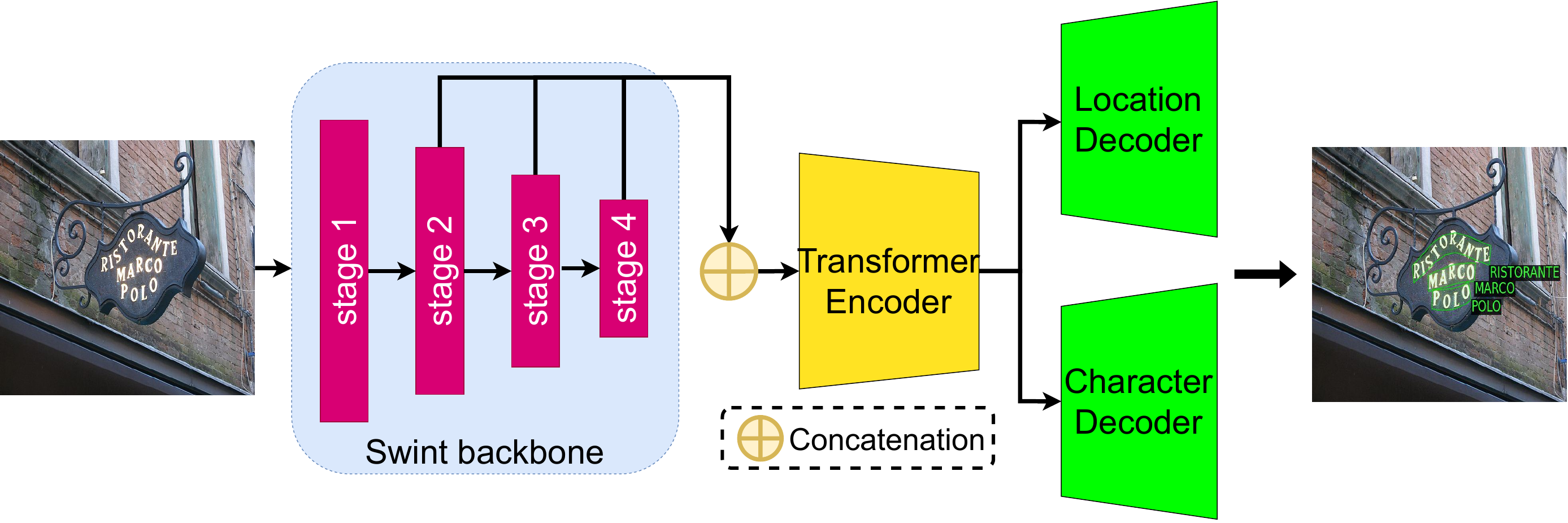}
\caption{Schematic overview of the model Swin-TESTR.}
\label{fig2}
\vspace{-6mm}
\end{figure}
\subsection{Model Architecture}
The overall architectural pipeline as illustrated in Fig.\ref{fig2} consists of three major components: (1) a visual feature extraction unit based on a Swin-Transformer~\cite{liu2021swin} backbone to extract multi-scale features; (2) a text spotting unit consisting of a Transformer encoder to encode the image features into positional object queries and then two Transformer decoder units to predict the location of text instances and recognize the corresponding characters respectively.\\ 
\noindent
\textbf{Visual Feature Extraction Unit.} It is hard to connect remote features with vanilla convolutions since they operate locally at fixed sizes (e.g., $3 \times 3$). Text spotting requires capturing the relationships between different texts since scene text from the same image has a similar representation with respect to the text background, style and texture. For our backbone, we chose to use a small and efficient Swin-Transformer~\cite{liu2021swin} unit denoted as Swin-tiny for extraction of more fine-grained image features.\\
\noindent
\textbf{Text Spotting Unit.} The text-spotting unit is mainly composed of a transformer encoder and two transformer decoders for text detection and recognition following a similar schema as proposed in the TESTR framework~\cite{zhang2022text}. Accordingly, we formulate our problem as a set prediction problem as in DETR~\cite{carion2020end}, to predict a set consisting of  point-character tuples, for a particular image. We formulate it as $X = \{(S^{(i)}, R^{(i)})\}^K_{i=1}$. Where i is the index of each instances, $S^{(i)} = ({{s^{(i)}_1}, ...., s^{(i)}_M)}$ is the coordinates of $M$ control points, and $R^{(i)} = ({{r^{(i)}_1}, ...., r^{(i)}_M)}$ is the $M$ characters of the text. The text location decoder (TLD) will detect (predict $S^{(i)}$) while the text recognition decoder (TRD) will recognize (predict $R^{(i)}$) the text in a unified manner.

%% file: tex/sota.tex
\section{Experiments and Analysis}
\begin{table*}[ht]
\centering
\caption{\textbf{Domain Adaptation on Total-Text and CTW-1500.} Method implies pre-trained on the first dataset and then trained on the next dataset, Mix pre-train means pre-trained on SynthText, ICDAR MLT17, ICDAR15, and Total-Text. Results style: \textbf{best}, \underline{second best}.}
\begin{adjustbox}{max width=\textwidth}
\begin{tabular}{lcccccccccc}
\toprule
& \multicolumn{5}{c}{Total-Text} & \multicolumn{5}{c}{CTW-1500}\\ \cmidrule{2-6} \cmidrule{7-11}
\multirow{2}{*}{Method}& \multicolumn{3}{c}{\multirow{1}{*}{Detection}} &\multicolumn{2}{c}{End-to-End} &  \multicolumn{3}{c}{Detection} &\multicolumn{2}{c}{End-to-End} \\\cmidrule(lr){2-4} \cmidrule(lr){5-6} \cmidrule(lr){7-9} \cmidrule(lr){10-11} 
& P& R & F&None & Full & P& R & F&None & Full \\  \hline
Synth-Text   & 76.79&28.09& 41.13& 27.54 & 38.54 & 50.37&15.25& 23.42& 12.45 & 19.25\\
Synth-Text $\rightarrow$ ICDAR MLT17 &90.02&59.17&82.74&62.97& 77.98 &38.48&49.00&43.10&28.24& 33.84\\ 
Synth-Text $\rightarrow$ ICDAR15 &90.24&60.16&72.20&55.76& 69.74&36.20&28.53&31.91&19.64 & 24.3\\ 
Synth-Text $\rightarrow$ ICDAR MLT17 $\rightarrow$ ICDAR15 &90.52&67.71&77.47&58.14& 74.48 &39.94&31.10&34.97&23.47& 27.23\\ 
Synth-Text $\rightarrow$ ICDAR MLT17 $\rightarrow$ Total-Text &92.01&\textbf{85.82}&\textbf{88.81}&\underline{71.33}& 83.17 &39.76&56.85&46.7&28.23 & 35.46\\ 
Synth-Text $\rightarrow$ ICDAR MLT17 $\rightarrow$ CTW1500 &66.23&36.77&47.28&24.11 & 41.54 &\underline{92.61}&\underline{82.55}&\underline{87.29}& \underline{54.88}& \underline{81.94}\\
Synth-Text $\rightarrow$ Total-Text &\underline{92.19}&78.95&85.06&70.51& \underline{81.06} &43.86&51.04&47.18&29.33& 35.63\\ 
Synth-Text $\rightarrow$ CTW1500 &62.14&33.29&43.35&21.03 & 37.3 &\textbf{94.11}&77.27&84.86 &47.56 & 80.2\\ 
Mix pre-train & \textbf{93.59} & 75.2 & 83.4 & 67.06 & 80.51 & 39.19 & 36.35 &	37.71 &	25.83 & 29.57\\
ICDAR MLT17 &	79.35 &	55.37 & 65.23 & 15.44 & 36.02& 33.46 & 32.28 & 32.81 & 0.077 & 20.03\\
ICDAR MLT17 $\rightarrow$ Synth-Text &	77.98 &	26.56 &	39.62 &	21.09 & 34.45&	59.27 &	26.41 &	36.54 & 17.81 & 30.37\\
Synth-Text $\rightarrow$ ReCTS & 87.08 & 36.22 & 51.16 & 33.17 & \textcolor{black}{0.02} & 41.89 & 23.36 & 30.00 & 19.11& \textcolor{black}{0.04}\\ \hline
Mix pre-train $\rightarrow$ Fine-tune &90.58& {\underline{85.46}}& \underline{87.95} & \textbf{75.14}&\textbf{86.0}&91.45 & \textbf{85.16} & \textbf{88.19} &\textbf{56.02}  &\textbf{82.91}\\
\bottomrule
\end{tabular}
\end{adjustbox}

\label{tab:my-table_analisys1}
\end{table*}

\begin{table*}[ht]
\centering
\caption{\textbf{Domain Adaptation Experiments on Total-Text and CTW-1500.} Method implies pre-trained on the first dataset and then trained on the next dataset, Mix pre-train means pre-trained on SynthText, ICDAR MLT17, ICDAR15, and Total-Text.}
\begin{adjustbox}{max width=\textwidth}
\begin{tabular}{lccccccccccc}
\toprule
& \multicolumn{7}{c}{ICDAR 2015} & \multicolumn{4}{c}{ReCTS}\\ \cmidrule(lr){2-8} \cmidrule(lr){9-12}
\multirow{2}{*}{Method}& \multicolumn{3}{c}{\multirow{1}{*}{Detection}} &\multicolumn{4}{c}{End-to-End} &  \multicolumn{3}{c}{Detection} & End-to-End \\\cmidrule(lr){2-4} \cmidrule(lr){5-8} \cmidrule(lr){9-11} \cmidrule(lr){12-12} 
& P& R & F&S & W & G & None & P& R & F & 1-NED \\  \hline
Synth-Text   & 76.81 & 26.62 & 39.54 & 37.19 & 32.78 & 26.6 & 20.02 & 61.43&12.85&21.25&05.98\\
Synth-Text $\rightarrow$ ICDAR MLT17 &89.49&84.02&\underline{86.67}&82.95&75.15&67.69&\underline{57.36} & \underline{78.62}&\textbf{73.78}& \underline{76.12}& \underline{14.77} \\ 
Synth-Text $\rightarrow$ ICDAR15 &89.83&74.82&81.64&77.07&70.91&62.51&52.82 & 47.35&21.51&29.58&10.24\\ 
Synth-Text $\rightarrow$ ICDAR MLT17 $\rightarrow$ ICDAR15 &93.42&\underline{79.25} &85.75&\underline{81.48} &\underline{75.88}&\underline{68.14}&57.17 & 45.34&32.59&37.64&11.01\\ 
Synth-Text $\rightarrow$ ICDAR MLT17 $\rightarrow$ Total-Text &84.27&77.61&80.8&54.99&72.01&64.81& 54.09&60.96&29.17&39.46&14.56\\ 
Synth-Text $\rightarrow$ ICDAR MLT17 $\rightarrow$ CTW1500 &69.84&41.69&52.22&	24.78&	42.95&	35.69& 24.78& 52.16&30.91&38.81&12.94\\
Synth-Text $\rightarrow$ Total-Text &77.27&77.56&77.41&46.9&	66.59&	58.41&46.95&45.38&32.65&37.98&11.43\\ 
Synth-Text $\rightarrow$ CTW1500 &57.93&39.58&47.03&	19.22&37.09&30.61&19.22&41.87&23.34&29.98&09.13\\ 
Mix pre-train & \textbf{95.25}&76.26&84.71&81.39&75.36&68.04&57.29& 73.32& 64.46& \underline{68.61}& 13.30\\
ICDAR MLT17 &75.27&70.49&72.8&46.16&33.93&21.68&10.94&65.85&59.49&62.51&06.62\\
ICDAR MLT17 $\rightarrow$ Synth-Text &79.68&33.41&47.08&43.42&37.23&30.55&21.76& 60.92&19.65&29.72&07.22\\
Synth-Text $\rightarrow$ ReCTS & 84.48&44.29&58.12&\textcolor{black}{1.54}&\textcolor{black}{0.05}&\textcolor{black}{0.05}&18.19&\textbf{82.95}&\underline{72.67} &\textbf{77.47}&\textbf{48.75}\\ \hline
Mix pre-train $\rightarrow$ Fine-tune& \underline{95.03} & \textbf{85.70}  & {\textbf{90.13}} &\textbf{86.63} & \textbf{81.67} & \textbf{75.44} &\textbf{66.46} &\textcolor{black}{79.24} &\textcolor{black}{56.39} &\textcolor{black}{66.19}&\textcolor{black}{38.12}\\
\bottomrule
\end{tabular}
\end{adjustbox}
\label{tab:my-table_analisys2}
\end{table*}

For the purpose of validation, we have considered some
important benchmark dataset e.g. the ICDAR 2015~\cite{karatzas2015icdar}, Total-Text~\cite{ch2020total}, CTW1500~\cite{liu2019curved}, VinText~\cite{nguyen2021dictionary}, Curved SynthText 150K~\cite{liu2020abcnet}, ICDAR 2017 MLT~\cite{nayef2017icdar2017}, and ReCTS~\cite{zhang2019icdar} with different text orientation categorizations. Our experimental evaluation suggests that the
domain adaptation through intermediate representation advances the state-of-the-art. Moreover, extensive ablation studies have been performed on Total-Text~\cite{ch2020total}
to show the contribution of some important elements of Swin-TESTR. For more dataset details please refer to the supplementary materials. The code is be publicly available on \href{https://github.com/alloydas/TESTR_Eval}{github}.

\begin{figure*}[ht]
\subfloat[Synth to Real]{%
  \includegraphics[width=0.5\linewidth,height=4cm]{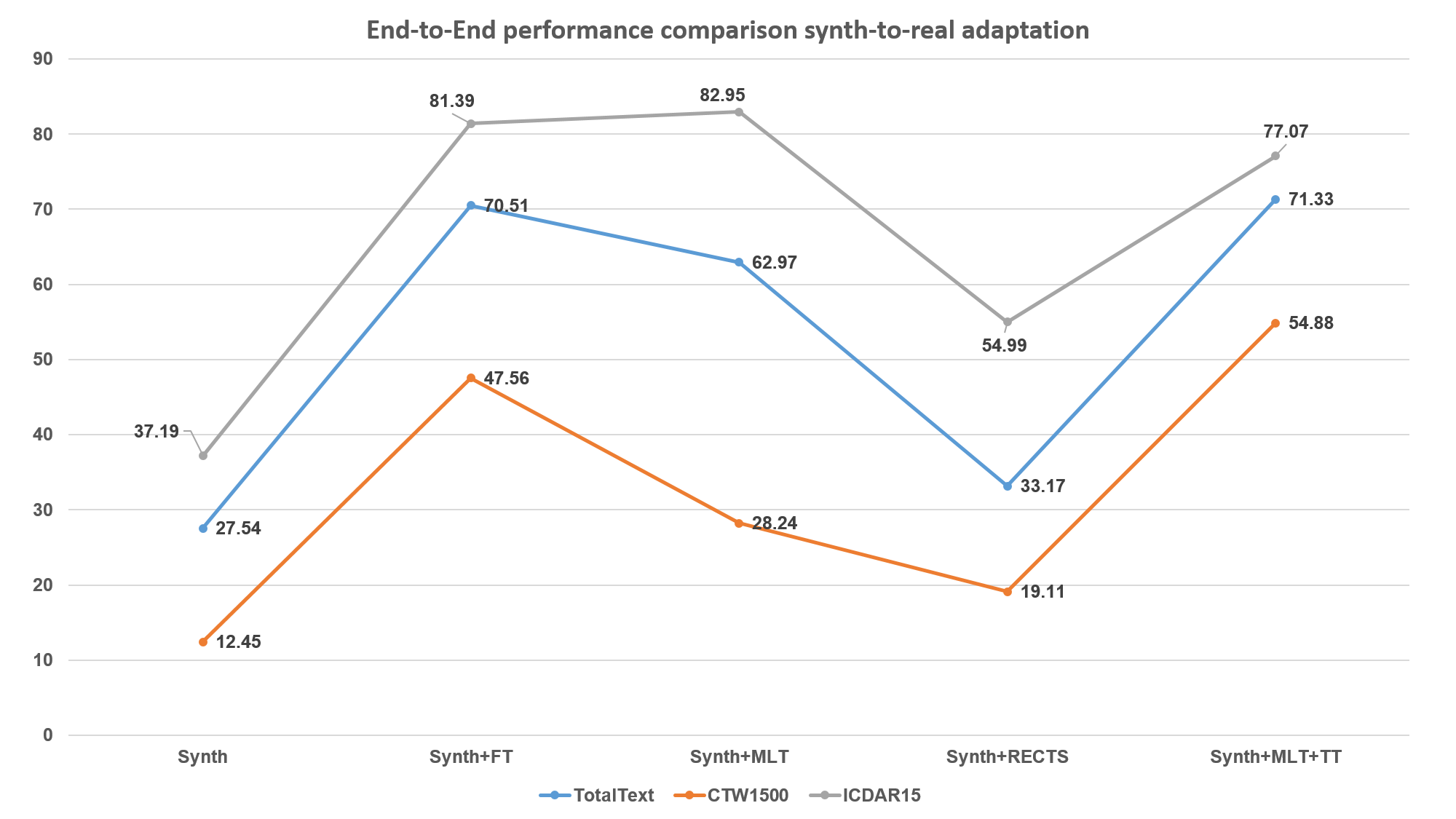}%
}\quad
\subfloat[Real-to-Real]{%
  \includegraphics[width=0.5\linewidth,height=4cm]{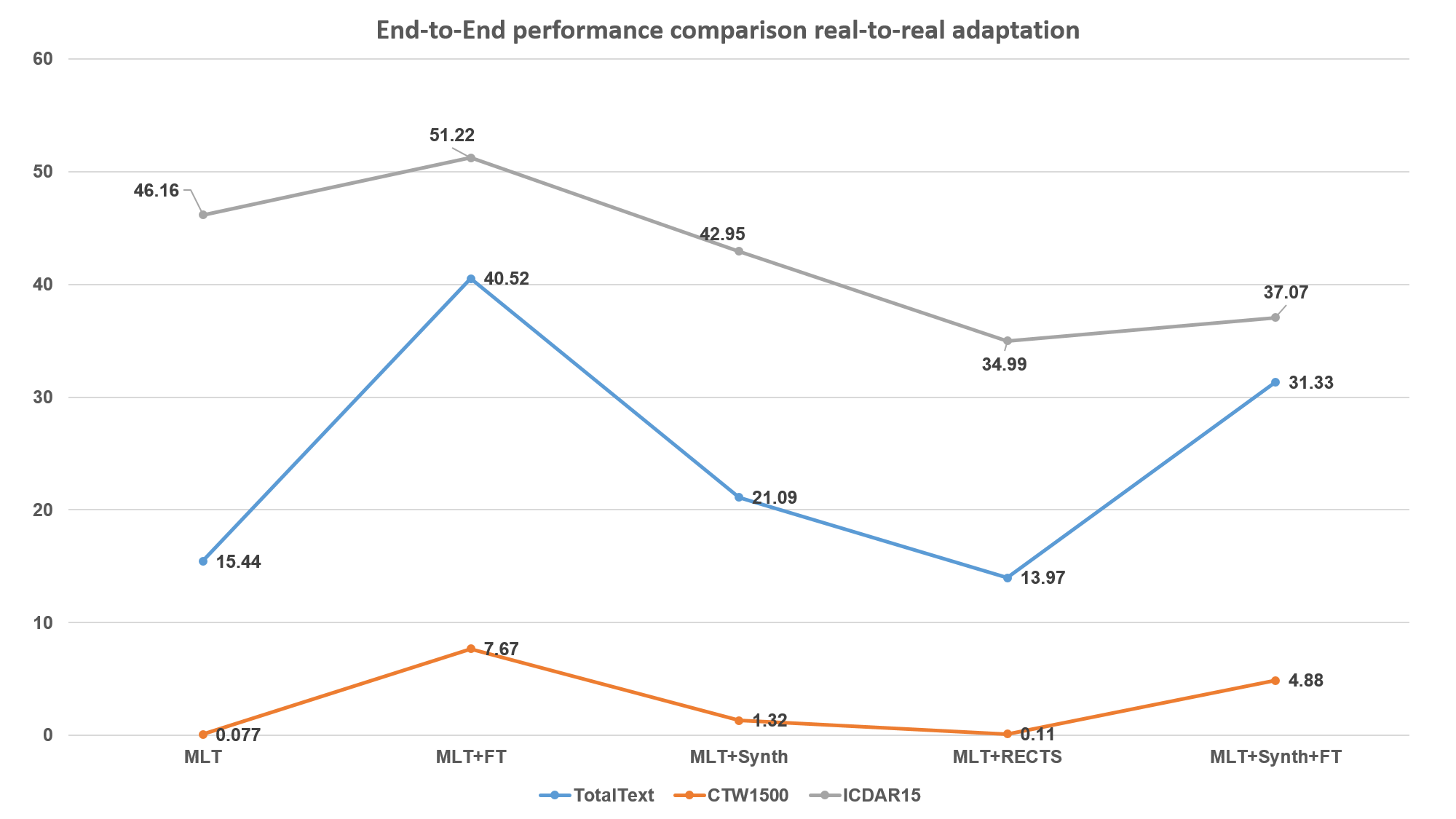}%
}
\caption{\textbf{Different Adaptation Settings:} Illustration of our baseline when combined with intermediate representations. Zoom in for better visualization.}
\label{fig:figlineplot}
\end{figure*}

\subsection{Implementation Details}
The hyper-parameters for the deformable transformer~\cite{zhu2020deformable} have been kept similar to the original work, with the number of heads = 8 and using 4 sampling points for deformable attention. The number of encoder and decoder layers is initialized to 6.

\begin{table*}[t]
\centering
\caption{ Scene text spotting results on Total-Text and CTW1500 and ICDAR 2015. ``None'' refers to recognition without any lexicon. The ``Full'' lexicon contains all the words in the test set.  “S”, “W”, and “G” represents recognition with “Strong”, “Weak”, “Generic”, lexica, respectively. For the models, which do not perform detection we denote it by N/A.}
\begin{tabular}{@{}ccccccccccc@{}}
\toprule
 & \multicolumn{3}{c}{Total-Text}&  \multicolumn{3}{c}{CTW1500} & \multicolumn{4}{c}{ICDAR 15}\\ \cmidrule(lr){2-4} \cmidrule(lr){5-7} \cmidrule(lr){8-11}
Methods& Detection&  \multicolumn{2}{c}{End-to-end} & Detection &  \multicolumn{2}{c}{End-to-end} &  Detection&  \multicolumn{3}{c}{End-to-end} \\ \cmidrule(lr){2-2} \cmidrule(lr){3-4}\cmidrule(lr){5-5}\cmidrule(lr){6-7}\cmidrule(lr){8-8}\cmidrule(lr){9-11}
 & H-mean& None& Full&  H-mean & None& Full &H-mean &S&W&G\\ \hline
Text Perceptron\cite{qiao2020text}&85.2&69.7&78.3 &84.6& 57.0& N/A & 87.5 & 83.4& 79.9& 68.0\\
ABCNet v2\cite{liu2021abcnet} &87.0&70.4 &78.1&84.7&57.5& 77.2 &88.1& 82.7& 78.5& 73.0 \\
MANGO\cite{qiao2021mango}&N/A&72.9&83.6& N/A& 58.9 & 78.7&N/A&81.8& 78.9& 67.3\\
TESTR\cite{zhang2022text}&  86.90&73.25& 83.9& 86.3&53.3& 79.9&90.0& 85.2& 79.4& 73.6\\
Swintextspotter\cite{huang2022swintextspotter}& 88.0&  74.3 & 84.1&  88.0&51.8& 77.0&N/A&83.9& 77.3& 70.5\\ 
Abinet++\cite{fang2022abinet++}& N/A &\textbf{79.4} & 85.4& N/A & \textbf{61.5} &81.2& N/A &86.1& \textbf{81.9}& \textbf{77.8 }\\\hline
\textbf{Swin-TESTR}& 87.95 & 75.14&\textbf{86.0}& \textbf{88.19} &56.02  &\textbf{82.91} & {\textbf{90.13}} &\textbf{86.63} & 81.67 & 75.44\\
\bottomrule
\end{tabular}
\label{tab:my-tablesota}
\end{table*}

\noindent
\textbf{Data Augmentation.} During pre-training, we augment the data with a random resize with the shorter edge (the length of the shorter edge is from 480 to 896, and the longest edge is kept within 1600 \textcolor{black}{(i.e. the length of the shorter edge (either width or height) of the input images is randomly resized to a value between 480 pixels and 896 pixels, and while resizing the shorter edge, it's ensured that the length of the longest edge (either width or height) does not exceed 1600 pixels.). This constraint is applied to prevent excessively large images, which can be computationally expensive to process. An instance-aware random crop (cropping operation takes into consideration the objects or instances within the image) has also been used to make the model more robust to text variations (e.g. curved, wavy, and so on) and positions within the images.}

\noindent
\textbf{Training Details.} \textcolor{black}{To compare with the state-of-the-art methods, shown in Table~\ref{tab:my-tablesota} we pre-trained Swin-TESTR with synthtext150k~\cite{liu2020abcnet}, MLT 2017~\cite{nayef2017icdar2017} for 4000K iteration with a learning rate of $2 \times 10^{-5}$ and is decayed at 3000-Kth iteration by a factor of 0.1. We use AdamW~\cite{loshchilov2017decoupled} optimizer, with $\beta_1 = 0.9$, $\beta_2= 0.999$ and a weight decay of $ 10^{-5}$. We use $Q = 100$ composite queries. The number of control points is 20 and the maximum text length is set to 25.} 

\begin{table}[ht]
\centering
\caption{Performance evaluation Vintext datasets.}
\begin{tabular}{cc}
\toprule
Methods& H-mean \\ \hline
ABCNet~\cite{liu2020abcnet}& 54.2\\
ABCNet $+$ D~\cite{nguyen2021dictionary}& 57.4\\
Mask Textspotter v3~\cite{nguyen2021dictionary}& 53.4 \\
Mask Textspotter v3 $+$ D~\cite{nguyen2021dictionary}&68.5 \\
Swintextspotter\cite{huang2022swintextspotter}&71.1\\ \hline
Swin-TESTR(w/o fine-tune)&21.54\\
Swin-TESTR&\textbf{73.20}\\
\bottomrule
\end{tabular}
\label{tab:my-tablevintext}
\vspace{-4mm}
\end{table}

\subsection{Domain Adaptation Experiments}\label{domainadaptation}
\textcolor{black}{Through our Domain Adaptation experiments, we have demonstrated the Swin-TESTR framework's ability to seamlessly transition between synthetic and real domains, as well as adapt within real domains, all without the need for re-training or fine-tuning on specific datasets. While it's well-known that fine-tuning typically results in superior performance compared to generalization, our findings indicate that the performance gap is remarkably narrow.}

\textcolor{black}{In Table \ref{tab:my-table_analisys1} and Table \ref{tab:my-table_analisys2}, we present our results using SynthText as the base, with re-training involving ICDAR 2017 MLT on the top of SynthText pre-trained weights. Traditionally, it has been common practice to pre-train models using a combination of synthetic and real data. However, in Table \ref{tab:my-table_analisys1}, we delve into the impact of pre-training with individual datasets, shedding light on how this approach affects our model's performance.}

\textcolor{black}{Some valuable insights gleaned from our experimental analysis include:}

\begin{table*}[ht]
\centering
\caption{Domain Adaptation performance from scene text to document}
\begin{adjustbox}{max width=\textwidth}
\begin{tabular}{@{}cccccccccc@{}}
\toprule
                     & Text & Image & Table & Math & Separator & other & AP   & AP@0.5 & AP@0.75 \\ \midrule
Layout Parser~\cite{shen2021layoutparser}       & 83.1 & 73.6  & 95.4  & 75.6 & 20.6      & 39.7  & 64.7 & 77.6   & 71.6    \\
Layout Parser (our OCR) & \textbf{85.2} & 64.7  & 90.2  & \textbf{77.1} & 11.2      & 28.1  & 59.8 & 68.1   & 61.7    \\
LayoutLMv3~\cite{huang2022layoutlmv3}           & 70.8 & 50.1  & 42.5  & 46.5 & 9.6       & 17.4  & 40.3 & 49.4   & 42.7    \\
    LayoutLMv3 (our OCR)    & \textbf{72.1} & 47.8  & \textbf{43.5}  & \textbf{47.2} & 1.8       & 20.2  & 38.7 & 45.2   & 40.8    \\ \bottomrule
\end{tabular}
\end{adjustbox}
\label{tab:my-tabledoc}
\end{table*}

\noindent
\textbf{MLT dataset gives the maximum performance boost during pretraining.} The existing SOTA  approaches do not highlight the impact of individual datasets during pre-training. MLT dataset has quite a lot of variability among the real scene-text benchmarks as demonstrated in Fig.~\ref{fig:fig1} and could be highly beneficial when it comes to domain adaptation performance as exhibited in both Table~\ref{tab:my-table_analisys1} and Table~\ref{tab:my-table_analisys2}. This approach also helps us to \textit{outperform SwinTextSpotter~\cite{huang2022swintextspotter} on TotalText (with fine-tuning) with 88.8\% detection performance} as reported. Not only that, the same pre-training strategy with MLT helps us to get state-of-the-art results on the ICDAR15 benchmark with a detection performance of 90.13\% as shown in Table~\ref{tab:my-tablesota}.   
\\
\noindent
\textbf{Domain Adaptation could be a great alternative to fine-tuning.} Without fine-tuning we achieve really competitive results on both the word-level benchmarks, arbitrary-shaped TotalText, and regular-shaped ICDAR 15. The difference between the performances for detection and end-to-end recognition (with and without fine-tuning) in the above benchmarks is quite marginal. This shows we could indeed have an efficient and robust model which do not require any fine-tuning on the corresponding evaluation dataset. Also, Fig.~\ref{fig:figlineplot} demonstrates how domain adaptation settings gain a performance boost under synthetic-to-real when compared to real-to-real scenario.\\
\noindent
\textbf{More variable text orientations make the dataset a better pre-training candidate.} As inferred from Fig.~\ref{fig:fig1}, the more the separability in terms of text orientations, corresponds to learning better representations during pretraining like as in Total-Text. However, even more dense and complex candidates like MLT or ReCTS as shown can help the model to learn more richer representations in terms of text orientation. \textcolor{black}{As the representations of the MLT17 and ReCTS both are quite dense and diverse in Fig.~\ref{fig:fig1}, we need to analyze further to understand which one provides better intermediate representations. In order to do that, we observe the perceptual and structural similarity scores with the FID, FSIM, and SSIM metrics(See Table \ref{Tab:3}). Results demonstrate that MLT17 has the closest perceptual similarity with Total-Text, which is why they are the most preferred choices for SOTA pretraining. Structurally with SSIM, Total-Text is mostly similar to CTW1500, which helps immensely to improve the performance for CTW1500 evaluation.}\\
\begin{table}[ht]
\vspace{-4mm}
\centering
\caption{The dataset diversion.  Results style: \textbf{best}, \underline{second best}}
\label{Tab:3}
\resizebox{\linewidth}{!}{%
\begin{tabular}{clccc}
\hline
Source Dataset              & Target Dataset & FID $\downarrow$   & FSIM $\uparrow$  & SSIM $\uparrow$ \\ \hline
\multirow{5}{*}{MLT2017}    & Total-Text     & \textbf{0.82}  & \textbf{25.23} & 21.22 \\ \cline{2-5} 
                            & CTW1500        & 6.97  & 18.12 & 16.98 \\ \cline{2-5} 
                            & ICDAR15        & 14.78 & 14.20 & 13.25 \\ \cline{2-5} 
                            & ReCTS          & 11.23 & 6.24  & 5.17  \\ \cline{2-5} 
                            & VinText        & 2.22  & 21.48 & 19.07 \\ \hline
\multirow{4}{*}{Total-Text} & CTW1500        & 1.29  & \underline{24.27} & \textbf{23.32} \\ \cline{2-5} 
                            & ICDAR15        & 9.74  & 16.67 & 15.12 \\ \cline{2-5} 
                            & ReCTS          & 12.14 & 4.37  & 3.54  \\ \cline{2-5} 
                            & VinText        & 2.43  & 19.32 & 19.14 \\ \hline
\multirow{3}{*}{CTW1500}    & ICDAR15        & 0.97  & 23.12 & \underline{22.87} \\ \cline{2-5} 
                            & ReCTS          & 16.32 & 2.22  & 1.46  \\ \cline{2-5} 
                            & VinText        & \underline{0.92}  & 20.27 & 20.12 \\ \hline
\multirow{2}{*}{ReCTS}      & ICDAR15        & 20.20 & 1.05  & 1.03  \\ \cline{2-5} 
                            & VinText        & 12.32 & 6.98  & 4.42  \\ \hline
\end{tabular}
}
\end{table}

\noindent
\textbf{Can Swin-TESTR representations help to read and benefit in document layout analysis task?}
As shown in Table~\ref{tab:my-tabledoc}, we compare the reading ability of our Swin-TESTR model for document layout analysis as we evaluate the performance of detecting several layout regions in documents by utilizing the OCR from our model. The performance demonstrates that it is quite comparable to the original document layout analysis~\cite{shen2021layoutparser,huang2022layoutlmv3} OCR and it outperforms the original model for text, math and table regions. This further justifies the information extraction capability of Swin-TESTR, although it gives a considerably poor performance on the separator, other and image regions due to its initialization with the text spotting weights which has never seen those kinds of instances.

\subsection{Comparison with State-of-the-art}
\begin{figure*}[ht]
\centering
\begin{adjustbox}{max width=\textwidth}
\begin{tabular}{c c c c } 
\multicolumn{2}{c}{(a) Total-Text~\cite{ch2020total}} & \multicolumn{2}{c}{(b) CTW 1500~\cite{liu2019curved}} \\
\includegraphics[height=2cm, width= 2cm]{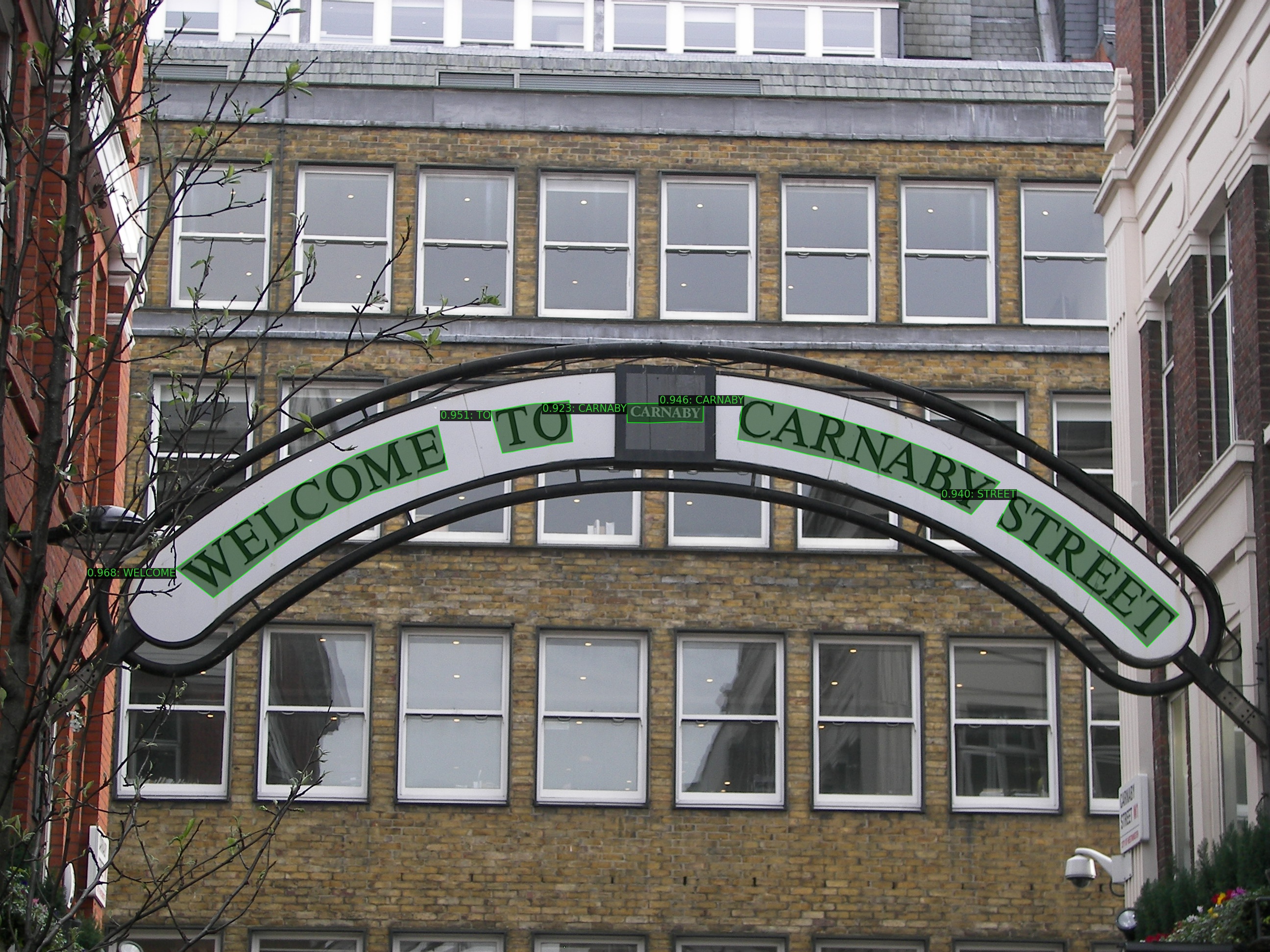}&
\includegraphics[height=2cm, width= 6cm]{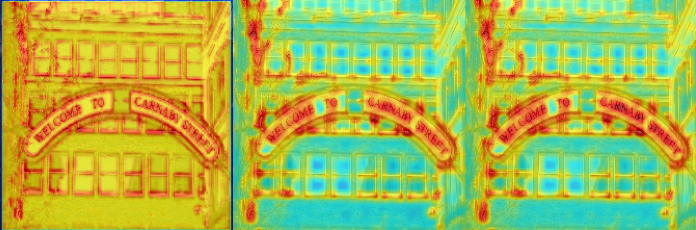}&
\includegraphics[height=2cm, width= 2cm]{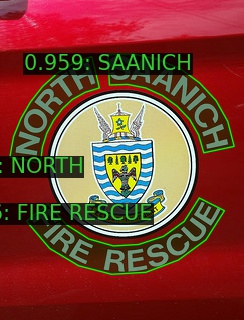}&
\includegraphics[height=2cm, width= 6cm]{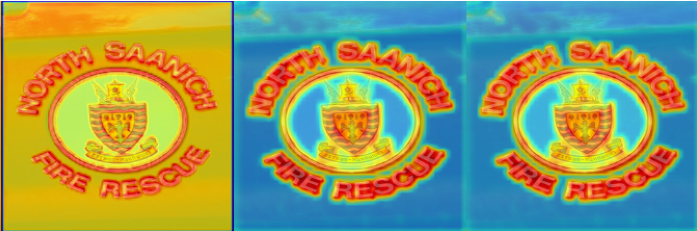}\\
\multicolumn{2}{c}{(c) ICDAR 2015~\cite{karatzas2015icdar}} & \multicolumn{2}{c}{(d) Vintext~\cite{nguyen2021dictionary}} \\
\includegraphics[height=2cm, width= 2cm]{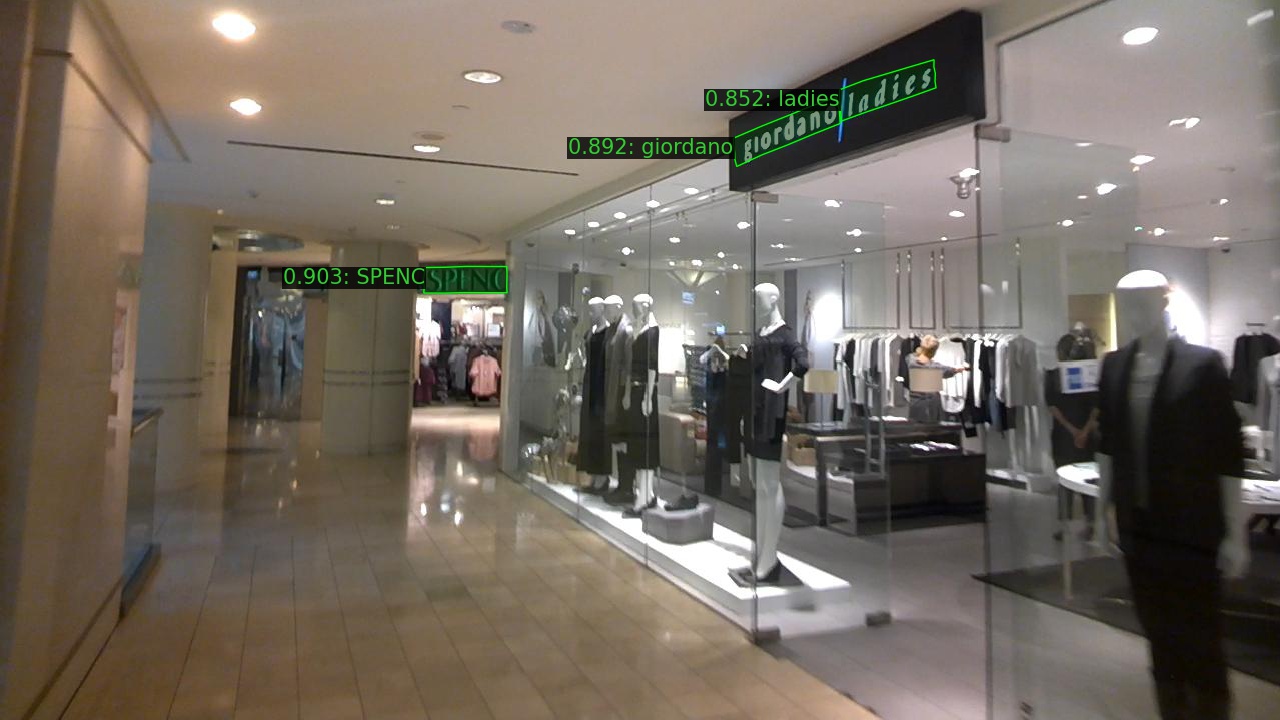}&
\includegraphics[height=2cm, width= 6cm]{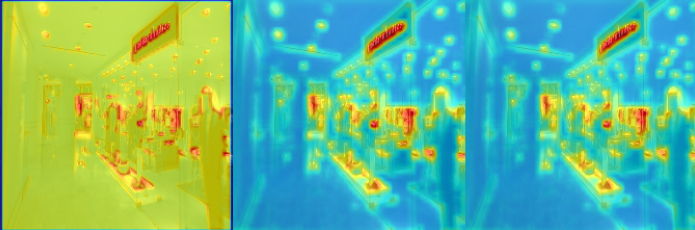}&
\includegraphics[height=2cm, width= 2cm]{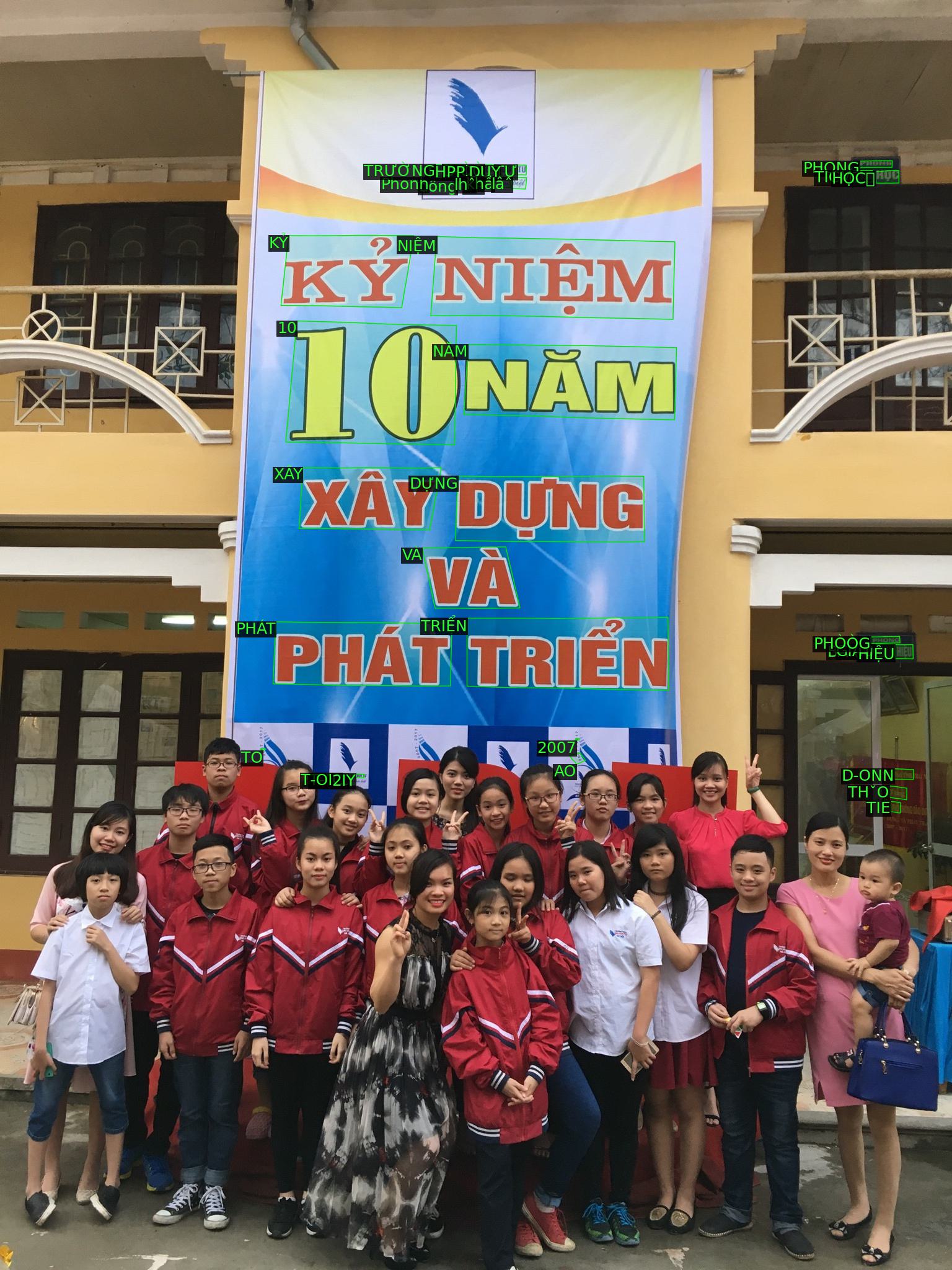}&
\includegraphics[height=2cm, width= 6cm]{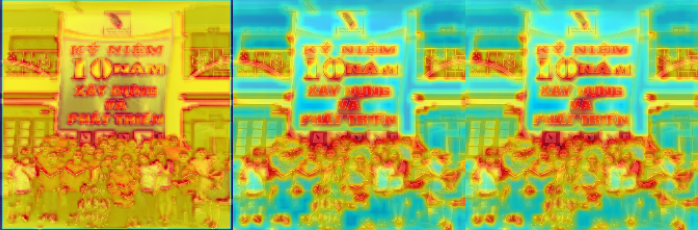}\\
\multicolumn{2}{c}{(e) MLT 2017~\cite{nayef2017icdar2017}} & \multicolumn{2}{c}{(f) ReCTS~\cite{zhang2019icdar}} \\
\includegraphics[height=2cm, width= 2cm]{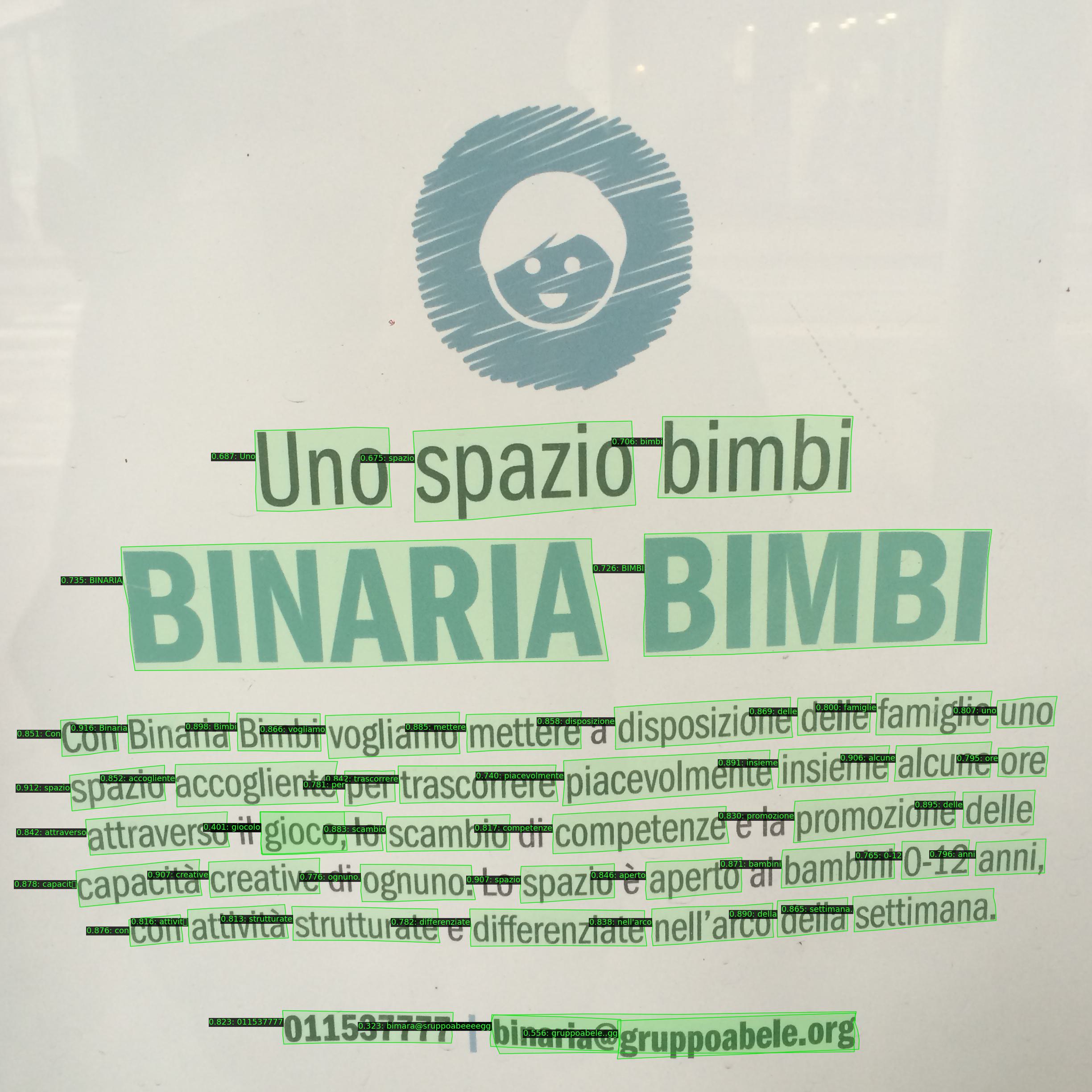}&
\includegraphics[height=2cm, width= 6cm]{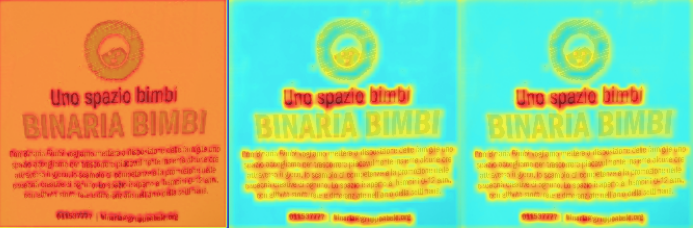}&
\includegraphics[height=2cm, width= 2cm]{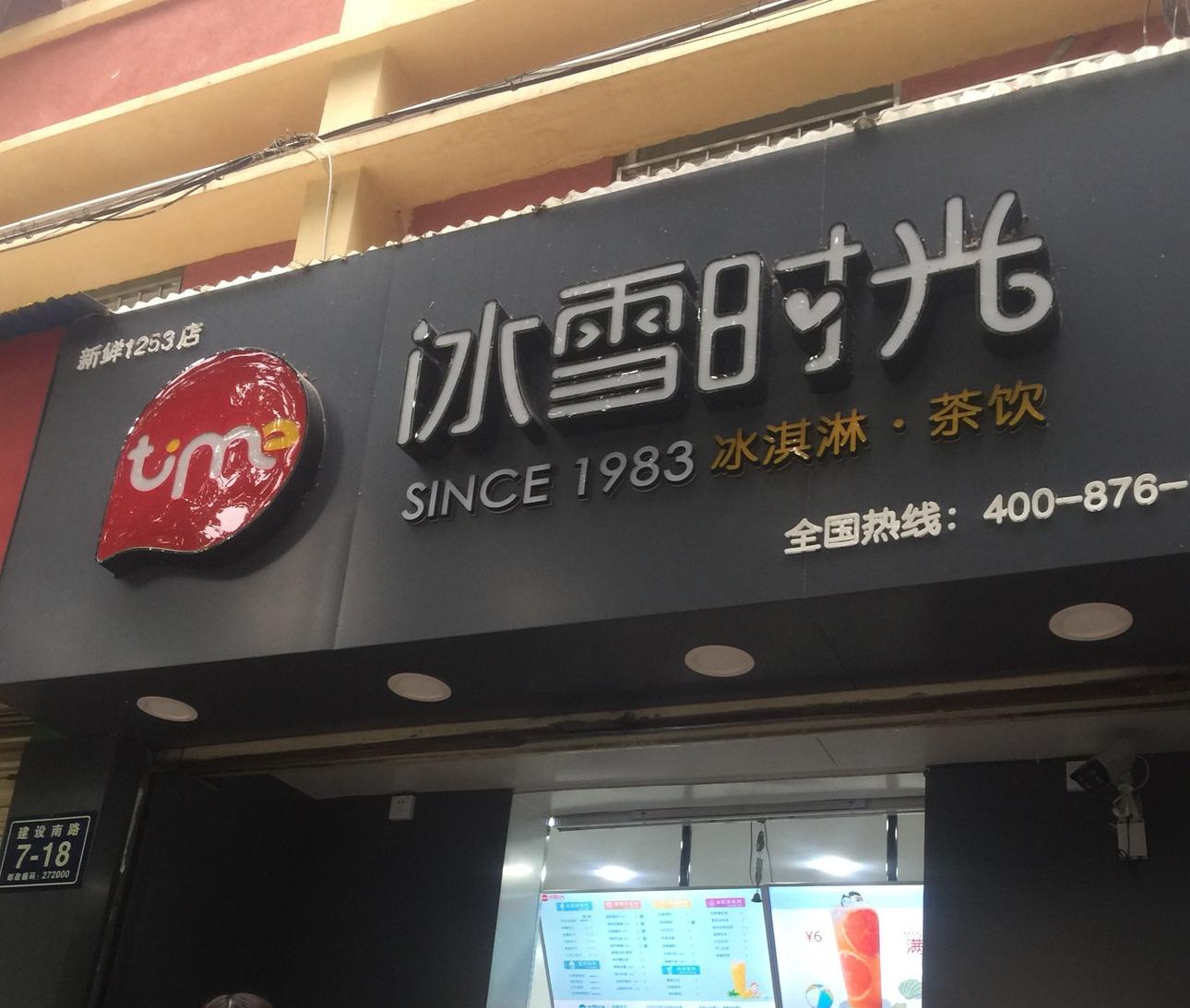}&
\includegraphics[height=2cm, width= 6cm]{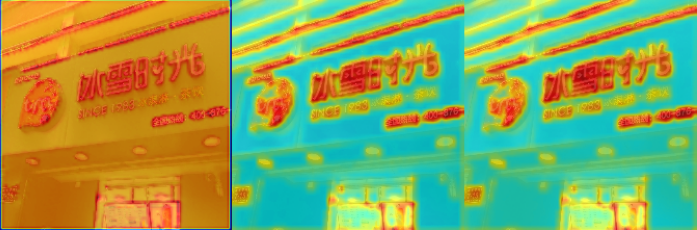}\\
\end{tabular}
\end{adjustbox}
\caption{Some illustration of our method on different datasets and their feature maps from the last three layers of backbone respectively. Zoom in for better visualization. }
\label{fig:figvis}
\end{figure*}

\vspace{-1mm}
\noindent
Table~\ref{tab:my-tablesota} shows the results summary of Swin-TESTR when compared to other related text spotting approaches. We outperform the existing state-of-the-art Abinet++~\cite{fang2022abinet++} on both word-level Total-Text~\cite{ch2020total} and sentence-level CTW1500~\cite{liu2019curved} benchmarks in end-to-end recognition task. On the other hand for scene text detection, we outperform SwinTextSpotter~\cite{huang2022swintextspotter} on CTW1500 while having the second-best results for Total-Text in terms of F-score. It is worth mentioning that our model performs the best in the recall metrics for both benchmarks. Qualitative example studies have been further shown in Fig.~\ref{fig:figvis}.\\ 
\noindent
\textbf{Regular-shaped Scene Text Spotting.} We evaluate our method on ICDAR15 benchmark~\cite{karatzas2015icdar} and display our results compared with the state-of-the-art in Table~\ref{tab:my-tablesota}. For Text detection, we achieve a 5\% gain in precision over TESTR~\cite{zhang2022text} while beating the state-of-the-art F-measure marginally. For the text spotting task, our approach gives the \textit{best result in the  most challenging ``Strong'' type} as in this setting every image contains a lexicon of only 100 words. We outperformed TESTR, Swintspotter, and Abinet++ by almost 1.5\%, 2.5\%, and 0.5\% respectively. In Fig~\ref{fig:figvis} we show how our method performs on this dataset  in the third column.
\\
\noindent
\textbf{Low-Resource Text Spotting.} We have also evaluated Vintext~\cite{nguyen2021dictionary}, and ReCTS~\cite{zhang2019icdar} a low-resource scene text detection benchmark for Vietnamese and Chinese text to show the high generalizability of our model. We can see the results in Table~\ref{tab:my-tablevintext} showing that our methods outperform the state-of-the-art by almost 2\%. In Fig~\ref{fig:figvis} we show the performance of our model in Vintext and ReCTS in the last column \textcolor{black}{(\textbf{NOTE:} due to the space limitations we shift the Table of the ReCTS in the supplementary materials (Table 3). It has been observed that Swin-TESTR outperforms the SOTA approaches with ChineseSynthTexT pre-training).} 
\\
We further throw light on the following insights after some intuitive analysis:
\\
\noindent
~\textbf{Why ABINet++ has better performance in End-to-End None performance?} Since ABINet++~\cite{fang2022abinet++} uses linguistic information for the text spotting framework, they highly improve in this metric. A similar method MANGO~\cite{qiao2021mango} also follows this trend. Our approach outperforms both these approaches with a purely visual approach.\\

\subsection{Ablation Studies}
\noindent
To understand the significance of the different components of the Swin-TESTR framework, we conduct ablation studies and deduce the following insights. 
\begin{table}[ht]
\vspace{-1mm}
\centering
\caption{Ablation for different feature extraction backbones.}
\begin{tabular}{@{}ccccc@{}}
\toprule
\multirow{2}{*}{Methods} & \multicolumn{3}{c}{Detection}& End-to-end\\ \cmidrule(lr){2-4} \cmidrule(lr){5-5} 
& P & R & F &  None \\ \hline
Resnet-50 \cite{he2016deep} & 88.87& 76.47& 82.20&  60.06  \\
ViT-T \cite{dosovitskiy2020image}& 90.17& 72.90  & 80.62  &59.40\\ 
\textbf{Swin-T} \cite{liu2021swin} & \textbf{93.59} &\textbf{75.20} &\textbf{83.40} & \textbf{67.06} \\ 
\bottomrule
\end{tabular}
\label{tab:table1}
\vspace{-4mm}
\end{table}
\paragraph{Swin Transformer is a strong visual backbone for text spotting.} The Swin-Tiny~\cite{liu2021swin} feature extraction backbone in the Swin-TESTR framework gives a significant performance gain over ViT-Tiny~\cite{dosovitskiy2020image} and Resnet-50~\cite{he2016deep} backbones in both detection and recognition metrics.  Almost a 4\% change in detection precision and a 1\% change in the detection F-measure over Resnet-50 is observed. We get almost a massive 7\% improvement in the End-to-End recognition results when it is run on the Total-Text~\cite{ch2020total} dataset as shown in Table~\ref{tab:table1}. This justifies that the window-based local attention computed with Swin-T can be really effective for special text regions with smaller local boundaries.
\begin{figure}[ht]
\vspace{-2mm}
\centering
\begin{tabular}{c c  c } 
  \includegraphics[height=2cm, width= 2cm]{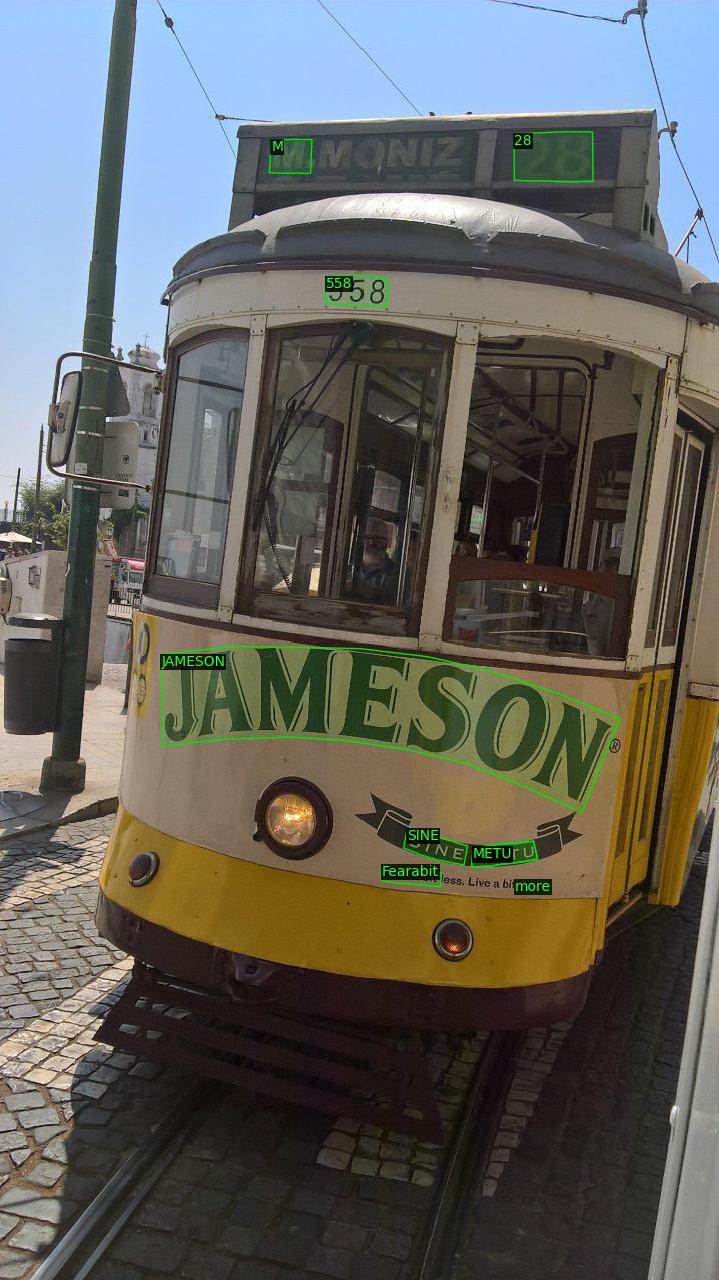}&
  \includegraphics[height=2cm, width= 2cm]{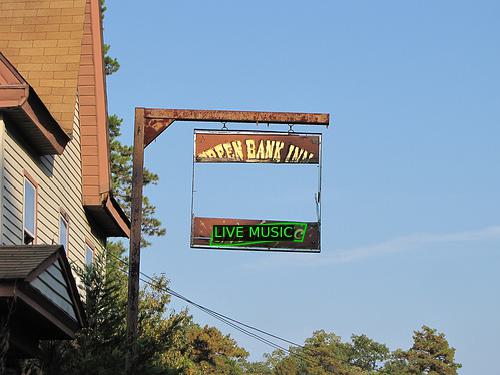}&
  \includegraphics[height=2cm, width= 2cm]{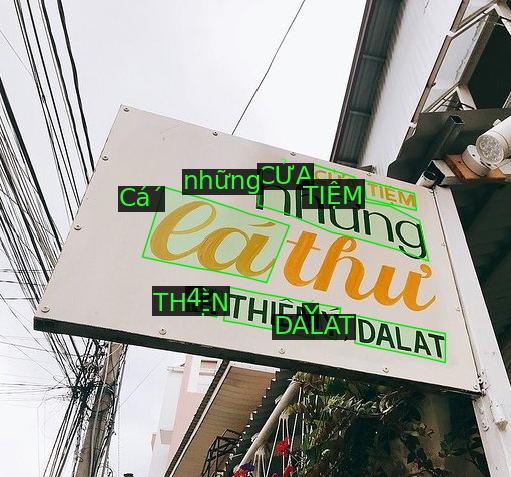}\\
\end{tabular}
\caption{Some failure cases of our method on Swin-TESTR. Zoom in for better visualization.}
\label{fig:fig9}
\vspace{-2mm}
\end{figure}

\paragraph{Semi-Supervised Training Strategy.}
To assess the data dependency of Swin-TESTR, we conducted a series of experiments involving different subsets of the dataset. These subsets consisted of 25\%, 50\%, 75\%, and the entire dataset, respectively. The objective was to investigate how Swin-TESTR's performance varied with the amount of training data available. Our findings provided compelling evidence of a strong data dependency exhibited by Swin-TESTR. Notably, as we increased the proportion of data used for training, the model's performance consistently improved. Training with 25\% and 50\% of the dataset resulted in noticeable enhancements, indicating the reliance on available data. 

\begin{table}[ht]
\centering
\caption{Performance Evaluation with different \% of labeled data}
\begin{tabular}{@{}ccccc@{}}
\toprule
\multirow{2}{*}{\% Labels } & \multicolumn{3}{c}{Detection}& End-to-end\\ \cmidrule(lr){2-4} \cmidrule(lr){5-5} 
& P & R & F &  None \\ \hline
25\%  & 89.88& 81.8& 85.65&  65.71  \\
50\% & 90.56& 82.75  & 86.48  &70.38\\ 
75\%  & 89.61 &85.73 &87.63  & 71.17\\
All  & 90.58&\textbf{85.46}&\textbf{87.95}&\textbf{74.13}\\ 
\bottomrule
\end{tabular}
\label{tab:tablesemi}
\end{table}

%% file: tex/conclusion.tex
\section{Conclusion and Future Work}
\textcolor{black}{In this work, we have explored a new direction towards benchmarking of text-spotting models in domain adaptation settings. Comprehensive experiments on different benchmarks exhibit how the proposed baseline achieves competitive performance against the SOTA approaches under different settings. Moreover, diverse multilingual datasets could help the model in getting better intermediate representations during pre-training. This opens further scope for more diverse linguistic and visually-rich datasets to be introduced for aggravating further research.} 


%% file: tex/supply.tex
\section{Some More Insights}

\paragraph{Representation quality and robustness.}
To evaluate the representation quality of the features learned during pre-training and the need for the encoder and dual decoder blocks in the Swin-TESTR pipeline we did an exhaustive evaluation by freezing and unfreezing them. We performed all the possible combinations of frozen and unfrozen decoder and encoder to show the representation quality. Results have been highlighted in Table~\ref{tab:table2} performed on the Total-Text dataset in the fine-tuning configuration. The consistent performances on both detection and recognition tasks for all the different configurations show the model's robustness. 

\begin{table}[ht]
\caption{\textbf{Evaluating Representations.} Performance obtained by fine-tuning a pre-trained model using frozen backbones. Here ``IE'', ``LD'', and ``CD'' represent Image Encoder, Text Location Decoder, and Text Recognition Decoder respectively also ``None'' refers to no lexicon has been used. }
\begin{adjustbox}{max width=.48\textwidth}
\begin{tabular}{ccccccc}
\toprule
\multicolumn{3}{c}{Blocks}&\multicolumn{3}{c}{Detection}&End-to-End  \\ \cmidrule(lr){1-3} \cmidrule(lr){4-6} \cmidrule(lr){7-7}
IE & LD & CD&  P &R & F &None\\\hline
\multicolumn{1}{c}{\color{red}\xmark}&\multicolumn{1}{c}{\color{red}\xmark}& \multicolumn{1}{c}{\color{teal}\cmark}&91.67&77.63 &84.07&71.88\\
\multicolumn{1}{c}{\color{red}\xmark}&\multicolumn{1}{c}{\color{teal}\cmark}& \multicolumn{1}{c}{\color{red}\xmark}&\textbf{93.33}&81.91&87.25&73.74\\

\multicolumn{1}{c}{\color{red}\xmark}&\multicolumn{1}{c}{\color{teal}\cmark}& \multicolumn{1}{c}{\color{teal}\cmark}&90.86&84.64&87.64&68.9\\

\multicolumn{1}{c}{\color{teal}\cmark}&\multicolumn{1}{c}{\color{red}\xmark}&\multicolumn{1}{c}{\color{red}\xmark}&91.89&83.51&87.50&74.27\\
\multicolumn{1}{c}{\color{teal}\cmark}&\multicolumn{1}{c}{\color{red}\xmark}& \multicolumn{1}{c}{\color{teal}\cmark}&92.52&83.12&87.57&74.56\\
\multicolumn{1}{c}{\color{teal}\cmark}&\multicolumn{1}{c}{\color{teal}\cmark}&\multicolumn{1}{c}{\color{red}\xmark}&91.23&84.69&87.84&\textbf{75.14}\\
\multicolumn{1}{c}{\color{teal}\cmark}& \multicolumn{1}{c}{\color{teal}\cmark}& \multicolumn{1}{c}{\color{teal}\cmark}&90.58&\textbf{85.46}&\textbf{87.95}&74.13\\
\bottomrule
\end{tabular}
\end{adjustbox}

\label{tab:table2}
\end{table}

\begin{figure}[ht]
\includegraphics[width=0.48\textwidth]{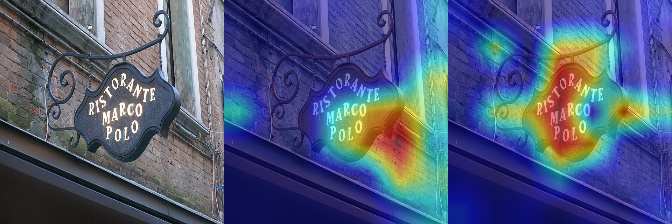}
\caption{\textbf{Illustration of the effectiveness of the Swin Backbone over Resnet 50.} first one is the original image, the second one is the feature map extracted by Resnet 50. and the last one is the feature map generated by Swin-tiny Backbone } \label{fig3}
\end{figure}

\paragraph{Effectiveness of Swin Transformer backbone over ResNet 50.} 
Connecting remote features with vanilla convolutions can be a challenging task due to their local operation at fixed sizes, such as $3 \times 3$. However, in the case of text spotting, it becomes crucial to capture the relationships between different texts. This is because scene text within the same image tends to exhibit similarities in terms of the text background, style, and texture. To address this issue, we have selected a backbone architecture known as Swin-Transformer~\cite{liu2021swin}, specifically the Swin-tiny variant, for feature extraction. The Swin-Transformer unit stands out as a small and efficient option that fits our requirements. By incorporating the Swin-tiny unit into our framework, we aim to enhance the ability of our model to understand the contextual connections between texts, thereby improving text spotting performance.
In summary, the challenge lies in connecting remote features using conventional convolutions, which have limited receptive fields. However, in the context of text spotting, it is crucial to capture the relationships between different texts. To tackle this, we utilize the Swin-tiny backbone, which is a compact and efficient variant of the Swin-Transformer, to extract features that enable our model better to comprehend the contextual information present in text scenes. we have illustrated how our Swin-tiny backbone manages to generate a better-localized representation over the text than the standard Resnet-50 model. The final output from the last layer of the encoder is then propagated to the next phase. Fig.~\ref{fig3} illustrates how we incorporate two dilated convolution layers, one vanilla convolution layer, and one residual structure into the original Swin-Transformer, which also introduces CNN properties to Transformer. 

\section{Datasets and Experimental Setup}

\noindent
\textit{ICDAR 2015}~\cite{karatzas2015icdar} is the official dataset for ICDAR 2015 robust reading competition for regular text spotting. It has 1000 training and 500 testing images. As discussed in~\cite{singh2021textocr} it has 6.9 words per image.\\
\textit{Total-Text}~\cite{ch2020total} is a well-known arbitrary-shaped text spotting benchmark with 1255 training and 300 testing images. The text instances in this benchmark are at the word level. As discussed in~\cite{singh2021textocr} it has 7.4 words per image. \\ 
\textit{CTW1500}~\cite{liu2019curved} is another arbitrary-shaped text spotting benchmark consisting of 1000 training and 500 testing images. The text instances in this benchmark are at the multi-word level. \\
\textit{VinText}~\cite{nguyen2021dictionary} is a recently released Vietnamese text dataset containing arbitrary-shaped text. It consists of 1200 training images and 500 testing images. \\
\textit{Curved SynthText 150K} dataset synthesized in~\cite{liu2020abcnet}, consisting of 94723 images with mostly straight text and 54327 images with major curved text.  \\
\textit{ICDAR 2017 MLT}~\cite{nayef2017icdar2017} is a multi-lingual text dataset. It contains 7200 training images and 1800 validation images. We only select the images containing English texts for training. The text instances in this benchmark are at the word level. As discussed in~\cite{singh2021textocr} it has 9.5 words per image which is very high.\\
\textit{ReCTS}~\cite{zhang2019icdar} is a Chinese arbitrary-shaped text dataset with 20000 training images and 5000 testing images.\\

\noindent
\textbf{Experimental Setup:} One RTX 3080 Ti GPU with a batch size of 1 for an overall 16 days for pre-training phase. For supervised training on the corresponding evaluation datasets to understand where we stand in the competition, we fine-tune Total-Text, and ICDAR2015 with 200K iterations. For the CTW1500 dataset, we fine-tune Swin-TESTR over 2000K iterations with a maximum length of text 100. As it is annotated sentence level it needs more iterations to fine-tune. For Vintext we fine-tune our model with 1000K iterations.

\section{More Performance Evaluation and Analysis}

\begin{table*}[t]
\centering
\caption{Performance Analysis on Out-of-Vocabulary End-to-End Recognition. Results style: \textbf{best}, \underline{second best}}
\label{Tab4}
\begin{tabular}{|l|c|c|c|}
\hline
Method& Recall & Precision &  Hmean  \\ \hline
Synth-Text& 0.0011    & 0.0033 & 0.0021 \\ \hline
Synth-Text $\rightarrow$ ICDAR MLT17& \underline{0.1146}    & \underline{0.2903} & \underline{0.1644} \\ \hline
Synth-Text $\rightarrow$ ICDAR15& 0.0217    & 0.1088 & 0.0361 \\ \hline
Synth-Text $\rightarrow$ ICDAR MLT17 $\rightarrow$ ICDAR15& 0.0563    & 0.2076 & 0.0886 \\ \hline
Synth-Text $\rightarrow$ ICDAR MLT17 $\rightarrow$ Total-Text& 0.0446    & 0.1638 & 0.0701 \\ \hline
Synth-Text $\rightarrow$ ICDAR MLT17 $\rightarrow$ CTW1500& 0.0399    & 0.2518 & 0.0688 \\ \hline
Synth-Text $\rightarrow$ Total-Text& 0.0445    & 0.1779 & 0.0711 \\ \hline
Synth-Text $\rightarrow$ CTW1500& 0.0375    & 0.1510 & 0.0601 \\ \hline
Mix pre-train& 0.0865    & 0.4674 & 0.1459 \\ \hline
ICDAR MLT17& 0.0364    & 0.1309 & 0.0433 \\ \hline
ICDAR MLT17 $\rightarrow$ Synth-Text& 0.0439    & 0.1712 & 0.0699 \\ \hline
Synth-Text $\rightarrow$ ReCTS & 0.0848& 0.2473 & 0.1263 \\ \hline
Mix pre-train $\rightarrow$ Fine-tune& \textbf{0.2492}    & \textbf{0.2724} & \textbf{0.2603} \\ \hline
\end{tabular}
\end{table*}

\paragraph{Performance Analysis on the Out-of-Vocabulary dataset.}

\begin{table}[ht]
\caption{\textbf{Text spotting results on ReCTS}. Results style: \textbf{best}, \underline{second best}}
\begin{adjustbox}{max width=.48\textwidth}
\begin{tabular}{@{}ccccc@{}}
\toprule
\multicolumn{1}{c}{\multirow{2}{*}{Methods}}& \multicolumn{3}{c}{Detection}& \multirow{2}{*}{1-NED} \\ \cmidrule(lr){2-4} 
 & P & R& F& \\ \hline
FOTS~\cite{liu2018fots}&78.3&82.5 & 80.31 &50.8\\
Mask TextSpotter\cite{lyu2018mask}&89.3 & \underline{88.8} & 89.0& 67.8\\
AE TextSpotter~\cite{wang2020ae}& 92.6 &\textbf{91.0}& \textbf{91.8}& \underline{71.8}\\
ABCNet v2\cite{liu2021abcnet}&\underline{93.6} &87.5 & \underline{90.4}& 62.7 \\
Swintextspotter\cite{huang2022swintextspotter}& \textbf{94.1} &87.1 & \underline{90.4} &\textbf{72.5} \\ \hline
\textbf{Swin-TESTR}& 92.61&67.72 &78.21&68.12\\
\bottomrule
\end{tabular}
\end{adjustbox}

\label{tab:my-tablerects}
\end{table}

Here we have also evaluated the domain adaptation setting in the Out-of-Vocabulary (OOV) text recognition dataset~\cite{garcia2022out}. An OOV dataset in the context of natural language processing and machine learning typically refers to \textit{words that are not in the model's training dictionary} or vocabulary. This situation often arises when the model is used in real-world applications, where it encounters words it has not seen during training. As shown in Table~\ref{Tab4}, results actually prove again how the MLT17 dataset is beneficial for the domain adaptive pretraining (Synth to Real), justifying our hypothesis that \textit{generally larger and more diverse training sets will result in better OOV performance for text recognition}.

%

\paragraph{Performance Analysis on the ReCTS dataset}

In the case of Chinese, where there are thousands of characters and many more possible words, this can be especially important.
. The results as shown in Table~\ref{tab:my-tablerects} demonstrate that our model doesn't perform the best compared to the other baselines. We hypothesize it's mainly due to the pretraining with the English Curved SynthText~\cite{liu2020abcnet}. The performances have the potential to match with the existing SOTA if pre-trained on a more domain-specific Chinese Synthetic dataset. There is more scope for exploration in future works. 



\begin{figure*}[b]
\centering
\subfloat[Industrial OCR 1]{\includegraphics[width=0.25\textwidth]{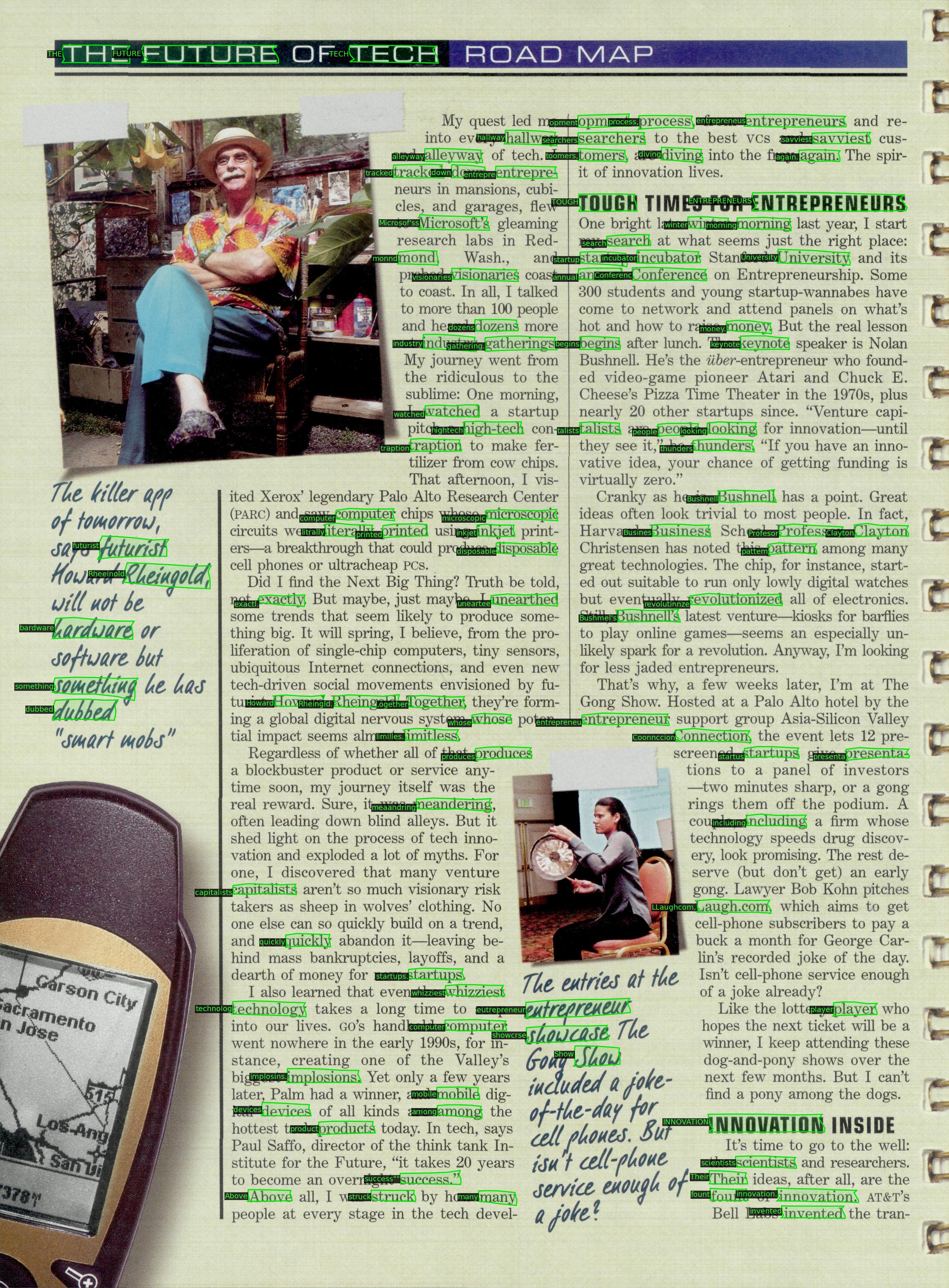}}\hfill
\subfloat[Industrial OCR 2]{\includegraphics[width=0.25\textwidth]{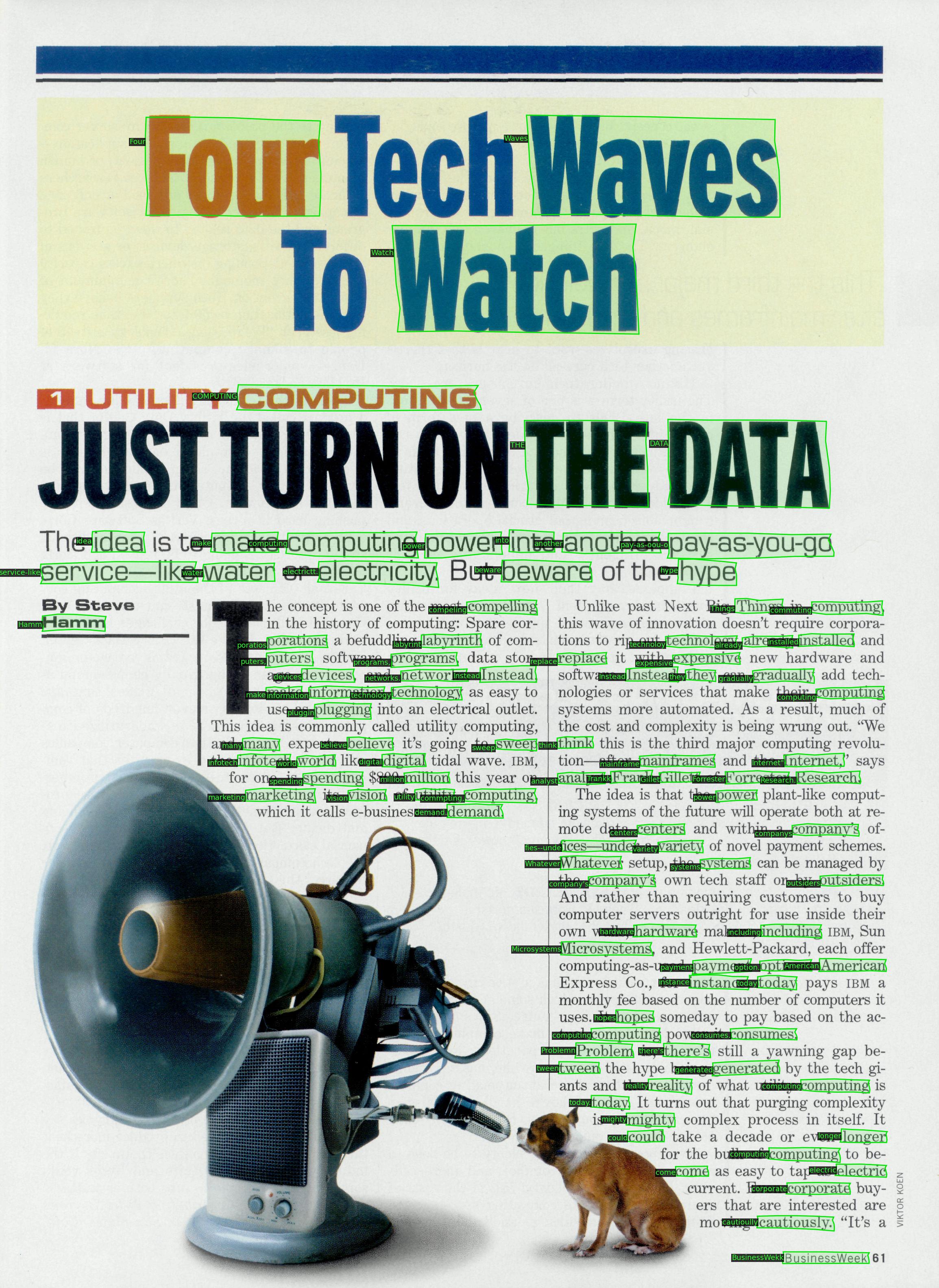}}\hfill
\subfloat[Swin-TESTR 1]{\includegraphics[width=0.25\textwidth]{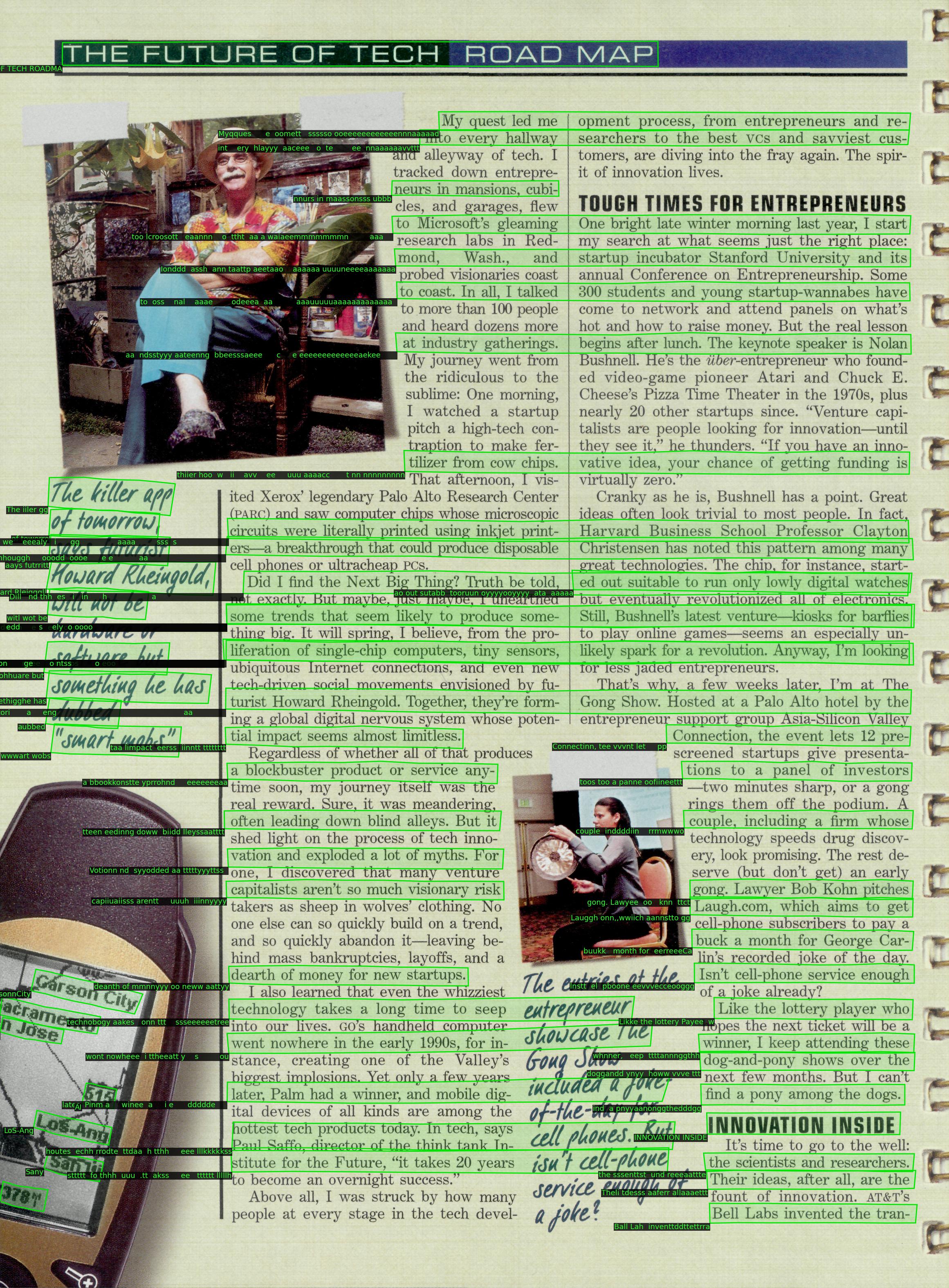}}\hfill
\subfloat[Swin-TESTR 2]{\includegraphics[width=0.25\textwidth]{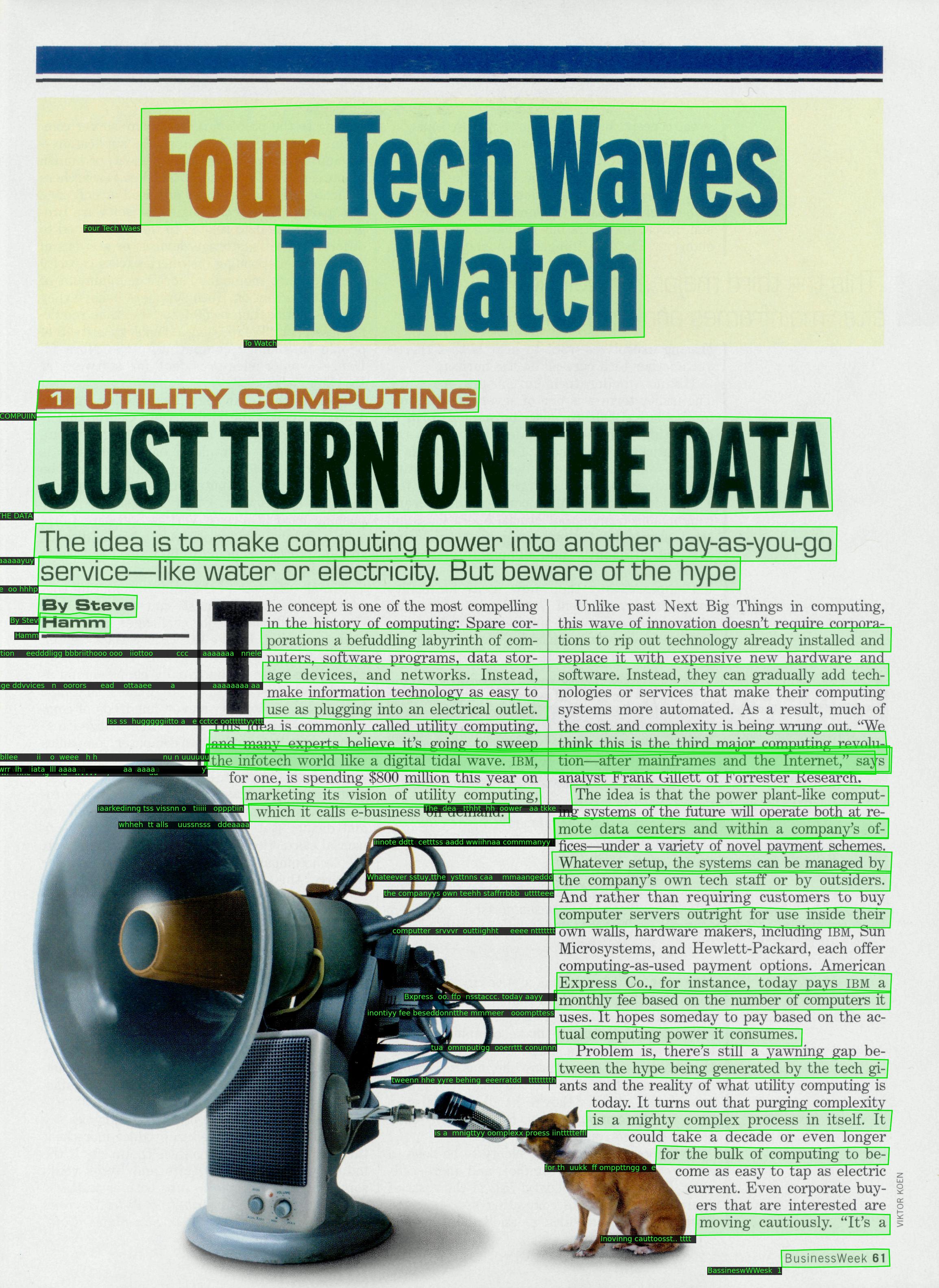}}
\caption{Comparative Analysis of Swin-TESTR as OCR for document layout analysis.}
\label{fig:q}
\end{figure*}

\section{Qualitative Analysis on Document Layout Analysis}
In Figure \ref{fig:q}, we present a comprehensive assessment of the reading proficiency achieved by our Swin-TESTR model in the context of document layout analysis. This evaluation revolves around the model's ability to detect various layout regions within documents by leveraging the OCR capabilities embedded in our framework.

Our performance evaluation reveals that Swin-TESTR stands as a robust contender, yielding results that closely rival those obtained through the original document layout analysis OCR. Notably, it surpasses the performance of the original model when it comes to identifying text, mathematical content, and table regions. This achievement underscores Swin-TESTR's prowess in extracting valuable information from documents.

However, it is essential to acknowledge that Swin-TESTR exhibits relatively poorer performance in the identification of separator regions, miscellaneous elements, and image-containing sections. This limitation can be attributed to the model's initialization with text spotting weights, which essentially means it was trained predominantly on text-centric data and, consequently, has limited exposure to instances of these other region types.

\begin{figure*}[ht]
\includegraphics[width=\textwidth]{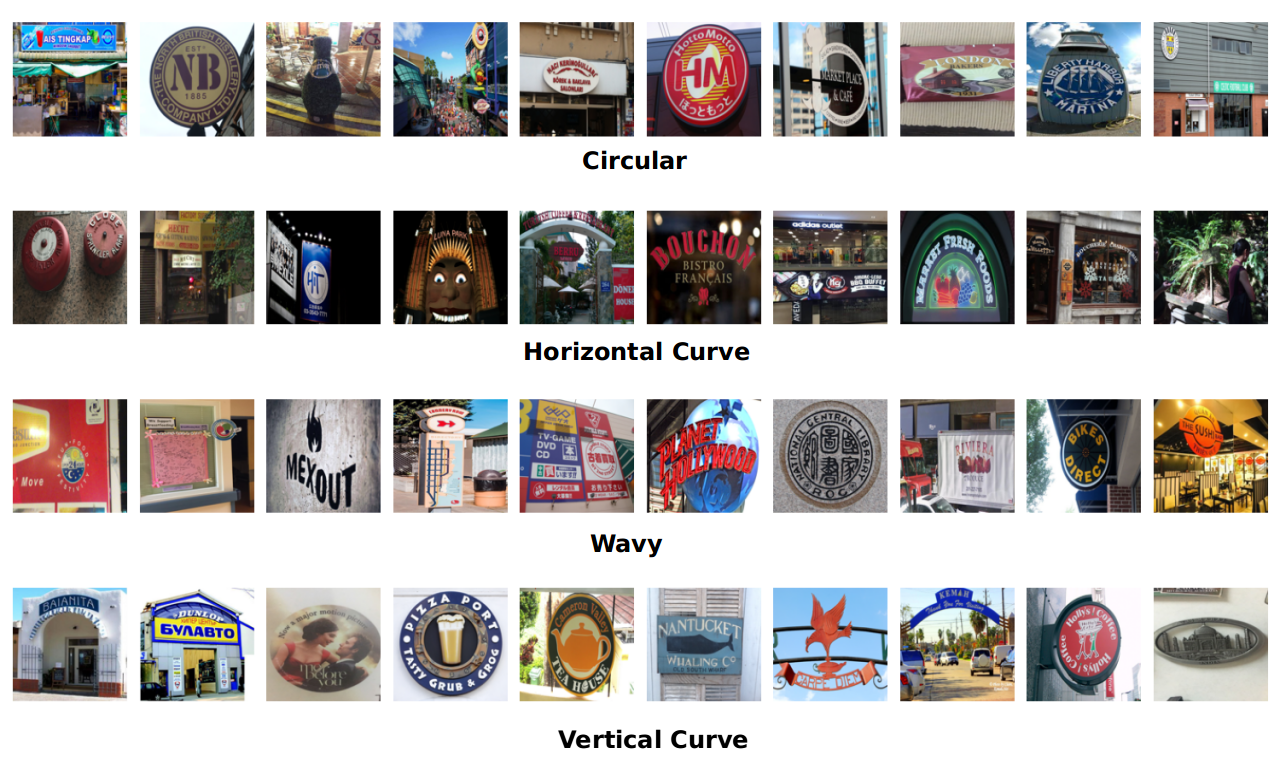}
\caption{Clusterization of Total-Text through the shape of the text.} \label{fig4}

\includegraphics[width=\textwidth]{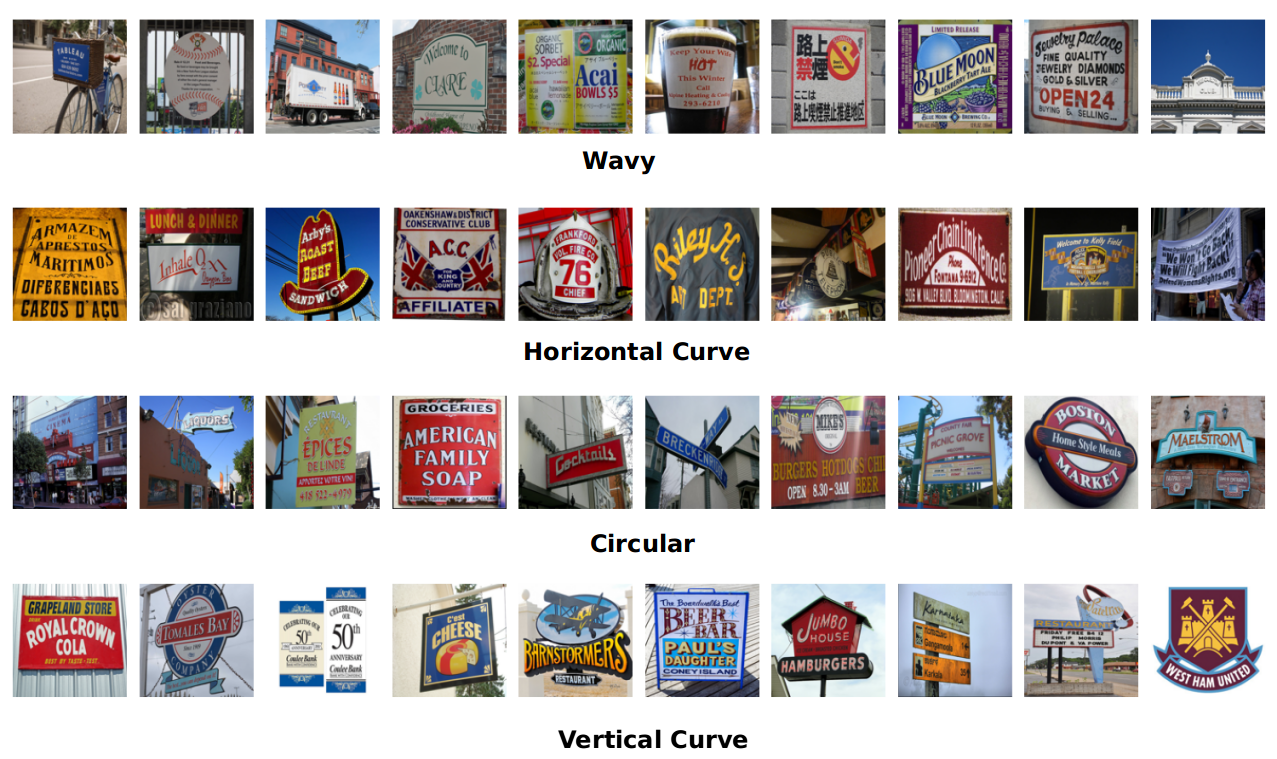}
\caption{Clusterization of CTW1500 through the shape of the text.} \label{fig5}
\end{figure*}

\begin{figure*}[ht]
\centering
\begin{tabular}{c c c c} 
\includegraphics[height=4cm, width= 4cm]{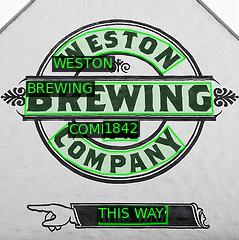}&
\includegraphics[height=4cm, width= 4cm]{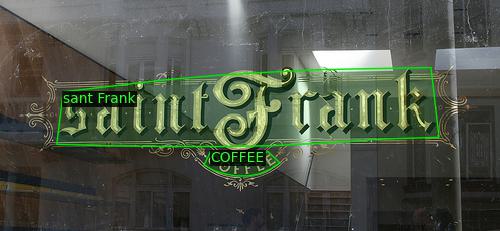}&
\includegraphics[height=4cm, width= 4cm]{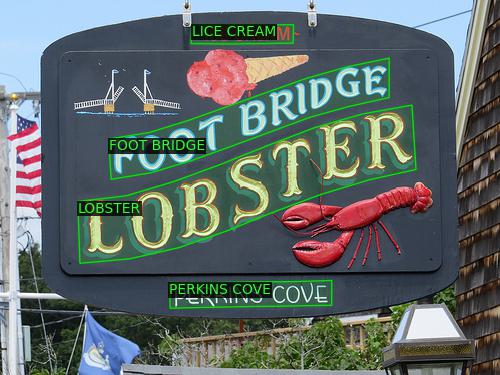}&
\includegraphics[height=4cm, width= 4cm]{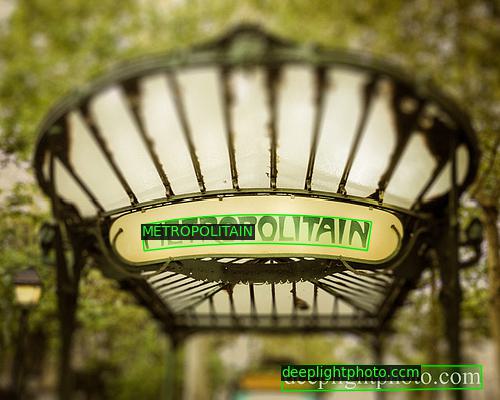}\\
(a)&(b)&(c)&(d)\\
\includegraphics[height=4cm, width= 4cm]{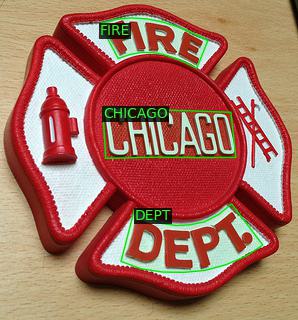}&
\includegraphics[height=4cm, width= 4cm]{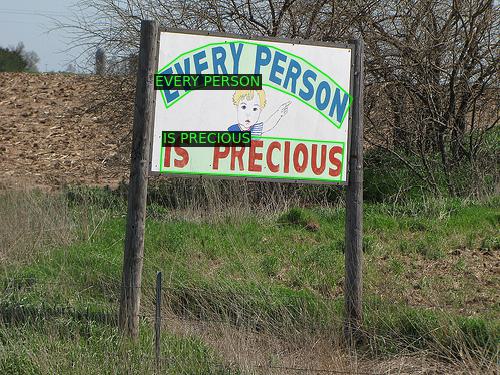}&
\includegraphics[height=4cm, width= 4cm]{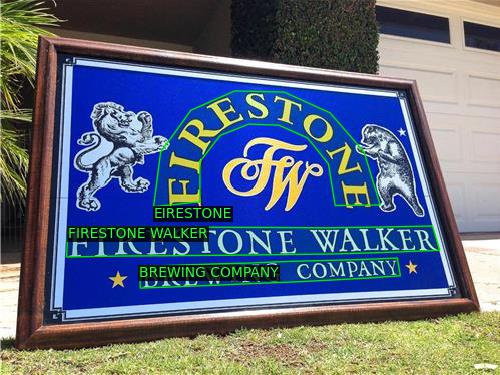}&
\includegraphics[height=4cm, width= 4cm]{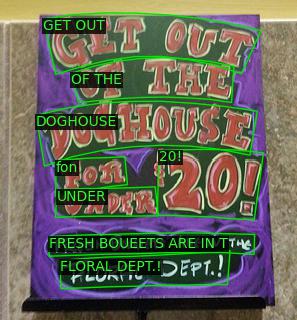}\\
(e)&(f)&(g)&(h)\\
\includegraphics[height=4cm, width= 4cm]{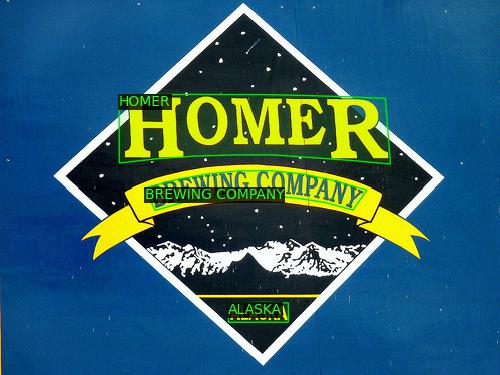}&
\includegraphics[height=4cm, width= 4cm]{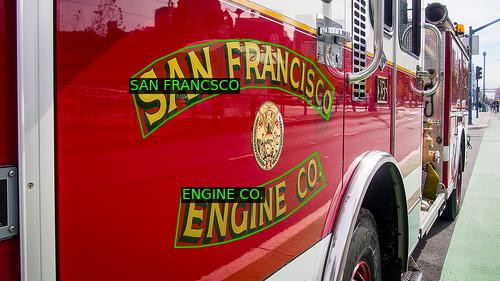}&
\includegraphics[height=4cm, width= 4cm]{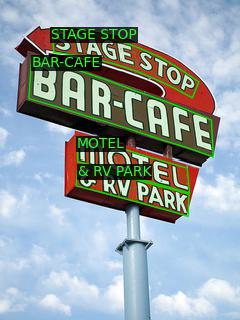}&
\includegraphics[height=4cm, width= 4cm]{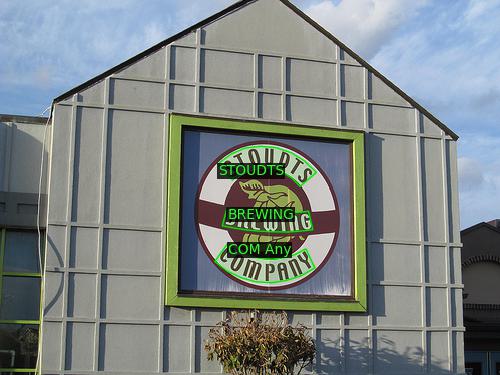}\\
(i)&(j)&(k)&(l)\\
\end{tabular}
\caption{\textbf{Performance Analysis on CTW1500.} Some qualitative test cases of our method on images from CTW1500.}
\label{fig:figsupctw}
\end{figure*}

\begin{figure*}[ht]
\centering
\begin{tabular}{cccc} 
\includegraphics[height=4cm, width= 4cm]{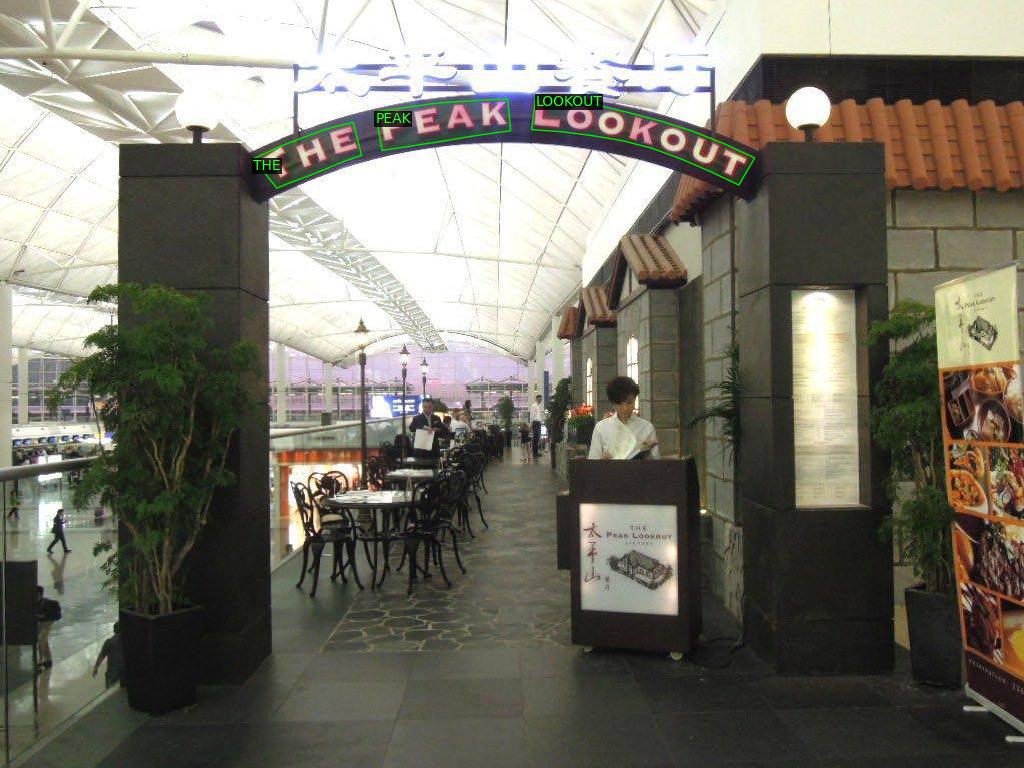}&
\includegraphics[height=4cm, width= 4cm]{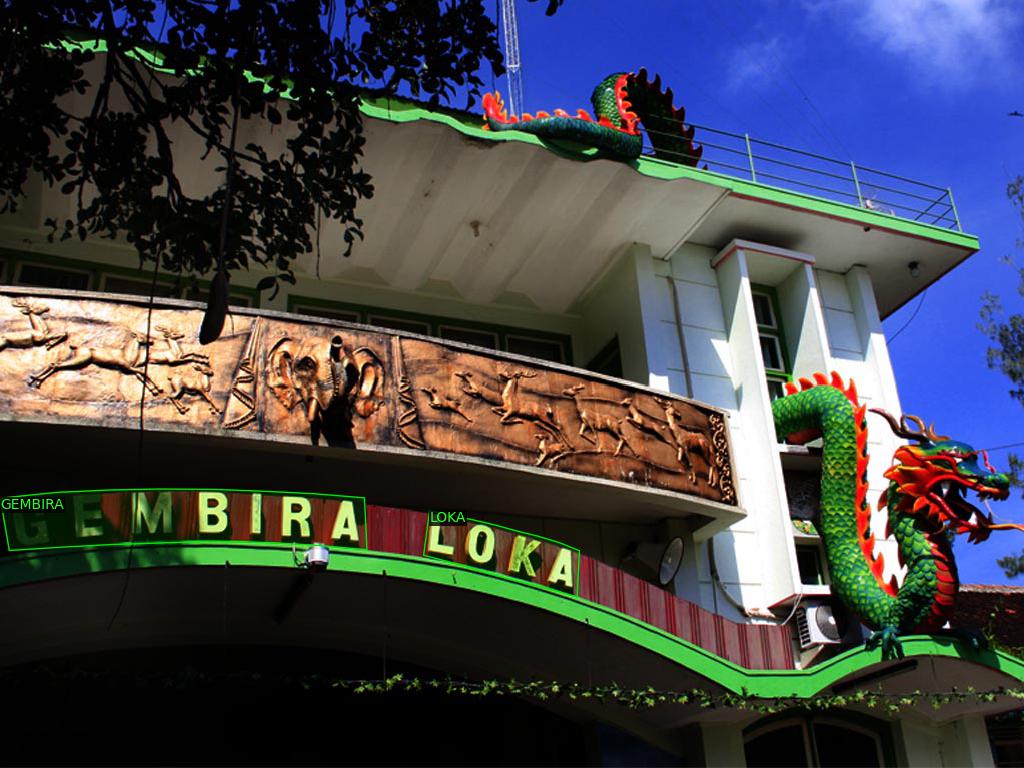}&
\includegraphics[height=4cm, width= 4cm]{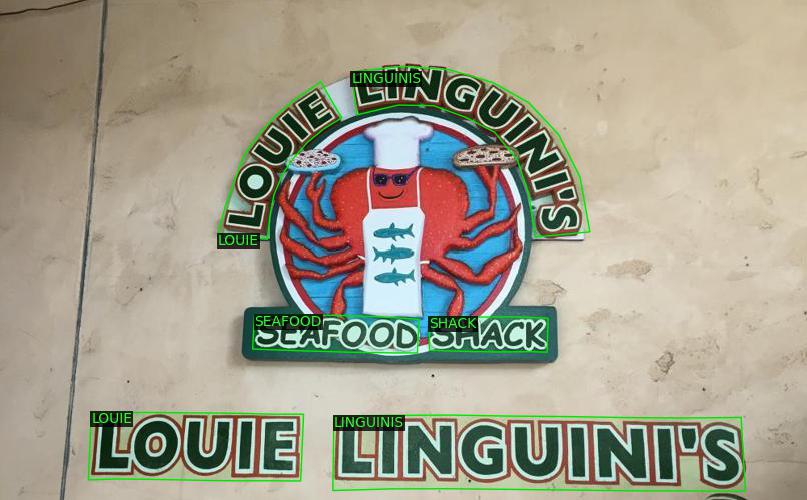}&
\includegraphics[height=4cm, width= 4cm]{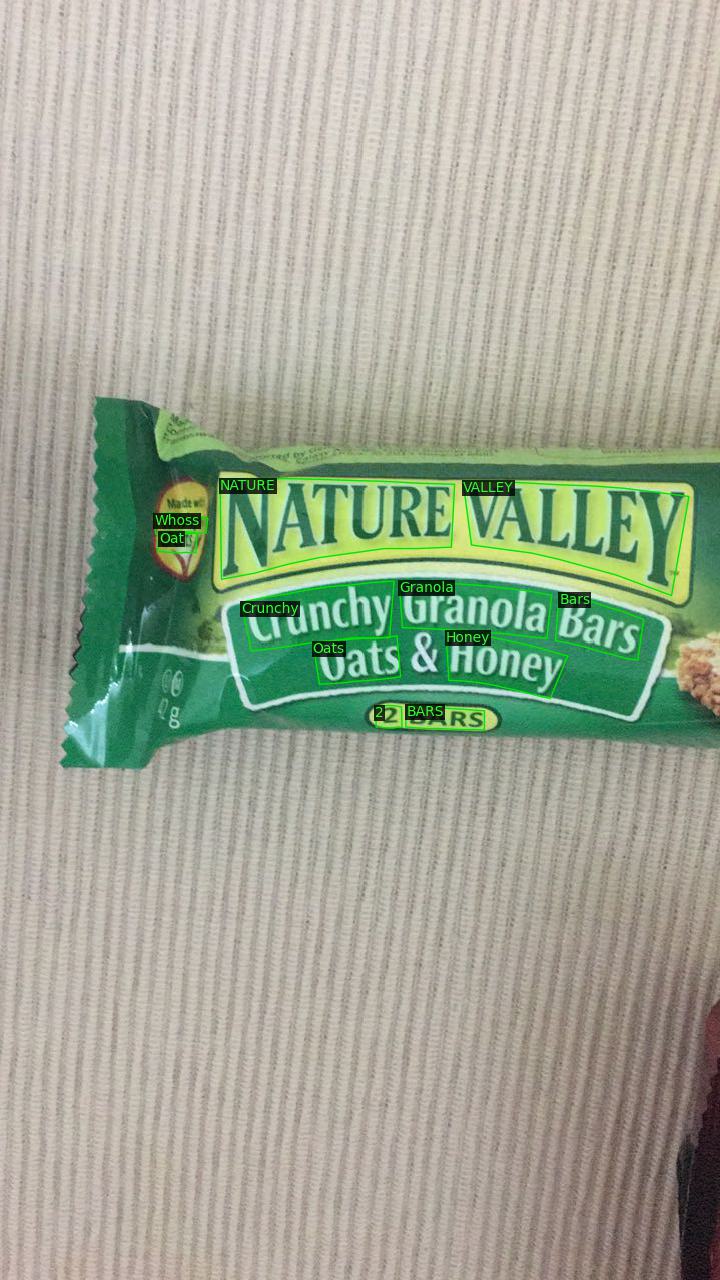}\\
(a)&(b)&(c)&(d)\\
\includegraphics[height=4cm, width= 4cm]{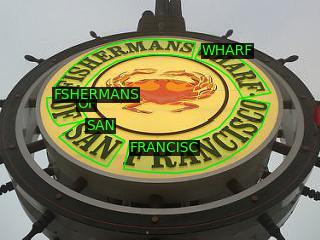}&
\includegraphics[height=4cm, width= 4cm]{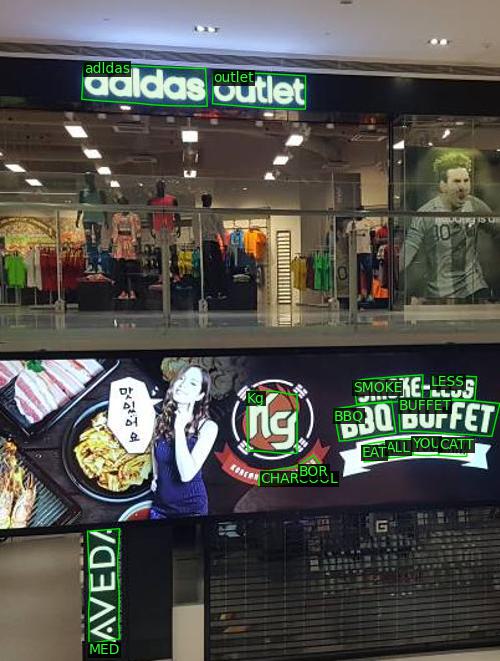}&
\includegraphics[height=4cm, width= 4cm]{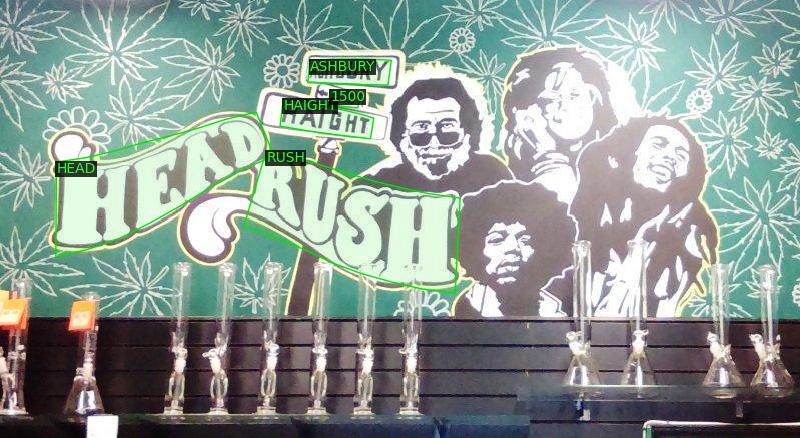}&
\includegraphics[height=4cm, width= 4cm]{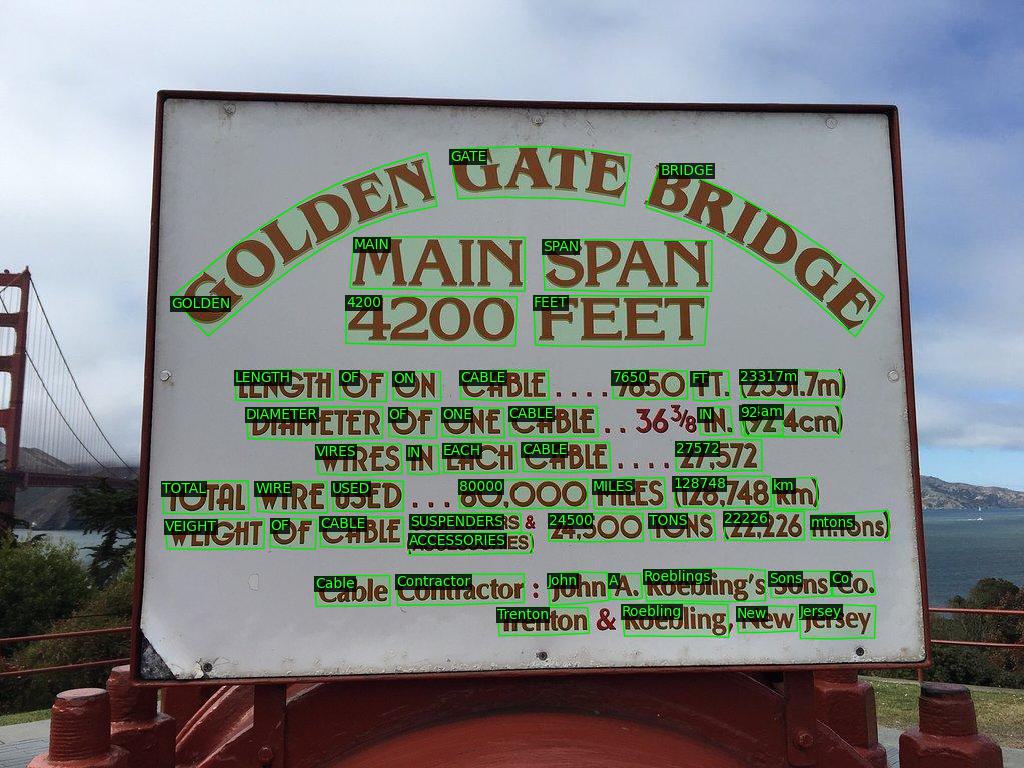}\\
(e)&(f)&(g)&(h)\\
\includegraphics[height=4cm, width= 4cm]{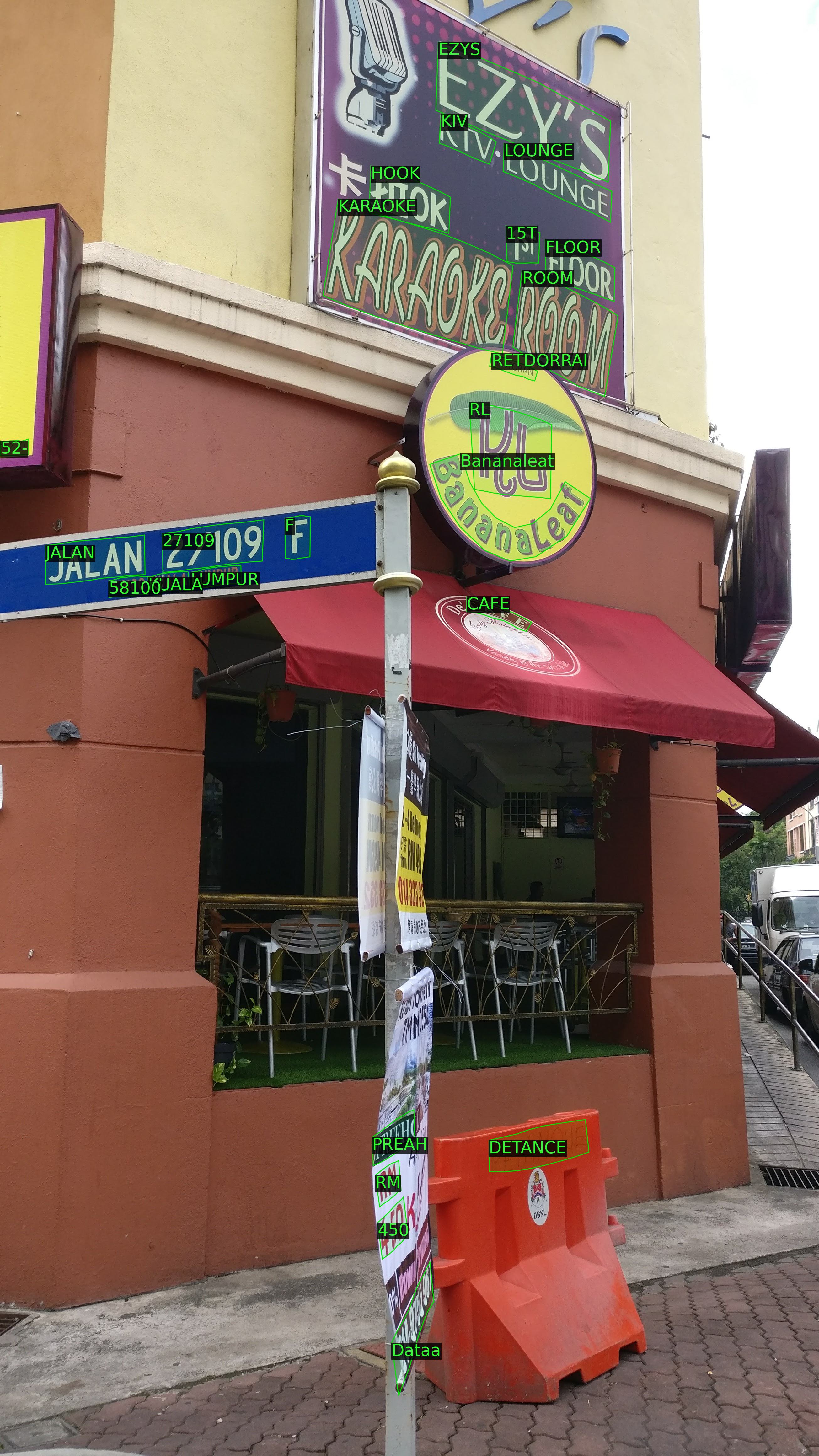}&
\includegraphics[height=4cm, width= 4cm]{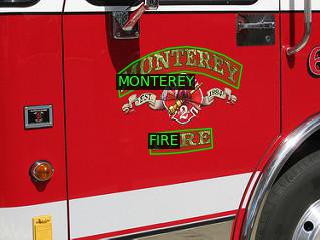}&
\includegraphics[height=4cm, width= 4cm]{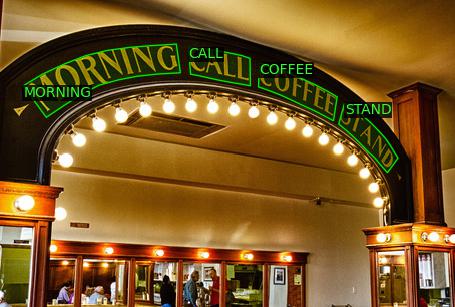}&
\includegraphics[height=4cm, width= 4cm]{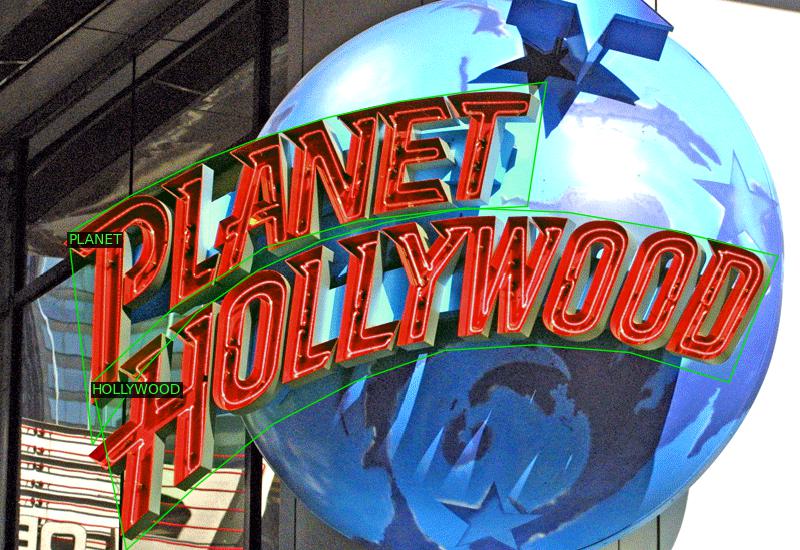}\\
(i)&(j)&(k)&(l)\\
\end{tabular}
\caption{\textbf{Performance Analysis on TotalText.} Some qualitative test cases of our method on images from TotalText.}
\label{fig:figsuptt}
\end{figure*}

\begin{figure*}[ht]
\centering
\begin{tabular}{cccc} 
\includegraphics[height=4cm, width= 4cm]{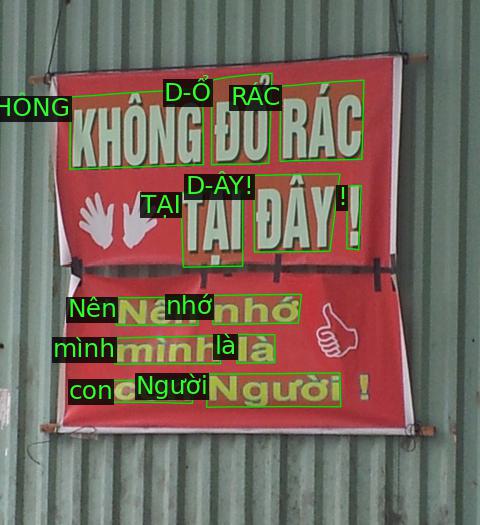}&
\includegraphics[height=4cm, width= 4cm]{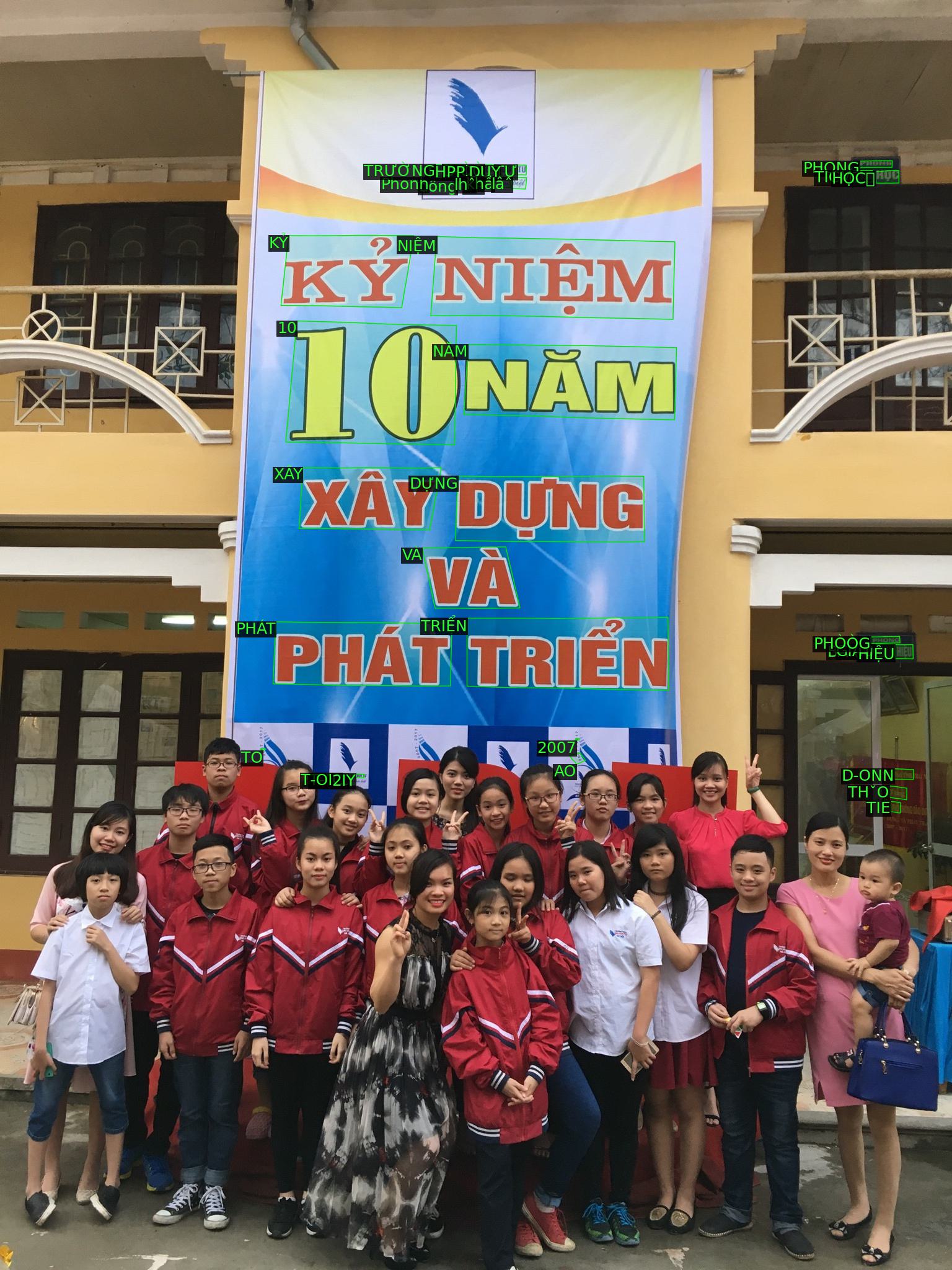}&
\includegraphics[height=4cm, width= 4cm]{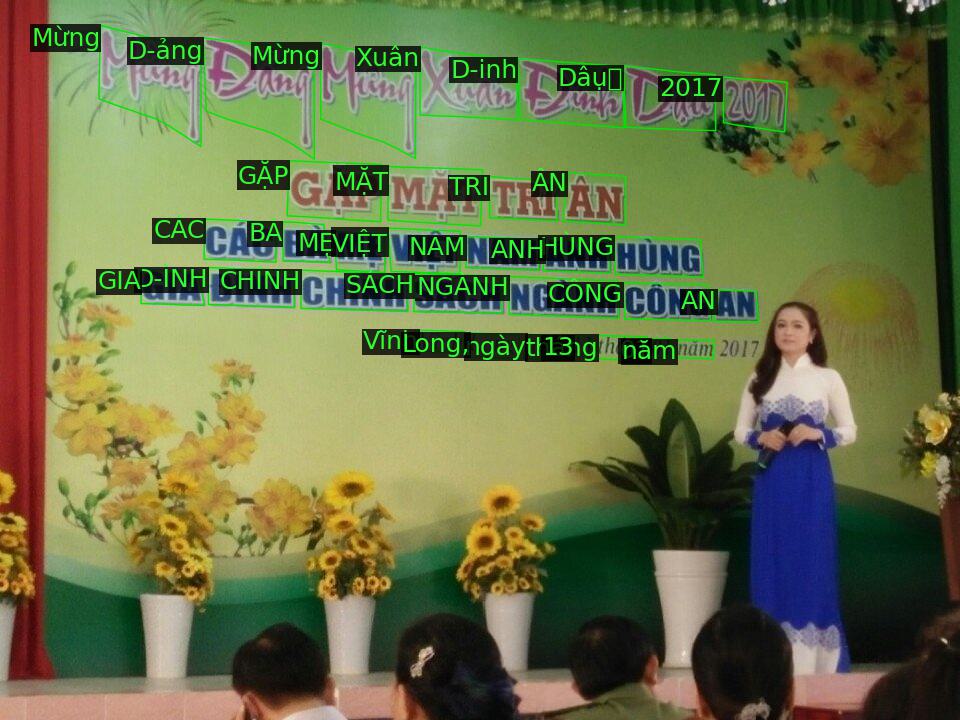}&
\includegraphics[height=4cm, width= 4cm]{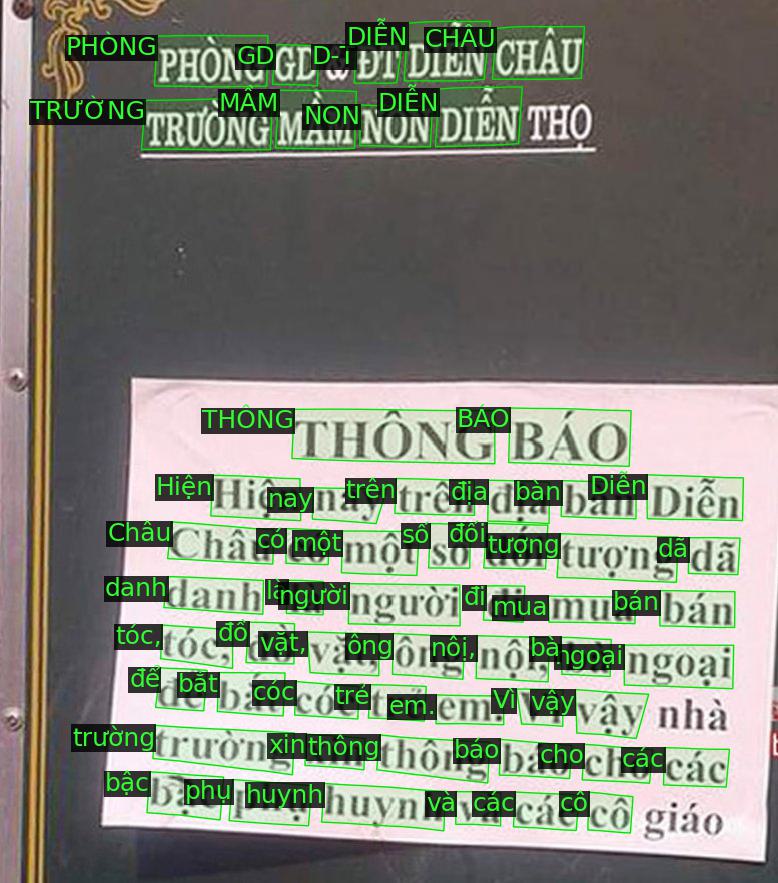}\\
(a)&(b)&(c)&(d)\\
\includegraphics[height=4cm, width= 4cm]{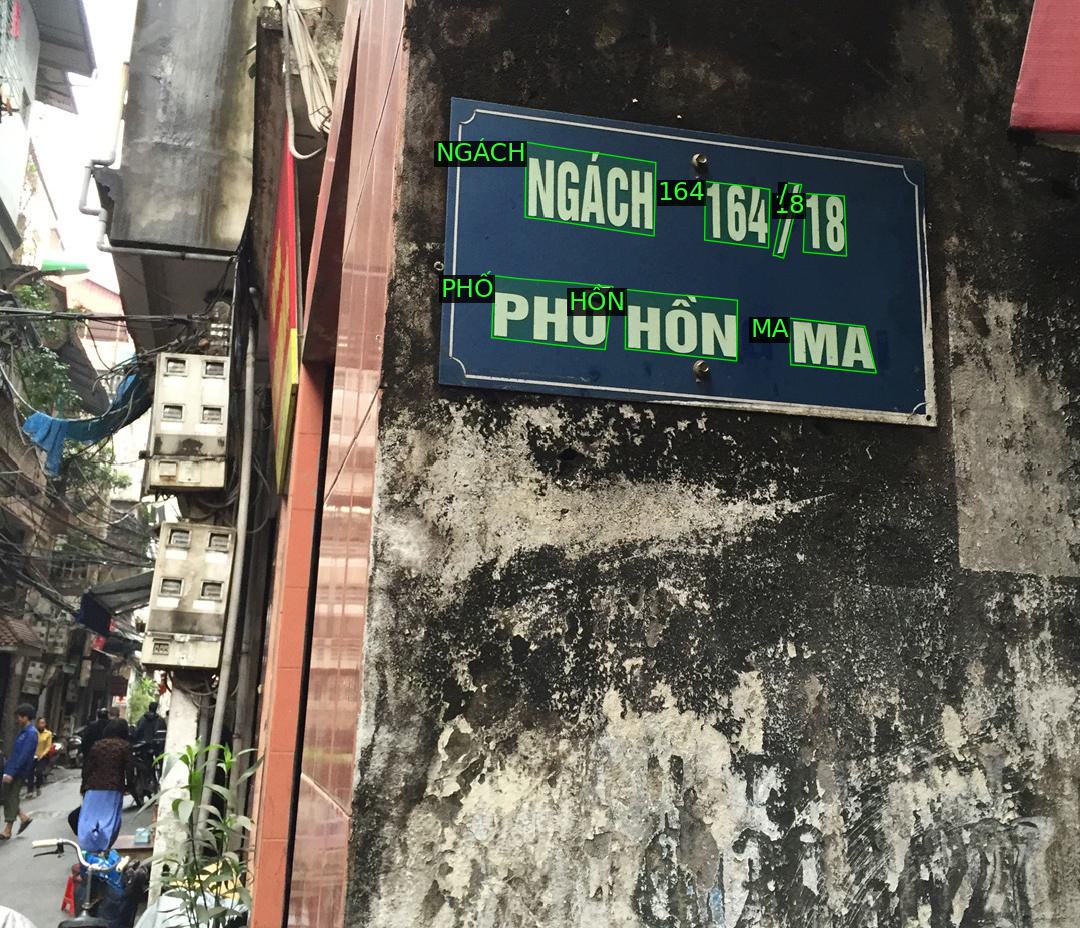}&
\includegraphics[height=4cm, width= 4cm]{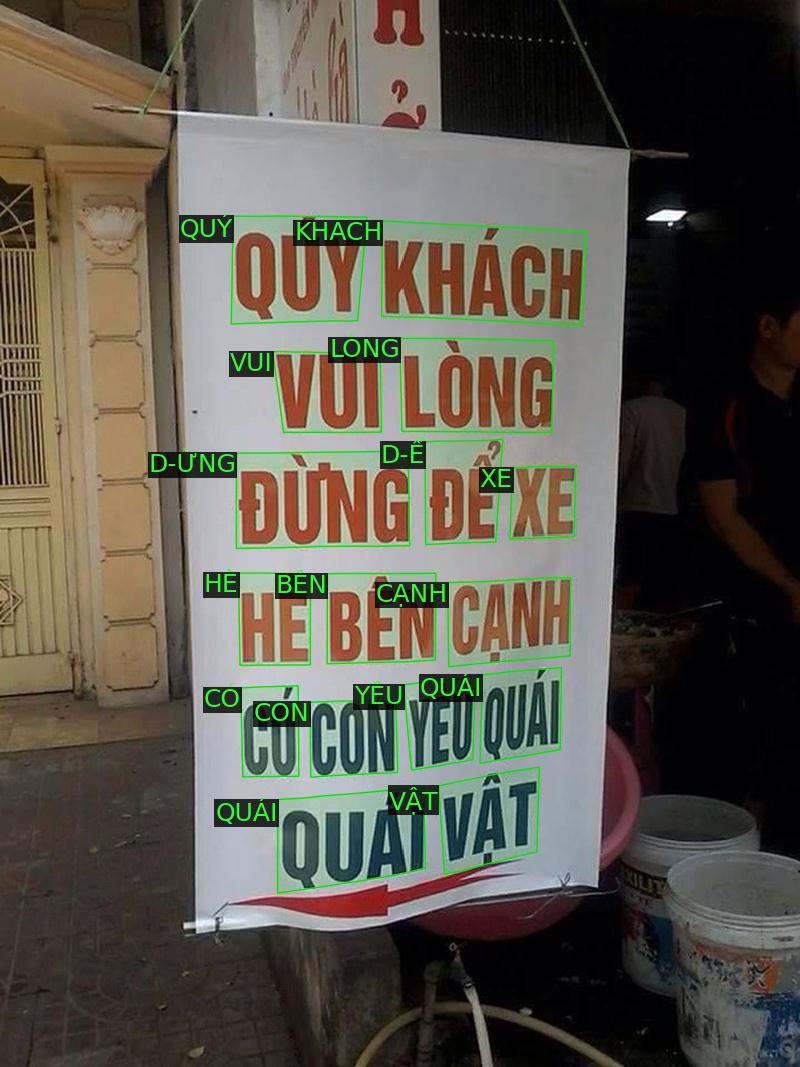}&
\includegraphics[height=4cm, width= 4cm]{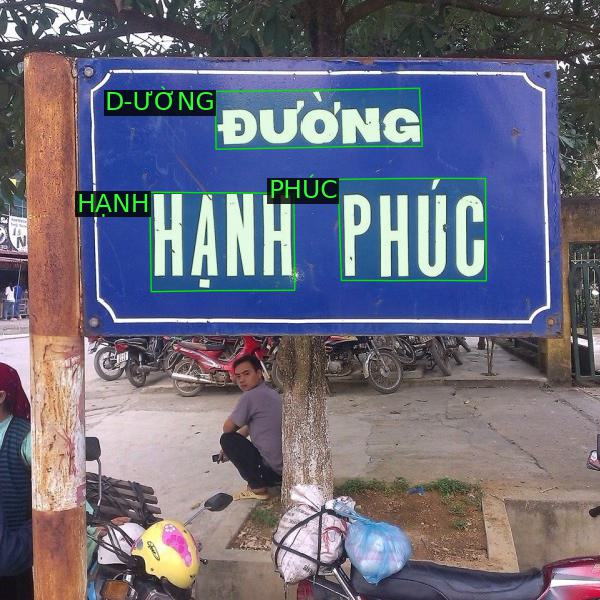}&
\includegraphics[height=4cm, width= 4cm]{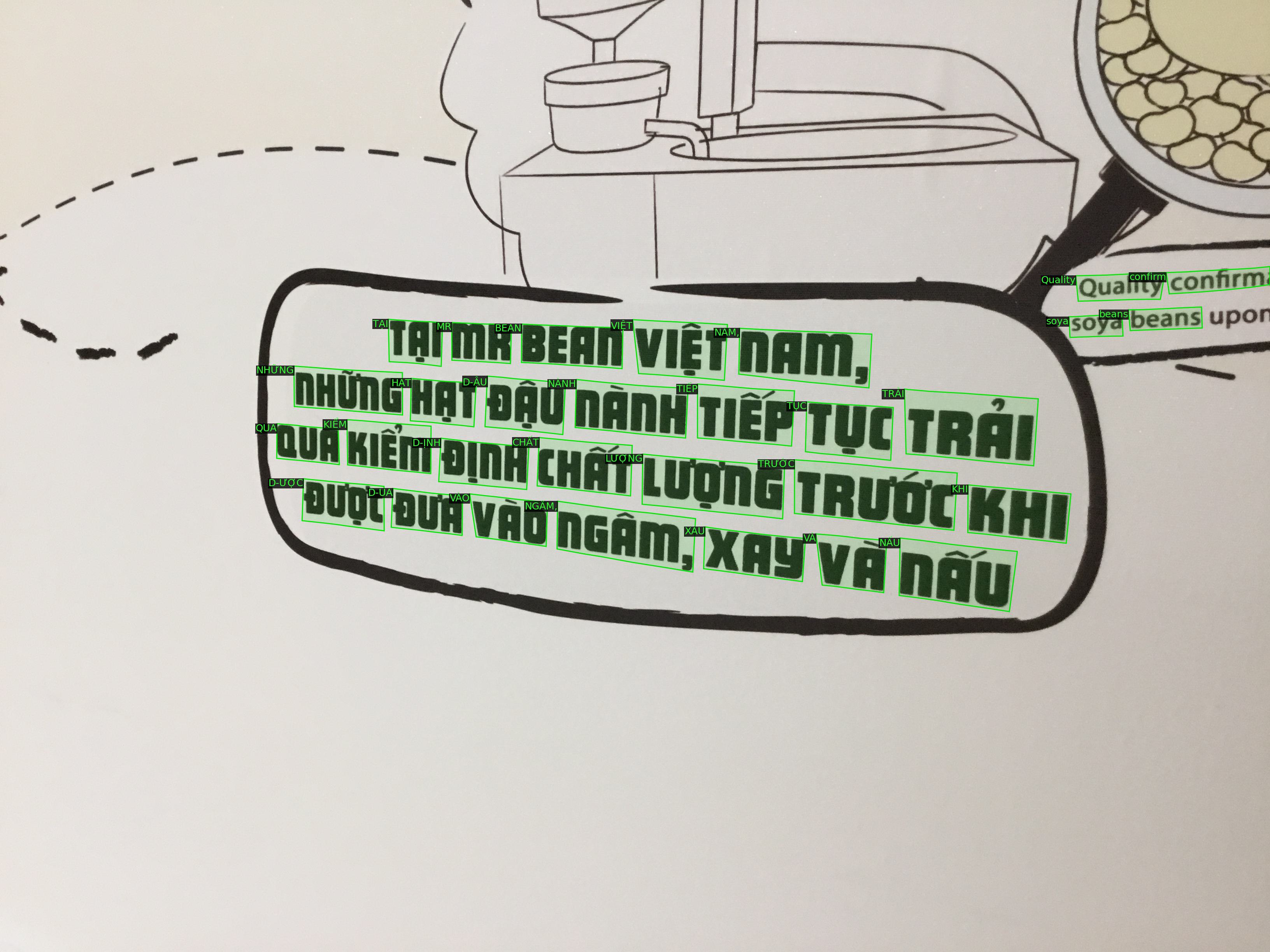}\\
(e)&(f)&(g)&(h)\\
\includegraphics[height=4cm, width= 4cm]{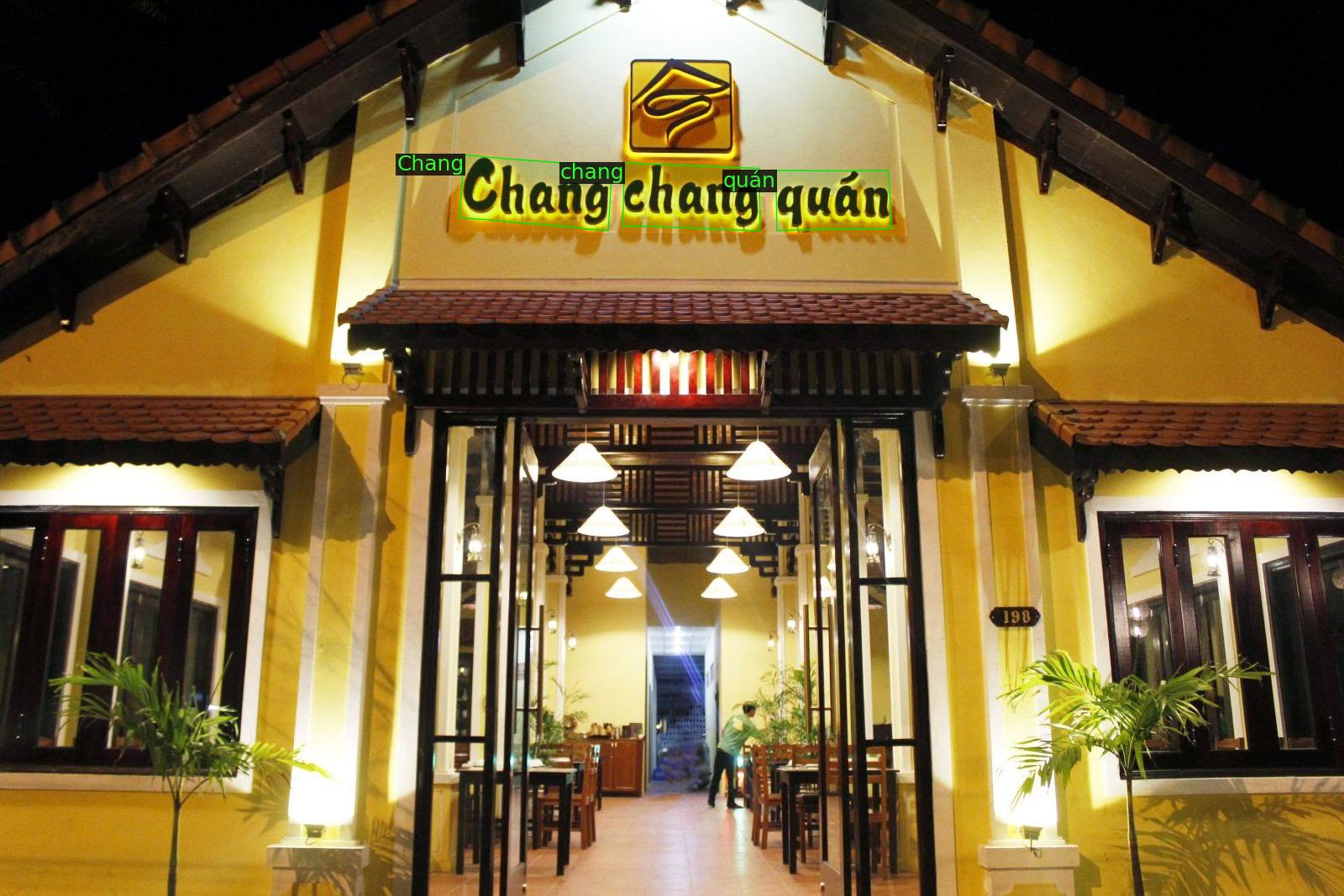}&
\includegraphics[height=4cm, width= 4cm]{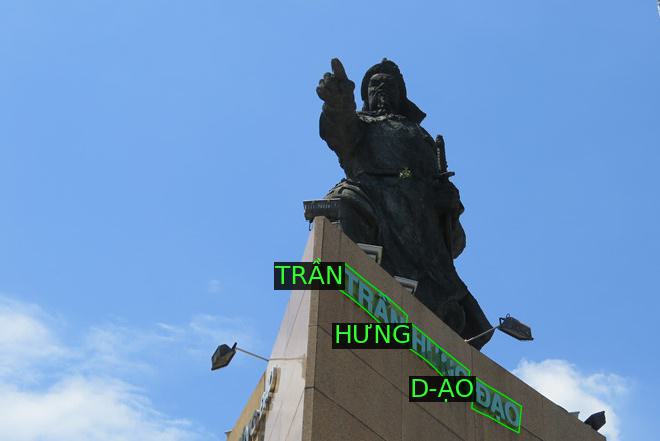}&
\includegraphics[height=4cm, width= 4cm]{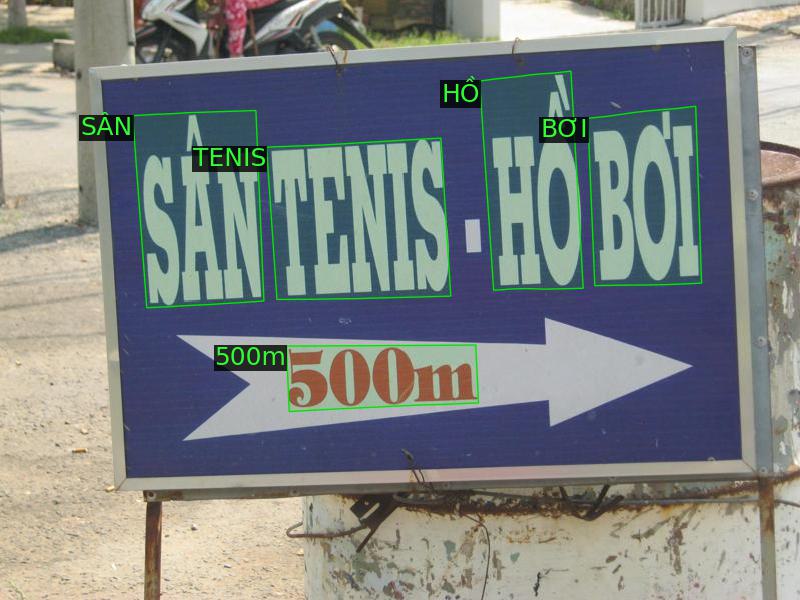}&
\includegraphics[height=4cm, width= 4cm]{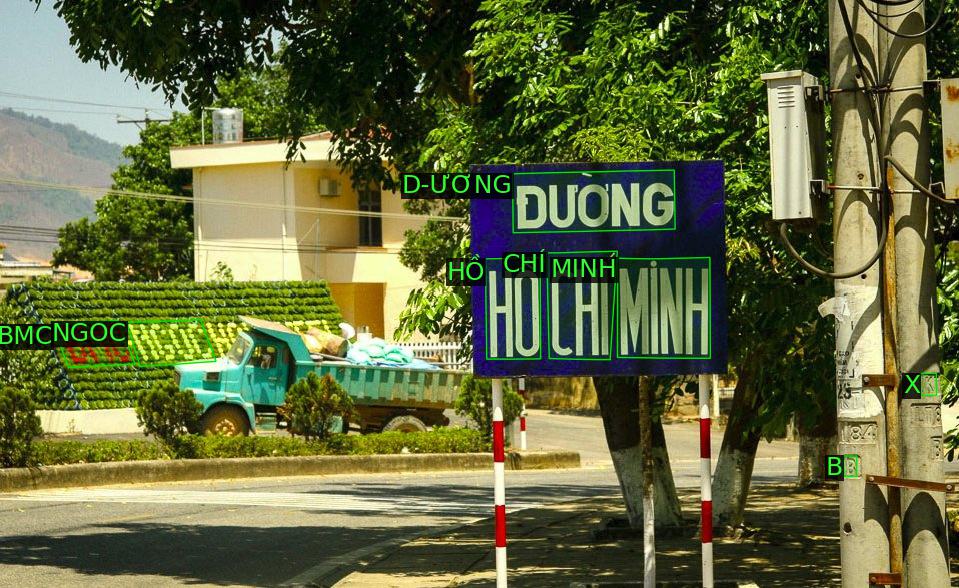}\\
(i)&(j)&(k)&(l)\\
\end{tabular}
\caption{\textbf{Performance Analysis on Low Resource VinText.} Some qualitative test cases of our method on images from the Vietnamese Text dataset.}
\label{fig:figsupvin}
\end{figure*}

\begin{figure*}[ht]
\centering
\begin{tabular}{cccc} 
\includegraphics[height=4cm, width= 4cm]{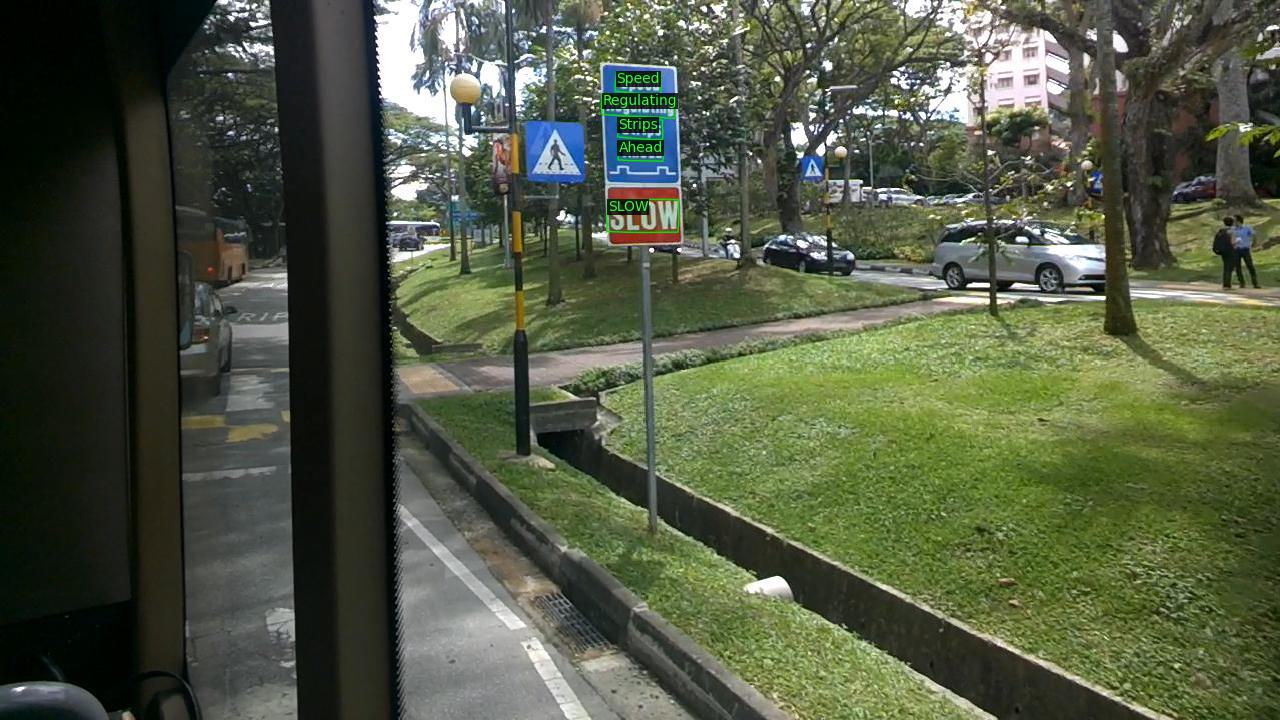}&
\includegraphics[height=4cm, width= 4cm]{supply_images/img_10.jpg}&
\includegraphics[height=4cm, width= 4cm]{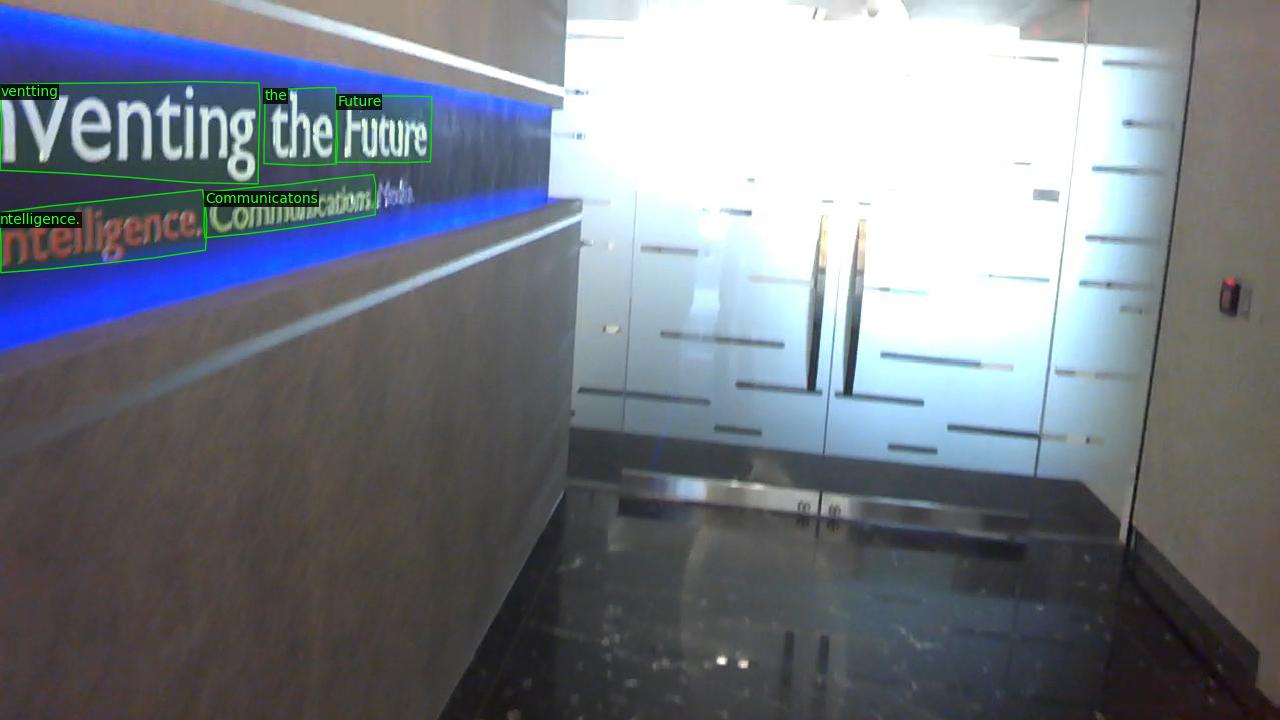}&
\includegraphics[height=4cm, width= 4cm]{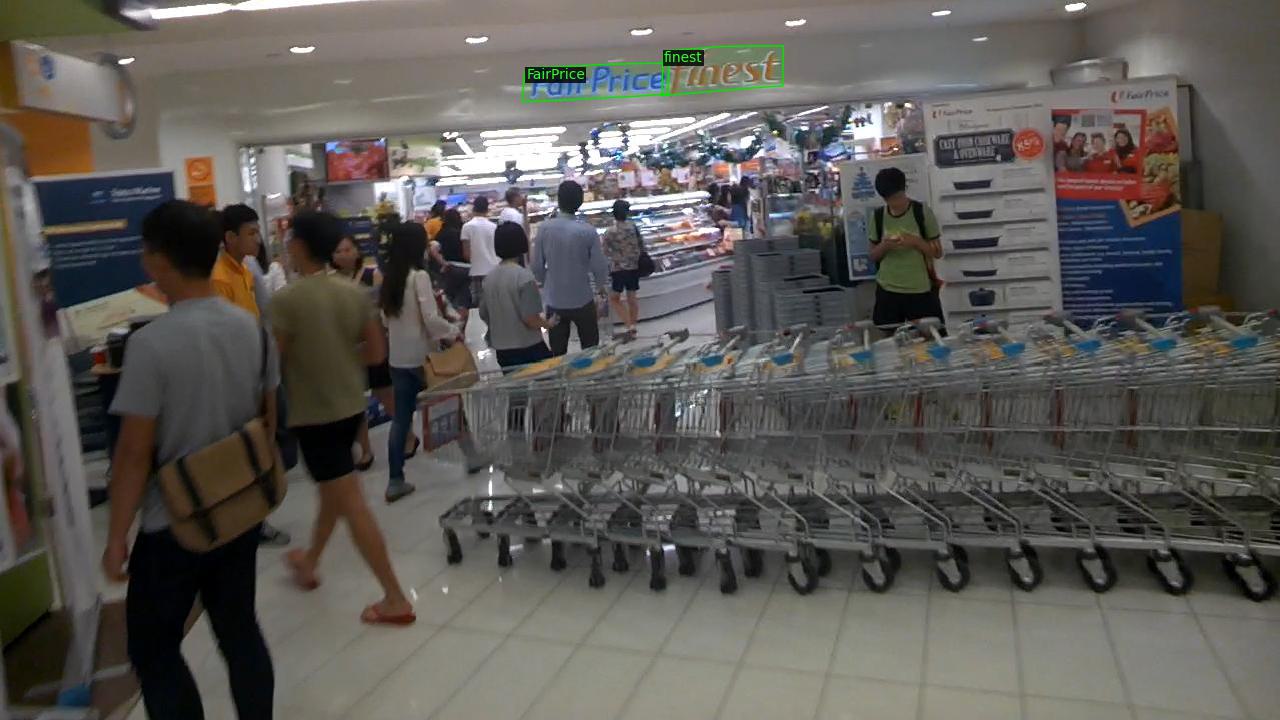}\\
(a)&(b)&(c)&(d)\\
\includegraphics[height=4cm, width= 4cm]{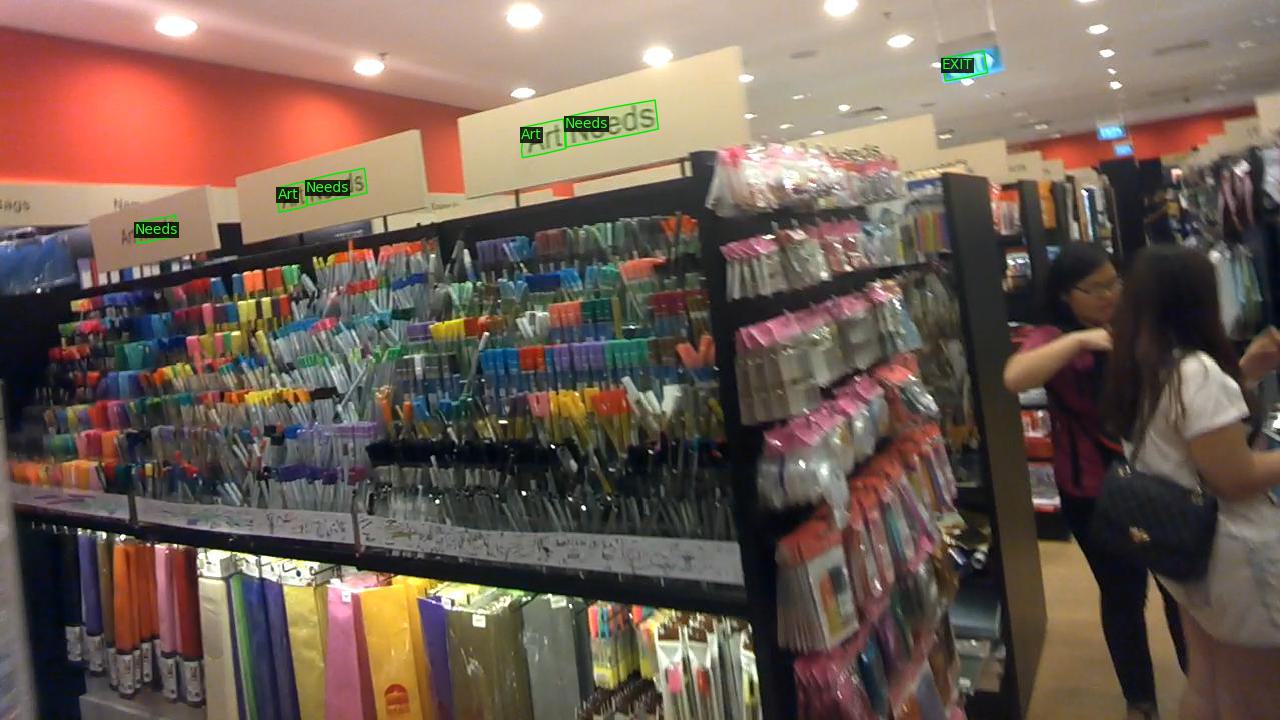}&
\includegraphics[height=4cm, width= 4cm]{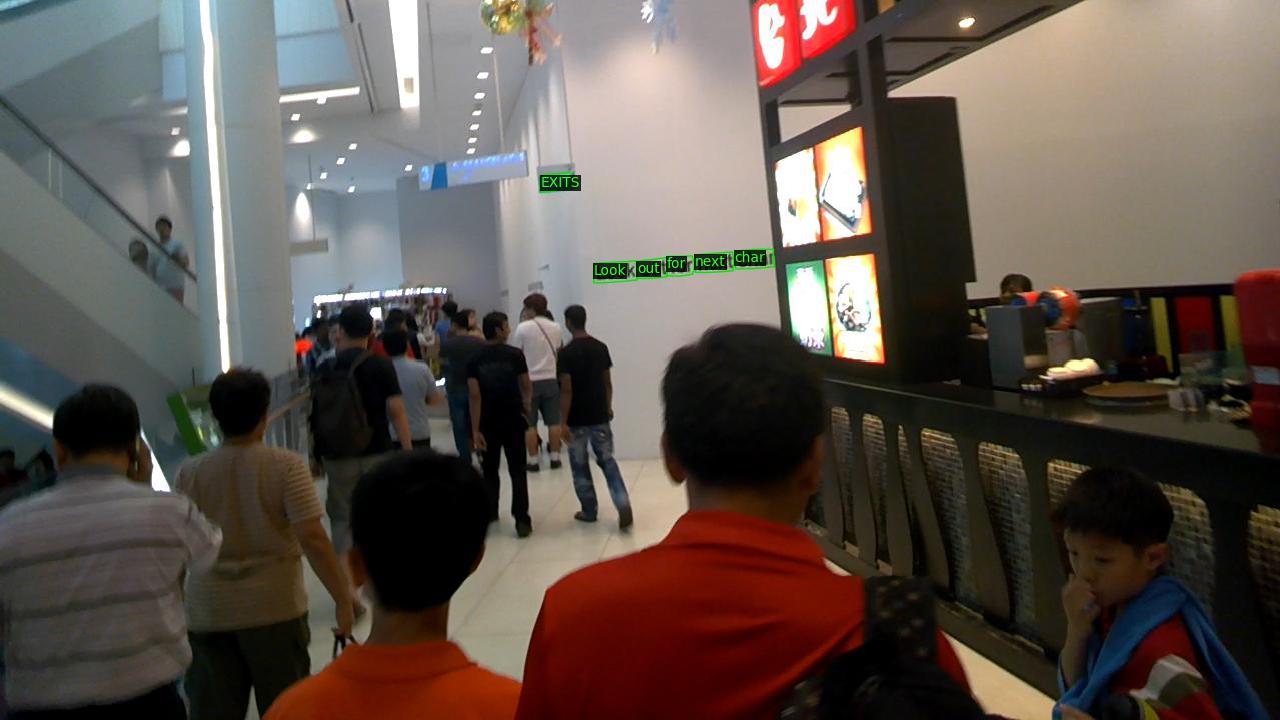}&
\includegraphics[height=4cm, width= 4cm]{supply_images/img_57.jpg}&
\includegraphics[height=4cm, width= 4cm]{supply_images/img_90.jpg}\\
(e)&(f)&(g)&(h)\\
\includegraphics[height=4cm, width= 4cm]{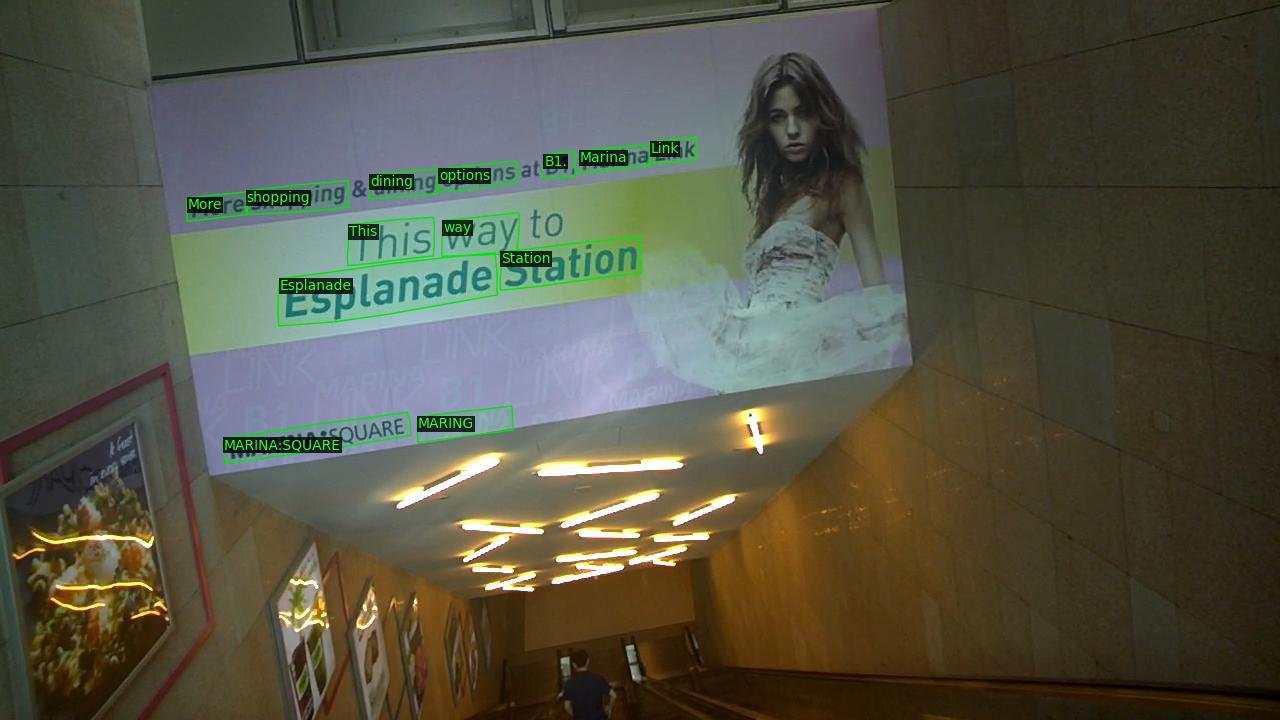}&
\includegraphics[height=4cm, width= 4cm]{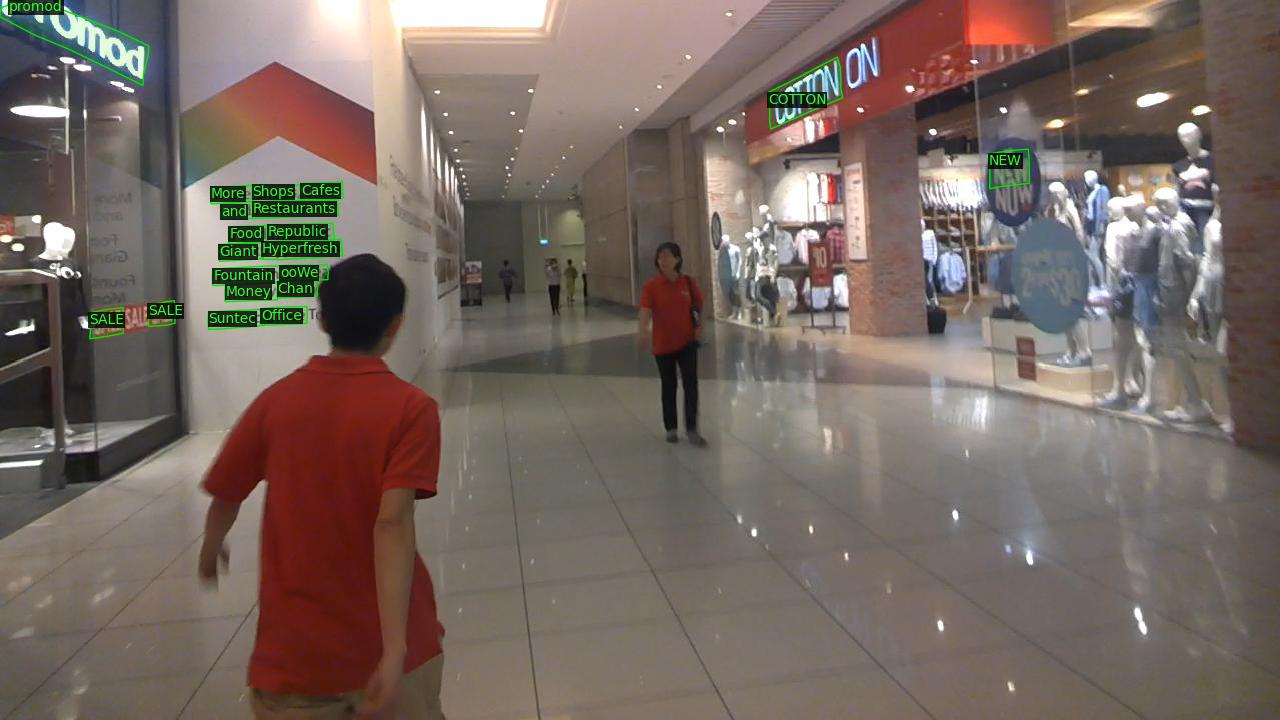}&
\includegraphics[height=4cm, width= 4cm]{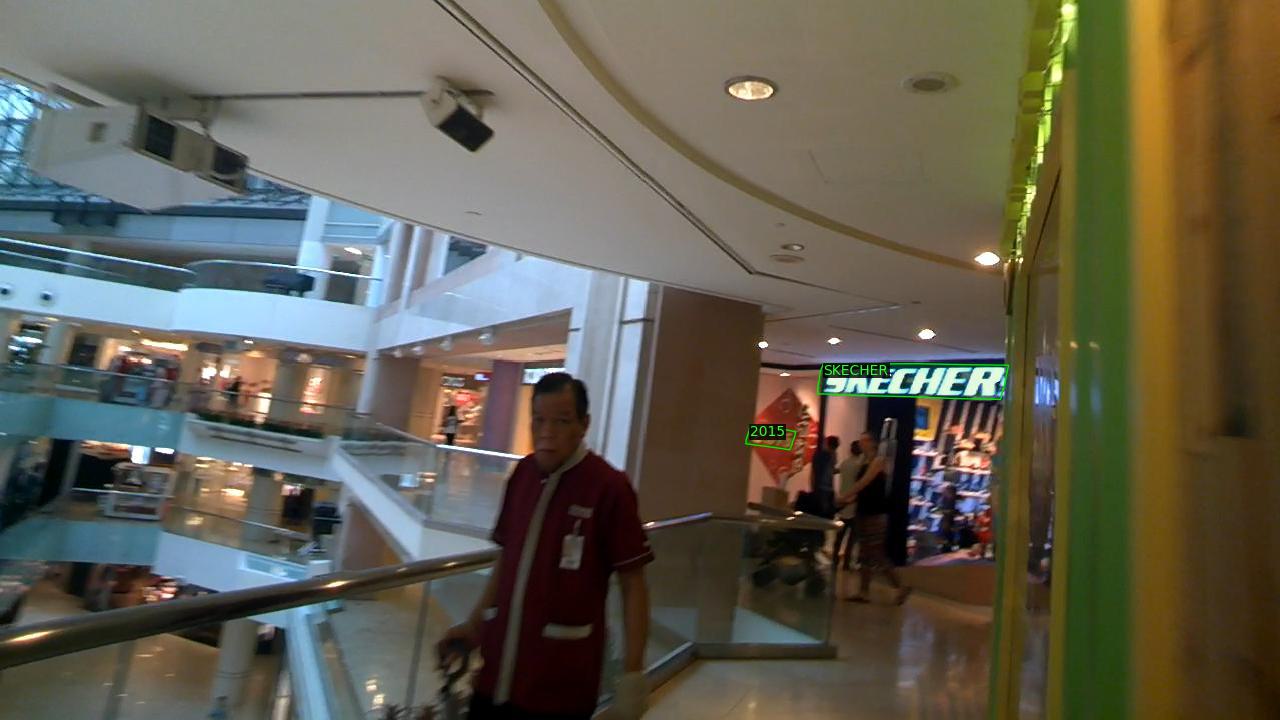}&
\includegraphics[height=4cm, width= 4cm]{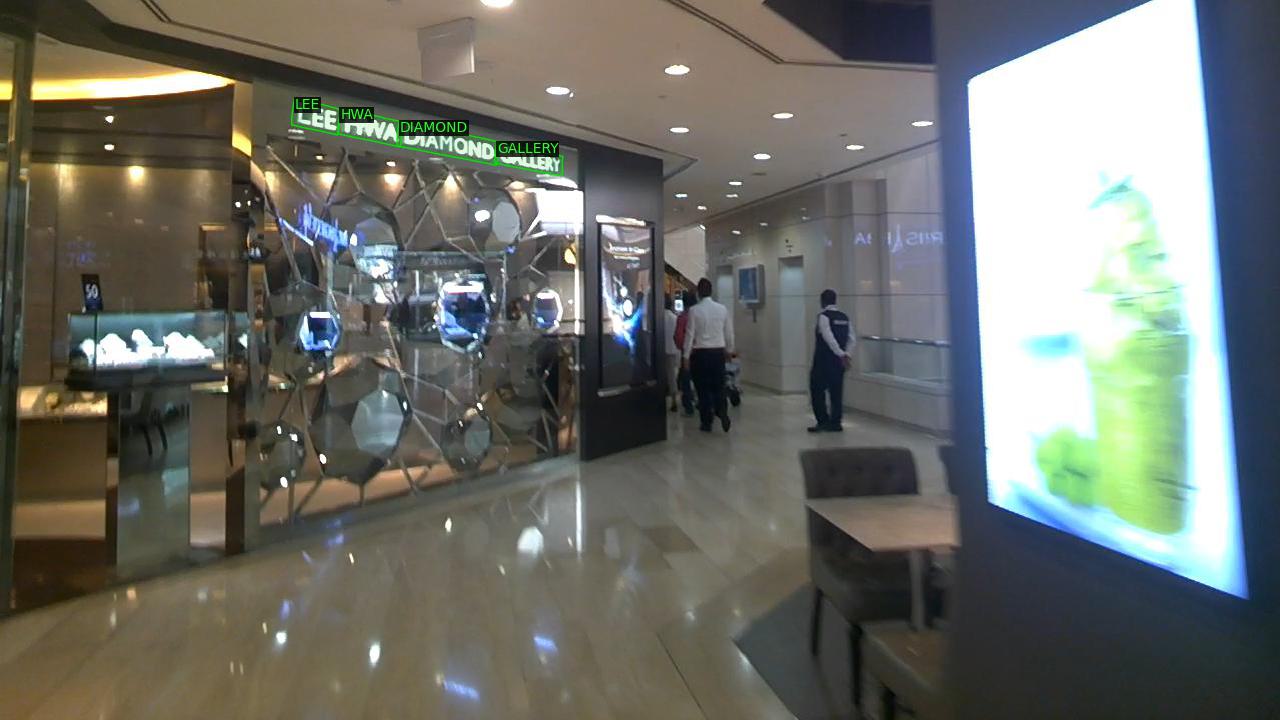}\\
(i)&(j)&(k)&(l)\\
\end{tabular}
\caption{\textbf{Performance Analysis on ICDAR15.} Some qualitative test cases of our method on images from ICDAR15.}
\label{fig:figsupic15}
\end{figure*}

\begin{figure*}[ht]
\centering
\begin{tabular}{c c c c} 
\includegraphics[height=4cm, width= 4cm]{failure/0000048.jpg}&
\includegraphics[height=4cm, width= 4cm]{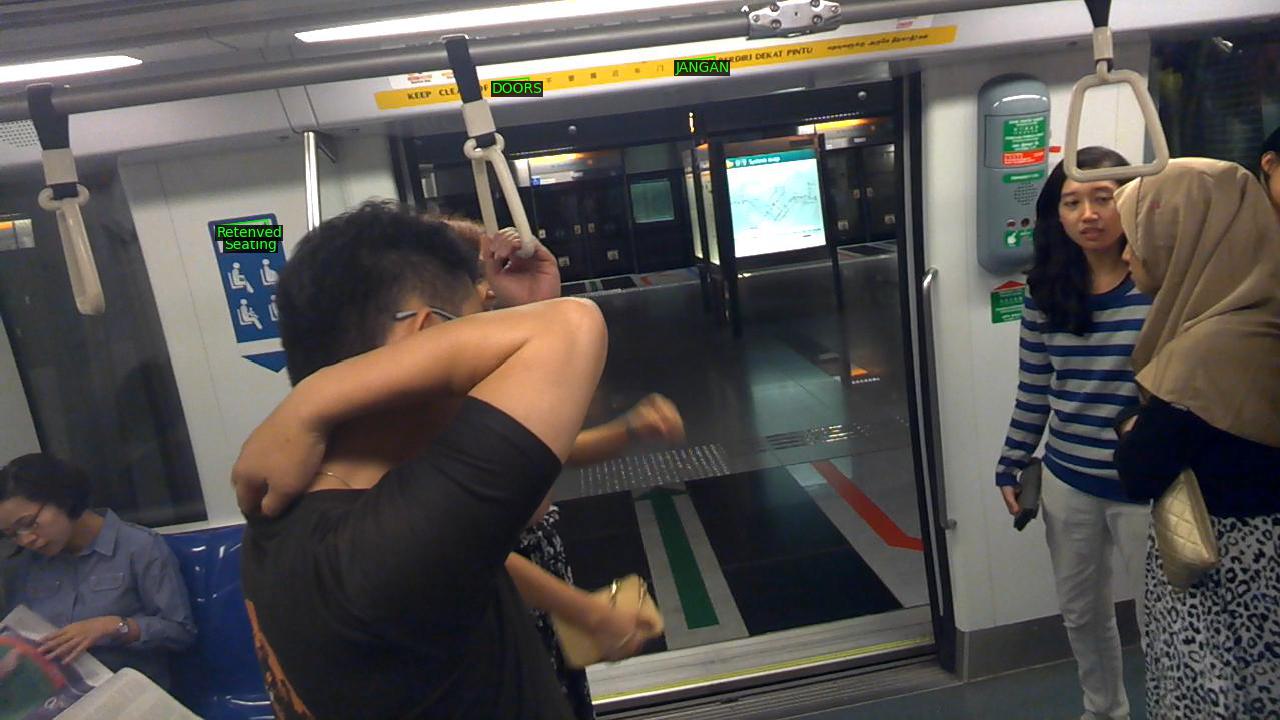}&
\includegraphics[height=4cm, width= 4cm]{failure/im1507.jpg}&
\includegraphics[height=4cm, width= 4cm]{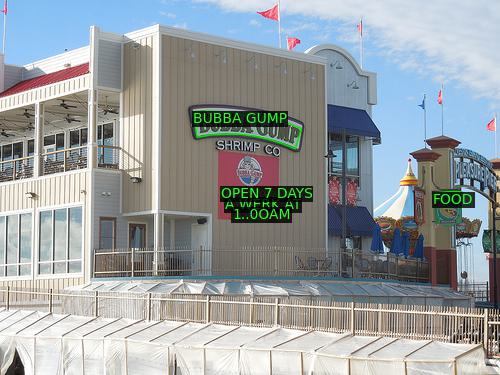}\\
(a)&(b)&(c)&(d)\\
\includegraphics[height=4cm, width= 4cm]{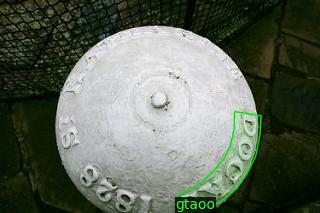}&
\includegraphics[height=4cm, width= 4cm]{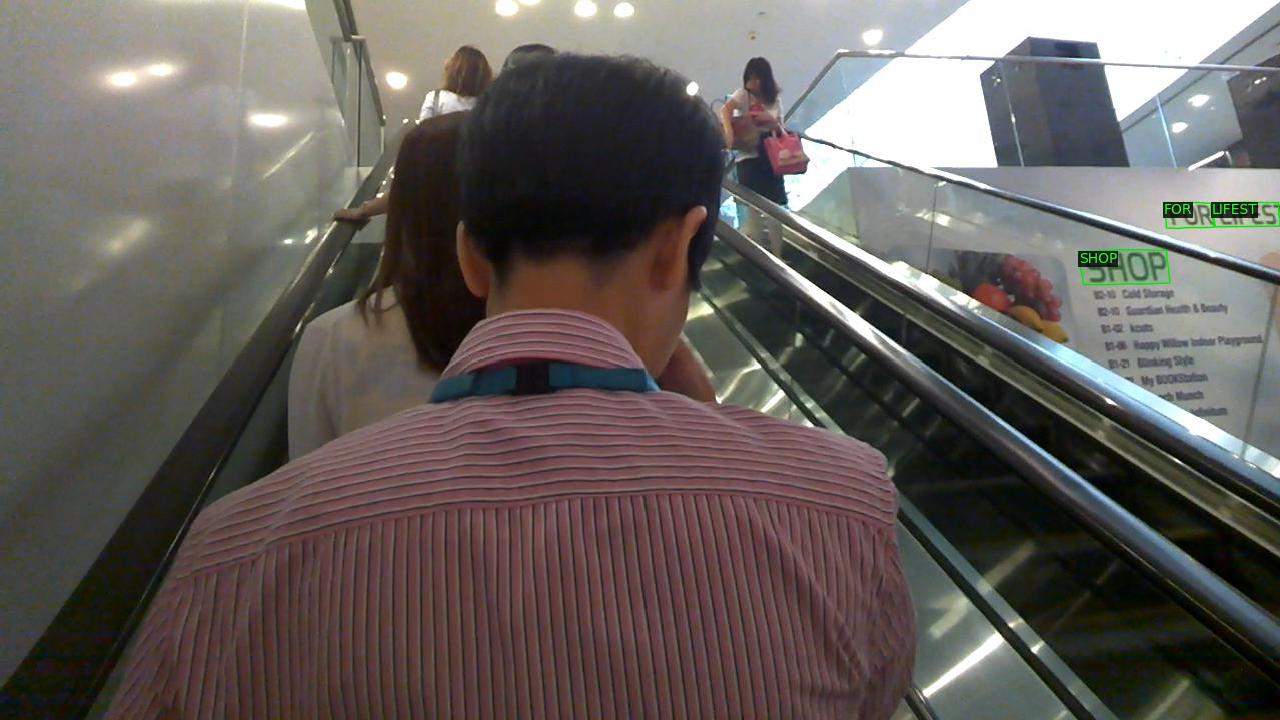}&
\includegraphics[height=4cm, width= 4cm]{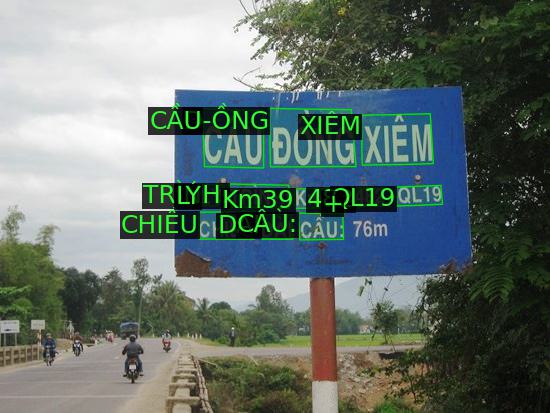}&
\includegraphics[height=4cm, width= 4cm]{failure/1009.jpg}\\
(e)&(f)&(g)&(h)\\
\end{tabular}
\caption{\textbf{Failure Cases.} Illustration of some failure cases of our method a and e from Total Text, b and f are from ICDAR15, c and g are from Vintext and d and h from CTW1500 dataset.}
\label{fig:figfail}
\end{figure*}
